\PassOptionsToPackage{bookmarksnumbered,unicode}{hyperref}
\RequirePackage{hyperref}
\documentclass[sigconf, nonacm]{acmart}

\usepackage{color}
\usepackage{graphicx}
\usepackage{subfigure}
\usepackage{mathtools}
\usepackage{makecell, multirow}
\usepackage{multirow}
\usepackage{bbding}
\usepackage{pifont}
\usepackage{wasysym}

\usepackage{amssymb}
\usepackage{algorithm}
\usepackage{algorithmic}
\usepackage[capitalise]{cleveref}
\usepackage{amsmath}

\usepackage{xspace}
\usepackage{enumitem}

\newcommand{\stitle}[1]{\vspace{2pt}\noindent\textbf{#1}}

\newcommand\vldbpagestyle{plain}
\newcommand{\error}{{\textsf{error}}}
\newcommand{\validerror}{{\textsf{validation\_error}}}

\DeclareMathOperator*{\argmin}{arg\,min}
\newtheorem{definition}{Definition}

\renewcommand\footnotetextcopyrightpermission[1]{} 
\begin{document}

\pagestyle{plain} 

\title{Auto-FP: An Experimental Study of Automated Feature Preprocessing for Tabular Data \\ (Technical Report)}

\author{{Danrui Qi$^{\dagger*}$, Jinglin Peng$^{\dagger*}$, Yongjun He$^{\diamondsuit*}$, Jiannan Wang$^{\dagger}$}}
\affiliation{%
  \institution{\hspace{0em} Simon Fraser University$^{\dagger}$ \hspace{7em}  ETH Z\"urich$^{\diamondsuit}$ \\
  {\{dqi, jinglin\_peng, jnwang\}@sfu.ca \quad \quad yongjun.he@inf.ethz.ch}}
  \country{} 
}
\thanks{* The first three authors contributed equally to this research.}


\begin{abstract}
Classical machine learning models, such as linear models and tree-based models, are widely used in industry. These models are sensitive to data distribution, thus feature preprocessing, which transforms features from one distribution to another, is a crucial step to ensure good model quality. Manually constructing a feature preprocessing pipeline is challenging because data scientists need to make difficult decisions about which preprocessors to select and in which order to compose them. In this paper, we study how to automate feature preprocessing (Auto-FP) for tabular data. Due to the large search space, a brute-force solution is prohibitively expensive. To address this challenge, we interestingly observe that Auto-FP can be modelled as either a hyperparameter optimization (HPO) or a neural architecture search (NAS) problem. This observation enables us to extend a variety of HPO and NAS algorithms to solve the Auto-FP problem. We conduct a comprehensive evaluation and analysis of 15 algorithms on 45 public ML datasets. Overall, evolution-based algorithms show the leading average ranking. Surprisingly, the random search turns out to be a strong baseline. Many surrogate-model-based and bandit-based search algorithms, which achieve good performance for HPO and NAS, do not outperform random search for Auto-FP. We analyze the reasons for our findings and conduct a bottleneck analysis to identify the opportunities to improve these algorithms. Furthermore, we explore how to extend Auto-FP to support parameter search and compare two ways to achieve this goal. In the end, we evaluate Auto-FP in an AutoML context and discuss the limitations of popular AutoML tools. To the best of our knowledge, this is the first study on automated feature preprocessing. We hope our work can inspire researchers to develop new algorithms tailored for Auto-FP.
We release our datasets, code, and comprehensive experimental results at {\color{blue}\url{https://github.com/AutoFP/Auto-FP}}.
\end{abstract}


\maketitle

\section{Introduction}

Despite the recent advancement of deep learning for image and text data, most machine learning (ML) use cases in industries are still about applying classical ML algorithms to tabular data. For example, a recent survey of over 25,000 data scientists and ML engineers~\cite{Kaggle-2021-survey} shows that the most commonly used models are linear models (linear regression or logistic regression); the most popular ML framework is scikit-learn~\cite{scikit-learn} (mainly designed for traditional machine learning). 

Building a classical ML model on tabular data involves several tasks, such as feature preprocessing, feature selection, and hyperparameter tuning. An important question faced by many real-world data scientists is how to automatically complete each task. Fortunately, many research efforts have been devoted to answering this question. They conduct  comprehensive surveys or experiments on feature selection~\cite{chandrashekar2014survey,li2017feature,forman2003extensive}, hyperparameter tuning~\cite{feurer2019hyperparameter,yang2020hyperparameter},  data cleaning~\cite{CleanML},  feature type inference~\cite{FeatureTypeBench}, etc.  However, \emph{feature preprocessing},  an essential task for classical ML, has not been well explored in the literature. This paper aims to fill this research gap.

There are many commonly used feature preprocessors in  scikit-learn, such as \textit{Binarizer}, \textit{MaxAbsScaler}, \textit{MinMaxScaler},  \textit{Normalizer}, \textit{PowerTransformer}, \textit{QuantileTransformer} and \textit{StandardScaler}. Intuitively, a feature preprocessor transforms features from one distribution to another.  For example,  \textit{MinMaxScaler} transforms features by scaling each feature to a given range. \textit{PowerTransformer} applies an exponential transformation to each feature to make its distribution more normal-like.
Given a training dataset and a set of feature preprocessors~\cite{scikit-learn-preprocessing}, the goal of feature preprocessing is to construct a pipeline (i.e., a sequence of selected feature preprocessors) to scale and transform the features in the training set.  

To construct a good pipeline, we need to answer two questions: i) which preprocessors to select; ii) in which order to compose them. Different answers will lead to different pipelines. Consider the following two pipelines:

\begin{align*}\nonumber
    P1: & ~~\textit{MinMaxScaler}  \rightarrow \textit{PowerTransformer}   \\
    P2: & ~~\textit{PowerTransformer}  \rightarrow \textit{MinMaxScaler} \rightarrow \textit{Normalizer}
\end{align*}

\noindent $P1$ selects two preprocessors and applies \textit{MinMaxScaler} first and then \textit{PowerTransformer}; $P2$ selects three preprocessors and applies \textit{PowerTransformer} first and then \textit{MinMaxScaler} and  \textit{Normalizer}.

It is not easy to use domain knowledge to determine which pipeline ($P1$ or $P2$) is better since it not only depends on the upstream training set but also on the downstream learning algorithm. To \emph{manually} construct a pipeline, data scientists have to get into a trial-and-error mode, which will be both tedious and time-consuming. If a bad feature preprocessing pipeline is selected, it could even hurt the downstream model accuracy. To this end, this paper studies how to \emph{automatically} search for the best feature preprocessing pipeline.

The brute-force solution that enumerates all possible pipelines is prohibitively expensive. Given a set of $n$ preprocessors, a pipeline could contain 1, 2, $\cdots$, or $n$ preprocessors, thus there will be a total of  $\sum_{i=1}^{n} i^i \approx n^{n}$ pipelines to enumerate. 
Apparently, it is too expensive for the brute-force solution to enumerate all pipelines. We interestingly find the insight that the Auto-FP problem can be modelled as either a Hyperparameter Optimization (HPO) problem or a Neural Architecture Search (NAS) problem. This interesting insight enables us to extend many existing search algorithms from HPO and NAS to Auto-FP. With these search algorithms, we need to answer the following essential questions:


\stitle{Q1. Which search algorithm performs better?}
There are many HPO and NAS search algorithms available, but it is unclear which algorithms will perform well for Auto-FP. Unlike HPO and NAS, Auto-FP focuses on optimizing a preprocessing pipeline, i.e., aiming to find the best combination as well as the best order of preprocessors, thus it has a unique search space for these search algorithms to explore. An algorithm performing well for HPO and NAS does not mean that it will also perform well for Auto-FP.  


\stitle{Q2. How to extend Auto-FP to support parameter search?}
To further enhance the performance of Auto-FP, we consider the scenario where a user wants to search for not only the best pipeline but also the best parameters associated with each preprocessor. The characteristics such as the cardinality of parameter search spaces can be varied and it is valuable to explore how to extend Auto-FP to support different search spaces. 

\stitle{Q3. What is the relationship between Auto-FP and AutoML?}
To further explore future opportunities, we put Auto-FP in an AutoML context to figure out the relationship between Auto-FP and AutoML. AutoML tools are typically equipped with a feature preprocessing module. One natural question is whether Auto-FP can outperform the FP part of a general-purpose AutoML. Another question is whether Auto-FP is important in the AutoML context. If yes, they can learn from our paper to improve their feature preprocessing module. 


\begin{figure}[t] 
\centering 
\includegraphics[width=1.0\linewidth]{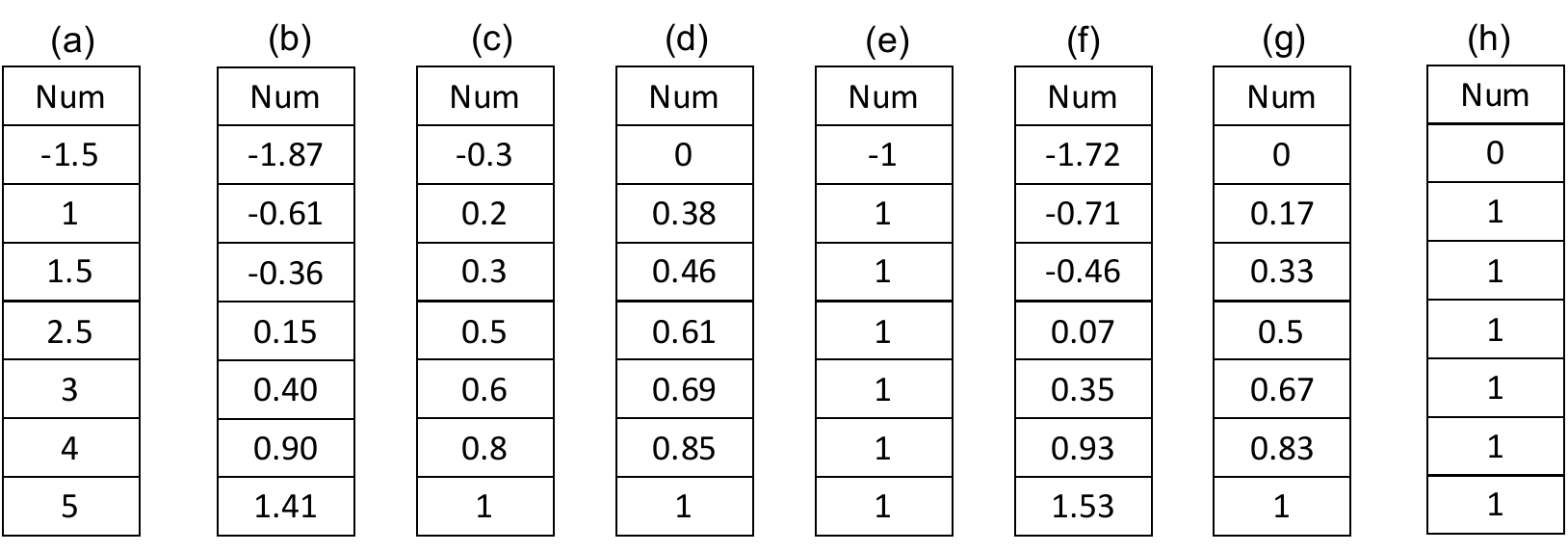}
\caption{Illustration of different feature preprocessors. (a) No preprocessor; (b) StandardScaler; (c) MaxAbsScaler; (d) MinMaxScaler; (e) Normalizer; (f) PowerTransformer; (g) QuantileTransformer; (h) Binarizer.} 
\label{fig:preprocessors_example}
\end{figure}
\stitle{Our Contributions.} For answering the above essential questions, our work makes the following contributions:

\begin{enumerate}[leftmargin=*]\itemsep0.15em
    \item \textbf{\textit{Indicate the importance of FP.}} Feature preprocessing is a crucial step  in classical ML. To the best of our knowledge, this is the first comprehensive study on this important topic. 
    \item \textbf{\textit{Formally define Auto-FP.}} We identify a number of widely used preprocessors in scikit-learn and formally define the Auto-FP problem. We find that this problem can be modelled as either a HPO or a NAS problem, and extend 15 search algorithms from HPO and NAS to Auto-FP.
    \item \textbf{\textit{Categorize Auto-FP search algorithms.}} We conclude these search algorithms into a unified framework and categorize them into 5 categories: traditional algorithms, surrogate-model-based algorithms, evolution-based algorithms, RL-based algorithms and bandit-based algorithms.
    \item \textit{\textbf{To answer Q1, Q2 and Q3}}, we first conduct comprehensive experiments comparing 15 Auto-FP algorithms on 45 public ML datasets and provide recommendations for different scenarios. Then we explore two parameter search spaces with varying cardinalities and propose two extensions for Auto-FP, namely One-step (combining parameter and pipeline search) and Two-step (separating the searches). Thirdly, by situating Auto-FP within an AutoML context, we identify the limitations of current AutoML tools, examine if Auto-FP outperforms FP in AutoML tools, and highlight the significance of Auto-FP by comparing it to hyperparameter tuning in AutoML.
\end{enumerate}

\stitle{Experimental Findings.}
Our empirical results reveal a number of interesting findings and we summarize our main experimental findings as follows:

\begin{enumerate}[leftmargin=*]\itemsep0.15em
    \item Evolution-based search algorithms typically achieve the highest overall average ranking. 
    \item Random search is a strong baseline. 
    \item Many sophisticated algorithms, such as RL-based, bandit-based and most surrogate-model-based algorithms, perform even worse than random search.
    \item Different algorithms have different bottlenecks and model evaluation is the bottleneck in most cases.
    \item \textit{One-step} fits for extended low-cardinality search space and \textit{Two-step} is preferred for extended high-cardinality search space.
    \item Auto-FP outperforms the FP part in AutoML and it is important in the AutoML context. 
\end{enumerate}

\stitle{Research Opportunities.} According to our findings, we identify four future research directions: 1) warm-start search algorithms 2) reduce data size intelligently to mitigate performance bottleneck 3) allocate pipeline and parameter search time budget reasonably 4) benchmark other tasks in AutoML context. The code, datasets and experimental raw data are published on GitHub as a reference for the community. We hope our work can attract more research interest not only for Auto-FP but also for automating other individual tasks (such as data cleaning~\cite{AlphaClean}, feature generation and feature selection~\cite{autofeat, ExploreKit, SAFE, LearningFE, FICUS}) in AutoML context and building more powerful AutoML.

The rest of this paper is organized as follows. In Section 2, We present the background and motivation. We formally define the Auto-FP problem and show that it can be modelled as a HPO or a NAS problem in Section~3. We show that existing search algorithms all fall into a unified framework and present a taxonomy to relate and differentiate them in Section~4. In Section 5, we conduct extensive experiments to evaluate different search algorithms. Section~6 and Section~7 evaluate Auto-FP in an extended search and AutoML context, respectively. Research opportunities are presented in Section~8. Section~9 shows the related work. In Section~10, we present the conclusion of our paper.

\section{Background and Motivation}\label{background-motivation}

In many machine learning applications, users enhance the accuracy and effectiveness of their ML models by incorporating appropriate feature preprocessing techniques~\cite{statisticalLearning}. Feature preprocessors can prepare data for learning process by mainly performing the following functionalities: 1) Scale Variance Reduction: features with different scales may adversely impact the performance of certain algorithms, especially those that rely on distance calculations, such as k-Nearest Neighbors~\cite{KNN} and Support Vector Machines (SVM)~\cite{SVM}, scalers like \textit{StandardScaler} normalizing features to mitigates the issue of varying scales~\cite{jolliffe2016principal}. 2) Data Distribution Transformation: some algorithms, such as Logistic Regression and Gaussian Naive Bayes, assume that the input features follow a specific distribution, usually Gaussian~\cite{MLAPP}. Transformers like \textit{PowerTransformer} can transform the data to meet these assumptions, leading to better model performance. 3) Dimensionality Reduction: \textit{Binarizer} and other discretization techniques can reduce the dimensionality of the data, which can be beneficial in cases where high-dimensional data negatively impacts the model's performance due to the curse of dimensionality~\cite{hammer1962adaptive}. This can also improve the computational efficiency of the learning process. Selecting appropriate preprocessing techniques is vital for obtaining accurate and reliable results across a wide range of applications.

In this section, we first present the background of feature preprocessors, then validate whether feature preprocessing really matters, and finally explore the relationship between data characteristics and feature preprocessing. 

\subsection{Feature Preprocessors}
\label{operator}

\begin{table}[t]
\caption{3-CV scores of decision tree models with different tree depths for downstream ML models (LR, XGB, or MLP).}
\label{tab:tree_and_cv_scores}
\small
\begin{tabular}{|p{0.12\linewidth}<{\centering}
                |p{0.12\linewidth}<{\centering}
                |p{0.12\linewidth}<{\centering}
                |p{0.12\linewidth}<{\centering}
                |p{0.12\linewidth}<{\centering}
                |p{0.12\linewidth}<{\centering}|}
    \hline
    \multicolumn{2}{|p{0.28\linewidth}<{\centering}}{\textbf{LR}} & 
    \multicolumn{2}{|p{0.28\linewidth}<{\centering}}{\textbf{XGB}} & 
    \multicolumn{2}{|p{0.28\linewidth}<{\centering}|}{ \textbf{MLP}}   \\
    \hline
    \textbf{Tree Depth} & \textbf{3-CV Score} & \textbf{Tree Depth} & \textbf{3-CV Score} & \textbf{Tree Depth} & \textbf{3-CV Score}\\
    \hline
    1 & 0.65 & 1 & 0.70 & 1 & 0.54\\
    \hline
    2 & 0.65 & 2 & 0.67 & 2 & 0.65\\
    \hline
    3 & 0.54 & 3 & 0.67 & 3 & 0.67\\
    \hline
    No Limit & 0.50 & No Limit & 0.67 & No Limit & 0.67\\
    \hline
\end{tabular}
\end{table}

Scikit-learn \cite{scikit-learn} is a widely used tool for developing classical ML models on tabular data. It contains a feature preprocessing module with many preprocessors\footnote{https://scikit-learn.org/stable/modules/preprocessing.html}. The Scikit-Learn Documentation~\cite{Sklearn-doc} indicates the functionalities of feature preprocessors used in practice, and highlights the effectiveness of feature preprocessors. For example, scaling feature preprocessors such as Standardization and Normalization can be used for improving model performance. The frequency of using feature preprocessors is indicated by \cite{FES, DataPreparation}. They show the relatively lower frequency of use of discretization or binarization feature preprocessors compared to the scaling feature preprocessors. We introduce seven feature preprocessors in detail in the following contents, and provide a rigorous justification for choosing the listed feature preprocessors.

\textbf{1. \textit{StandardScaler: }} ML models may perform badly if individual features do not follow the standard normal distribution. \textit{StandardScaler} standardizes a feature by removing its mean value and scales it by dividing the standard deviation~\cite{scikit-learn}. Let $\mu$ and $\sigma$ denote the mean and the standard deviation of a feature, respectively. For each value $x$ in the feature, the scaled result is calculated by $\frac{x - \mu}{\sigma}$. Figure \ref{fig:preprocessors_example}(b) shows the result of applying \textit{StandardScaler}. The $\mu$ and $\sigma$ are 2.21 and 1.98.  Thus, the transformed result of value -1.5 is $\frac{-1.5-2.21}{1.98} =-1.87$. 

\begin{figure*}[t]
\centering
\includegraphics[width=\linewidth]{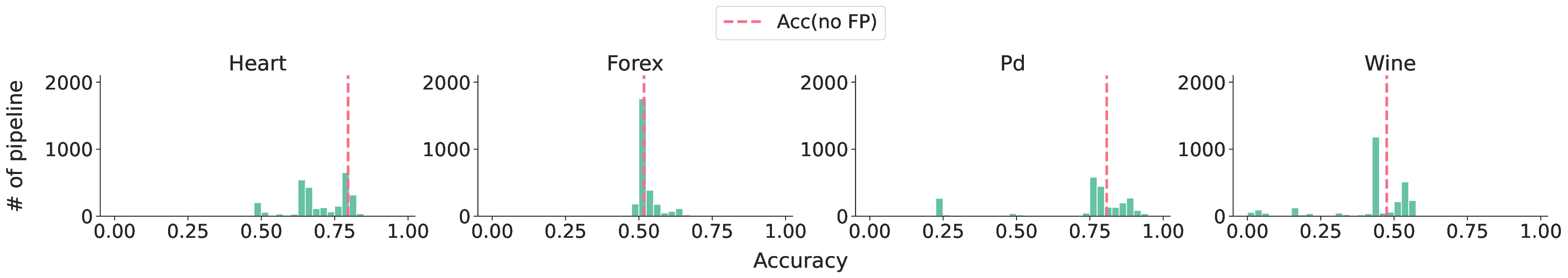}
\caption{Distribution of LR accuracy with different feature preprocessing pipelines.}
\label{fig:order_impact}
\end{figure*}

\textbf{2. \textit{MaxAbsScaler: }} When the standard deviations of some features are very small like $10^{-8}$, applying \textit{StandardScaler} can cause unreasonable transformed features. In this situation, it is better to scale the values of each feature into a specific range. \textit{MaxAbsScaler} scales each feature according to its maximum absolute value~\cite{Sklearn-doc}. Let $x$ be the maximum absolute value of the feature. For each value $v_1, v_2, \cdots$ of the feature, the scaled values will be $\frac{v_1 }{x}, \frac{v_2}{x}, \cdots$. Figure \ref{fig:preprocessors_example}(c) shows the result of applying \textit{MaxAbsScaler} to the original feature. The maximum absolute value of the original feature is 5. Thus, the value -1.5 is transformed to $\frac{-1.5}{5} = -0.3$.

\textbf{3. \textit{MinMaxScaler: }} Similar to \textit{MaxAbsScaler},  \textit{MinMaxScaler} transforms features by scaling each feature to a given range~\cite{Sklearn-doc} ([0, 1] by default).  Let $\max$ and $\min$ denote the maximum and minimum values of the feature, respectively. The scaled result of $x$ is calculated as $\frac{x - min}{max - min}$. Figure \ref{fig:preprocessors_example}(d) shows the result of applying \textit{MinMaxScaler} to the original feature. The max (min) value of the original feature is 5 (-1.5).  Thus, the value 1 is transformed to $\frac{1-(-1.5)}{5-(-1.5)}=0.38$.

\textbf{4. \textit{Normalizer: }} ML models can be impacted severely by different scales of the features. \textit{Normalizer} normalizes samples (\emph{rows} of data) individually into unit norm~\cite{Sklearn-doc}. Given a row vector, $x = [x_1, x_2, \cdots, x_n]$, each value $x_i$ is scaled to $\frac{x_i}{||x||_{2}}$. Suppose the data only has a single column as shown in Figure~\ref{fig:preprocessors_example}(a). Figure~\ref{fig:preprocessors_example}(e) shows the result of applying \textit{Normalizer} to the data. For the first row $x = [-1.5]$, the Euclidean length is $||x||_{2} = \sqrt{(-1.5)^2} = 1.5$, thus the value -1.5 is transformed to $\frac{-1.5}{1.5}=-1$.

\textbf{5. \textit{PowerTransformer: }} In some modelling scenarios, the normality of the features is desirable. When some features exhibit a large skewness, \textit{PowerTransformer} can perform an exponential and monotonic transformation to each feature to make its distribution more normal-like. It can help ML models with normal-like input assumptions and help to handle highly skewed data~\cite{Sklearn-doc}. The default PowerTransformer of Scikit-Learn, called Yeo-Johnson transformation, is defined as:
\begin{equation}\small
    \label{yeo-johnson}
    Yeo-Johnson(x) = 
        \begin{cases}
        \frac{(x+1)^{\lambda} - 1}{\lambda}& x >= 0, \lambda \neq 0\\
        log(x + 1) & x >= 0, \lambda = 0 \\
        \frac{1 - (1 - x)^{2 - \lambda}}{2 - \lambda} & x < 0, \lambda \neq 2 \\
        -log(1 - x) & x < 0, \lambda = 2
        \end{cases}
    \end{equation}
where $\lambda$ is automatically calculated. It aims to fit the transformed feature into a zero-mean, unit-variance normal distribution as much as possible. Figure \ref{fig:preprocessors_example}(f) shows the result of applying \textit{PowerTransformer} to the original feature. At first, the \textit{PowerTransformer} finds the optimal $\lambda$ automatically, which is 1.22. Then, the transformed results are calculated by Equation~\ref{yeo-johnson}. For example, the transformed result of value -1.5 is $\frac{1-(1-(-1.5))^{2-1.22}}{2-1.22}=-1.34$. 
 
\textbf{6. \textit{QuantileTransformer: }} Similar to the applied scenarios of \textit{PowerTransformer}, \textit{QuantileTransformer} also performs a non-linear transformation. \textit{QuantileTransformer} transforms features independently into a uniform or a normal distribution~\cite{Sklearn-doc}. Since the above \textit{PowerTransformer} can be used for normal distribution, we utilize \textit{QuantileTransformer} for uniform distribution. Intuitively, each transformed value represents its quantile position in the original feature column. For example, Figure~\ref{fig:preprocessors_example}(a) shows the original feature column: [-1.5, 1, 1.5, 2.5, 3, 4, 5]. As shown in Figure~\ref{fig:preprocessors_example}(g), it is transformed to  $[\frac{0}{6}, \frac{1}{6}, \frac{2}{6}, \frac{3}{6}, \frac{4}{6}, \frac{5}{6}, \frac{6}{6}]  = [0, 0.17, 0.33, 0.5, 0.67, 0.83, 1]$.

\textbf{7.\textit{Binarizer: }} Certain datasets can benefit from binarization, especially when binarization results present a high correlation with the target label~\cite{hammer1962adaptive}.  \textit{Binarizer} binarizes data into two values (0 or 1) according to a threshold. Values greater than the threshold are mapped to 1, otherwise to 0~\cite{Sklearn-doc}. The default threshold is 0, which means that negative values are mapped to 0, and non-negative values are mapped to 1. E.g., Figure~\ref{fig:preprocessors_example}(a) is the original feature without any preprocessing. Figure~\ref{fig:preprocessors_example}(h) shows the result of applying \textit{Binarizer} to the feature. With the default threshold 0, the number -1.5 is transformed to 0, while the other values are transformed to 1. 

\stitle{How Do We Choose the Feature Preprocessors?}
 Our work considers a total of seven widely used preprocessors. When selecting the above feature preprocessors, we prioritize two perspectives including effectiveness and interpretability~\cite{Sklearn-doc}. We select feature preprocessors that have shown effectiveness on widely applied models such as LR and XGB. The feature preprocessors should also be easy to interpret, thus users can understand their functionalities easily by combining them with the knowledge of specific datasets and models. Though there indeed exists other kinds of feature prerpecessors under the area of deep feature extraction and deep feature learning, such as x-bit encoding, they are with poor interpretability which is the reason that we do not select them in this work. Moreover, the seven selected preprocessors are more than what the popular AutoML tools support (see Section~7 for more detail), which means that the finding derived from this paper are applicable to a wide range of scenarios. For situations when more preprocessors are needed, one can easily extend our benchmark to derive additional insights.

\subsection{Motivation}
\stitle{Does FP Really Matter?}
To motivate Auto-FP, we conduct an exploratory experiment to examine whether feature preprocessing really matters. Since it is expensive to enumerate all possible pipelines, we only consider pipelines whose length is no larger than 4, leading to a total of 2800 different pipelines. For each pipeline, we apply it to a training set and then train a LR model on the preprocessed dataset. We plot the distribution of model accuracy of 2800 pipelines. Figure~\ref{fig:order_impact} shows the results on four datasets.  The x-axis represents different model accuracy values; the y-axis represents how many pipelines can achieve the accuracy. The red line indicates the ML accuracy without any preprocessing. 
We can see that different pipelines lead to significantly different model accuracy. For example, on the Heart dataset, the model accuracy is spread out from 0.49 to 0.88. 

Furthermore, a good pipeline can significantly improve the model accuracy and a bad pipeline can even hurt the model accuracy. For example, on \textit{Pd} dataset, the model accuracy of the best (worst) pipeline is 0.94 (0.24), while no pipeline (i.e., the red line) can achieve an accuracy of 0.81. These results validate the importance of feature preprocessing. To further motivate Auto-FP, we also examine whether the 2800 pipelines can indicate a better option compared to a given combination of FP. To make the results convincing, for each dataset, we employ FP pipelines generated by a popular AutoML tool TPOT\cite{TPOT} to represent ``a given combination of FP". Table~\ref{tab:acc_tpot_2800} shows the accuracy comparison between the TPOP FP pipeline and the best FP pipeline in the 2800 FP pipelines we consider. For the four datasets, the best FP pipelines in the 2800 FP pipelines win. Thus, extending the length of FP pipelines to obtain better performance is promising, indicating the need for Auto-FP. 

\begin{table*}[t]
\caption{Accuracy comparison between the TPOT FP pipeline and the best pipeline in Figure 2.}
\label{tab:acc_tpot_2800}
\small
\begin{tabular}{|c|c|c|}
\hline
\textbf{Dataset} & \textbf{TPOP FP Pipeline / Accuracy}                                   & \textbf{Best FP Pipeline in Figure 2 / Accuracy}                                                                \\ \hline
Heart            & Binarizer / 0.8333                                                     & Normalizer -\textgreater StandardScaler -\textgreater Binarizer / 0.8367                                        \\ \hline
Forex            & Binarizer -\textgreater StandardScaler / 0.7001                        & MaxAbsScaler -\textgreater Normalizer -\textgreater Normalizer -\textgreater StandardScaler / 0.7042            \\ \hline
Pd               & MinMaxScaler / 0.9250                                                  & StandardScaler -\textgreater Normalizer -\textgreater MinMaxScaler / 0.9421                                     \\ \hline
Wine             & Binarizer -\textgreater Normalizer -\textgreater MaxAbsScaler / 0.5591 & PowerTransformer -\textgreater Normalizer -\textgreater MaxAbsScaler -\textgreater QuantileTransformer / 0.5693 \\ \hline
\end{tabular}
\end{table*}

\stitle{Are there explicit data characteristic rules that can be used to infer the effectiveness of FP?}

To further motivate Auto-FP, we investigate the relationship between data characteristics and the effectiveness of FP.  We consider 40 data characteristics used in Auto-Sklearn~\cite{Auto-Sklearn}, including simple characteristics like \textit{NumberOfClasses} describing basic information, statistical characteristics like \textit{SkewnessMean} measuring normality of columns, landmarking characteristics like \textit{Landmark1NN} depicting data sparsity and separability, and information-theoretic characteristics like \textit{ClassEntropy} calculating class imbalance. The detailed list of these data characteristics is shown in the technical report \cite{Auto-FP}. The datasets, downstream ML models, and the experimental environment we use are the same as in Section 5.

We investigate whether there are \emph{data characteristic rules} that can be used to infer the effectiveness of feature preprocessing. For example, one possible rule is ``\emph{if features are highly skewed, then FP will be very useful for improving the downstream model accuracy}''. We collect 45 commonly used ML datasets with different characteristics and construct training data, where each training example corresponds to one dataset. The training data has 40 data characteristics as features and a binary label which represents whether FP will improve the downstream model accuracy by a relatively large margin. We set the label to 1 (0) if the downstream model accuracy can be improved by more (less) than 1.5\% with 200 randomly selected FP pipelines. 
We consider no-FP as the baseline. After inputting a specified dataset directly into model without any FP, we get an accuracy score A. After applying the 200 randomly selected pipelines on a specified dataset and model, we get an accuracy score B. We calculate B-A. If B-A > 1.5\%, we give label 1 to the specified dataset. If B-A < -1.5\%, we give label 0 to the specified dataset.
Since our goal is to generate data characteristic rules, we train a decision tree on the training data. We vary the tree depth (i.e., the length of a characteristic rule) and report the 3-fold cross-validation (3-CV) scores of decision tree models for different downstream models in Table~\ref{tab:tree_and_cv_scores}. Unfortunately, the 3-CV scores are quite low in all cases. It indicates that there is no data characteristic rule that can confidently infer whether FP works. If a data characteristic becomes a common “rule” like functional dependency, the Decision Tree can figure out easily and the 3-CV score should be approximately 1. However, our paper shows 3-CV score is around 0.67, which means that there are 1/3 wrong improvement predictions wasting users’ time to do unnecessary AutoFP, i.e. search FP pipelines. Thus, this further justifies the need for Auto-FP.

\section{Automated Feature Preprocessing}
In this section, we first formally define the Auto-FP problem and then 
view Auto-FP problem as HPO and NAS. 

\subsection{Problem Formulation} 

\begin{definition}[Feature Preprocessor]

Given a dataset $D$, let $r_i$ be its $i$-th row and $c_j$ be its $j$-th column. A preprocessor $\mathcal{P}$ is a mapping function that maps dataset $D$ to $D'$, where each row $r_i$ is mapped to $r_i'$ and each column $c_j$ is mapped to $c_j'$.
\end{definition}

\begin{example}
This paper considers the seven feature preprocessors from Section~2.1. Given a dataset $D$ and a feature preprocessor $\textit{StandardScaler}(\cdot)$, we have $D' = \textit{StandardScaler}(D)$.
\end{example}

\begin{definition}[Feature Preprocessing Pipeline]

Given a set of preprocessors, a pipeline $\mathcal{L}$ of size $n$ is a composite function that contains a sequence of $n$ preprocessors, denoted by $\mathcal{P}_1 \rightarrow \mathcal{P}_2 \rightarrow \cdots \rightarrow \mathcal{P}_n$. For a dataset $\mathcal{D}$, $\mathcal{L}$ maps it to dataset $\mathcal{D}'$, where $\mathcal{D}' = \mathcal{P}_n \circ (...\mathcal{P}_2 \circ (\mathcal{P}_1(\mathcal{D}))$.
\end{definition}

\begin{example}
Given $\mathcal{D}$ and a feature preprocessing pipeline $PowerTransformer \rightarrow MinmaxScaler \rightarrow Normalizer$, we have $\mathcal{D}' = Normalizer \circ (MinmaxScaler \circ (PowerTransformer(\mathcal{D})))$. 
\end{example}

\stitle{Pipeline Error.} Auto-FP problem is the problem of searching for the best pipeline. To solve this problem, we first need to define what the ``\emph{best}'' means, i.e., how to measure the quality of a given pipeline. An FP pipeline will be used to transform the training data and valid data. The downstream model is trained on the transformed training data and gets a valid error on the transformed valid data. The valid error here is called pipeline error. Thus searching for the best pipeline means searching FP pipeline with minimal pipeline error. We formally define \textit{Pipeline Error} as follows:

Consider the model training process that gives a classifier $\mathcal{C}$, a training dataset $\mathcal{D}_{train}$ with its labels, and a validation dataset $\mathcal{D}_{valid}$ with its labels. For a pipeline $\mathcal{L}$, it will be used to transform the training data $\mathcal{D}_{train}$ and get a new dataset $\mathcal{L}(\mathcal{D}_{train})$. Then, the classifier $\mathcal{C}$ will be trained on the new dataset. We denote the trained classifier as $\mathcal{C}_{\mathcal{L}(\mathcal{D}_{train})}$, or $\mathcal{C}_{\mathcal{L}}$ for simplicity when the context is clear. Clearly, a good pipeline should be the one that minimizes the error of the trained classifier $\mathcal{C}_{\mathcal{L}}$. Since the test data is not available in the training stage, we measure the validation error of $\mathcal{C}_{\mathcal{L}}$ on $\mathcal{D}_{valid}$, denoted by $\validerror(\mathcal{C}_{\mathcal{L}}, \mathcal{D}_{valid})$. Then, we define the error of a pipeline $\mathcal{L}$ as:
\begin{equation}
    \error(\mathcal{L}) := \validerror(\mathcal{C}_{\mathcal{L}(\mathcal{D}_{train})}, \mathcal{D}_{valid})
\end{equation}

Now we have defined the error of a pipeline. Next, we formally define the pipeline search problem (i.e., automated feature preprocessing) in \cref{def:pipe-search}.

\begin{definition}[Automated Feature Preprocessing]
\label{def:pipe-search}
Given a set of preprocessor $\mathcal{S}_{prep}$, suppose that a feature preprocessing pipeline contains at most $N$ preprocessors. Let $\mathcal{S}_{pipe}$ be the set of all pipelines  constructed with $n = \{1, \cdots, N\}$ preprocessors chosen from $\mathcal{S}_{prep}$. The automated feature preprocessing (Auto-FP) problem aims to find the best pipeline with minimal error, i.e.,
\begin{equation*}
    \argmin_{\mathcal{L} \in \mathcal{S}_{pipe} } \error(\mathcal{L})
\end{equation*}
\end{definition}

\subsection{Auto-FP as HPO and NAS}
Interestingly, Auto-FP can be viewed in two different ways. 
\begin{figure}[t]
\vspace{-2em}
\centering
\subfigure{
\label{Fig.sub.4}
\includegraphics[width=0.9\linewidth]{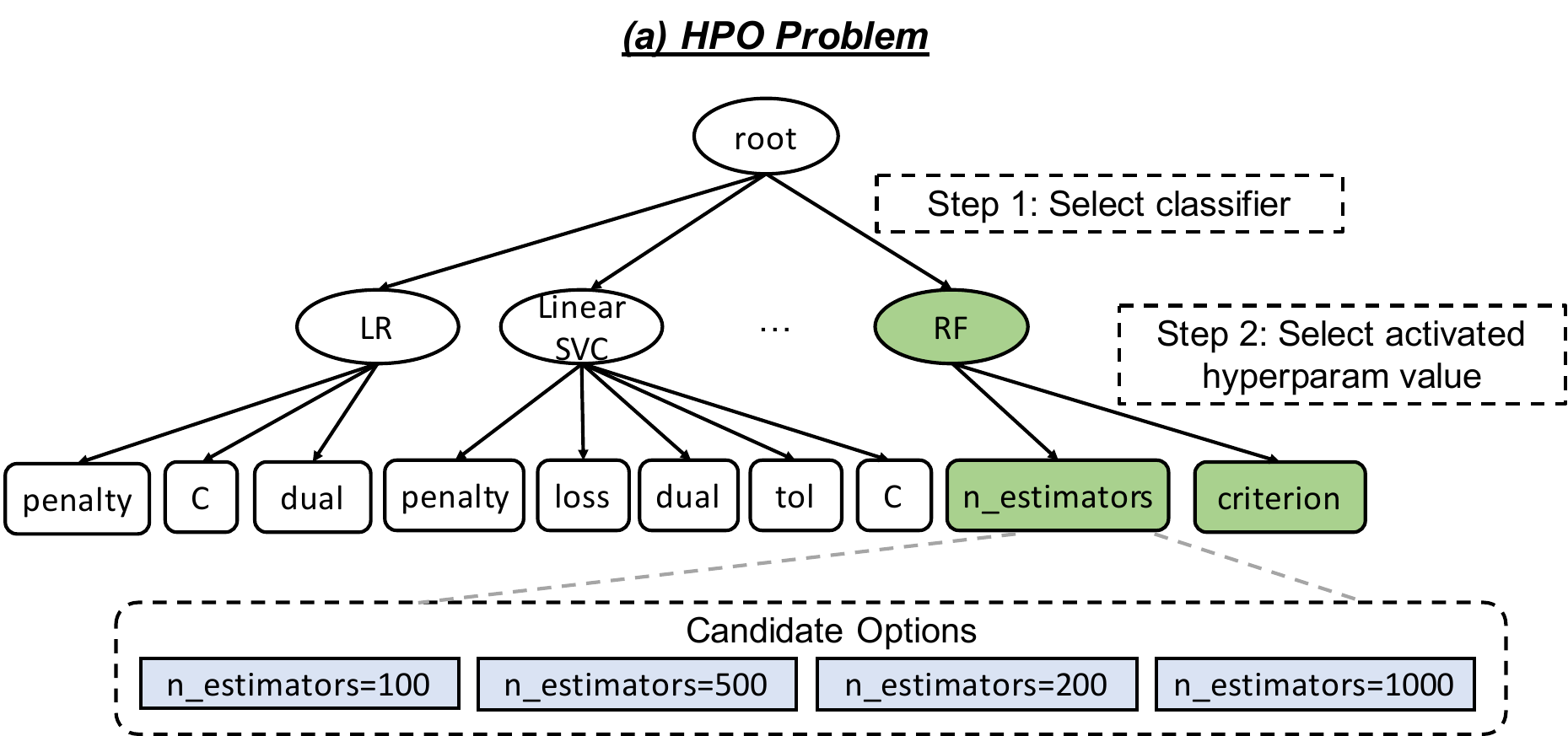}}
\subfigure{
\label{Fig.sub.5}
\includegraphics[width=0.9\linewidth]{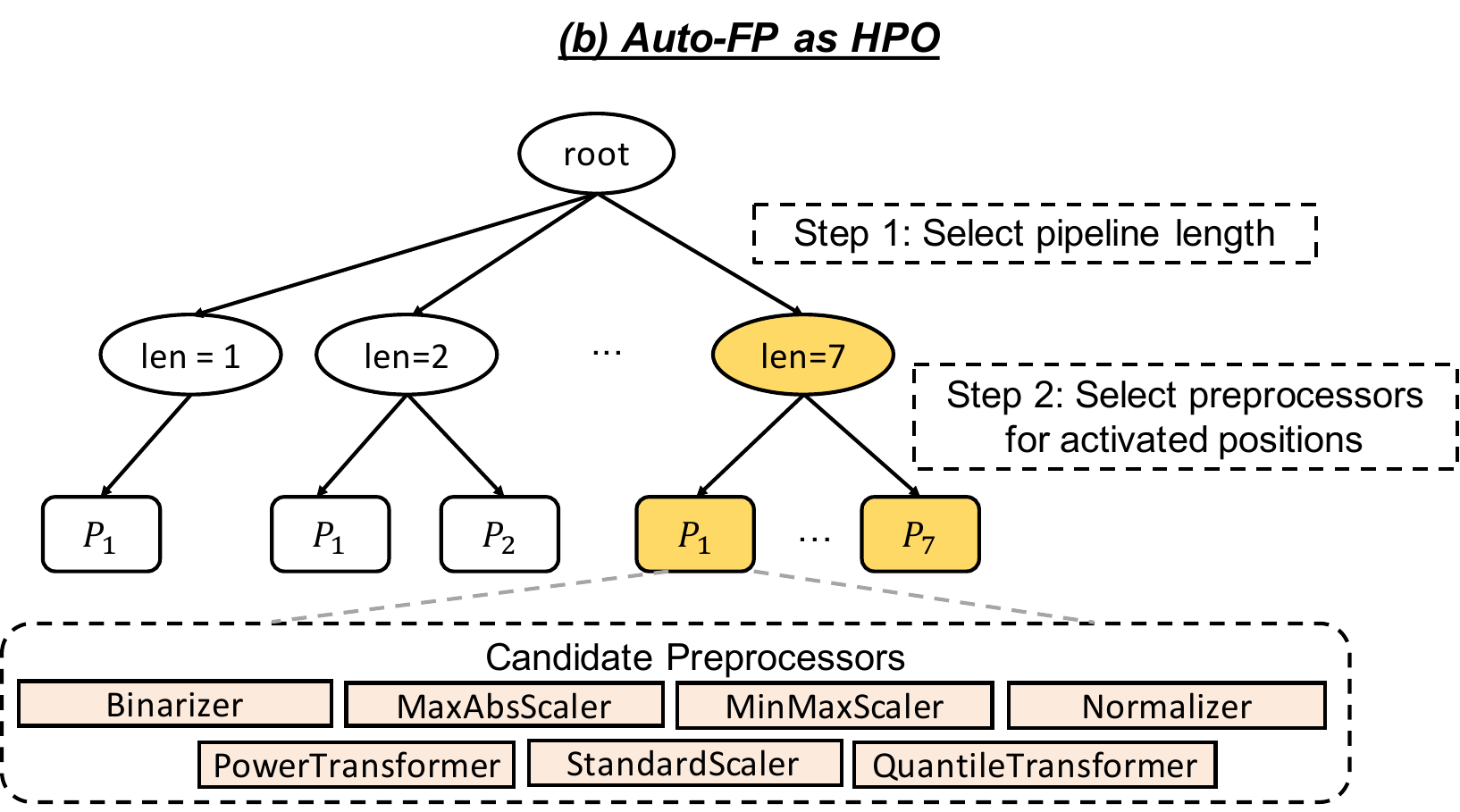}}
\caption{The analogy between HPO and Auto-FP.}
\label{fig:autofp_as_hpo}
\end{figure}

\stitle{Auto-FP as HPO.} HPO aims to find the best combination of classifier and its related hyperparameters for a given dataset. Since different classifiers have different HP space (e.g., LR needs to tune ``penalty'' while Random Forest (RF) does not), its search process contains two steps and can be modelled as a search tree like Figure~\ref{fig:autofp_as_hpo}(a).  The first step is to select a classifier to activate its corresponding HP space (e.g., RF). The second step is to determine the candidate option for each HP (e.g.,  n\_estimators = 200).  The search process of Auto-FP can also be divided into two steps as shown in Figure~\ref{fig:autofp_as_hpo}(b). The first step is to select the pipeline length (e.g., len = 7). The second step is to determine the specific preprocessor for each position (e.g., $P_1$ = Normalizer). 

\begin{figure}[t]

\vspace{-2em}
\centering
\subfigure{
\label{Fig.sub.4}
\includegraphics[width=0.9\linewidth]{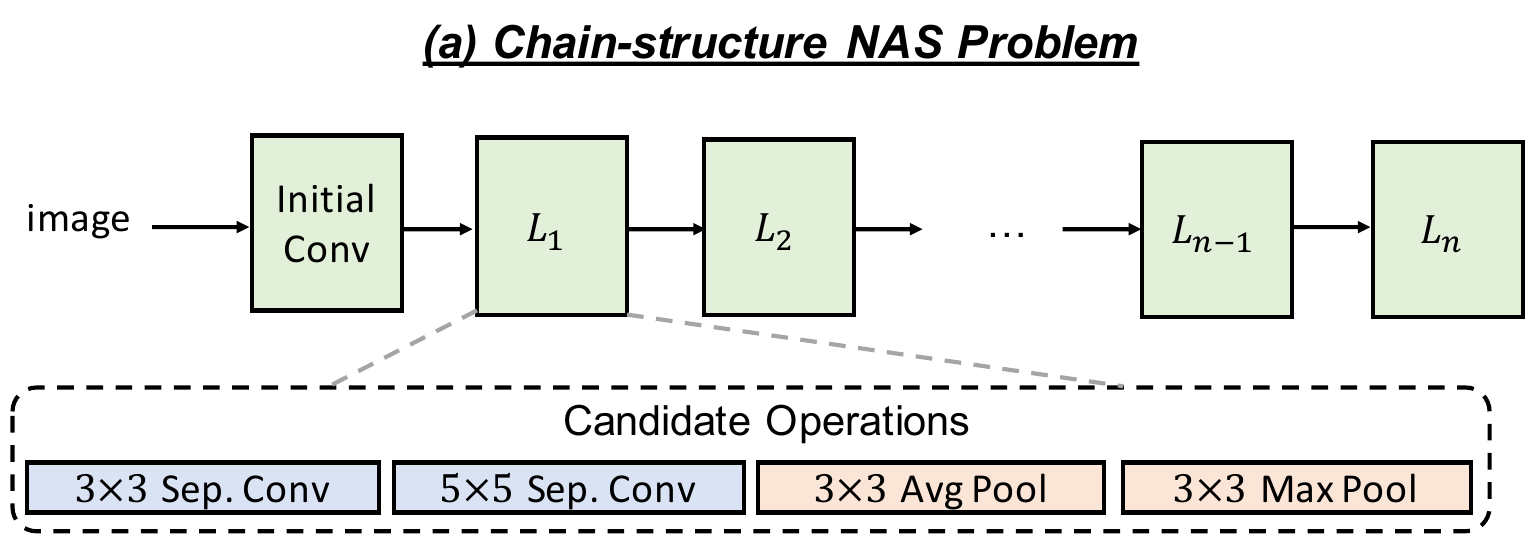}}
\subfigure{
\label{Fig.sub.5}
\includegraphics[width=0.9\linewidth]{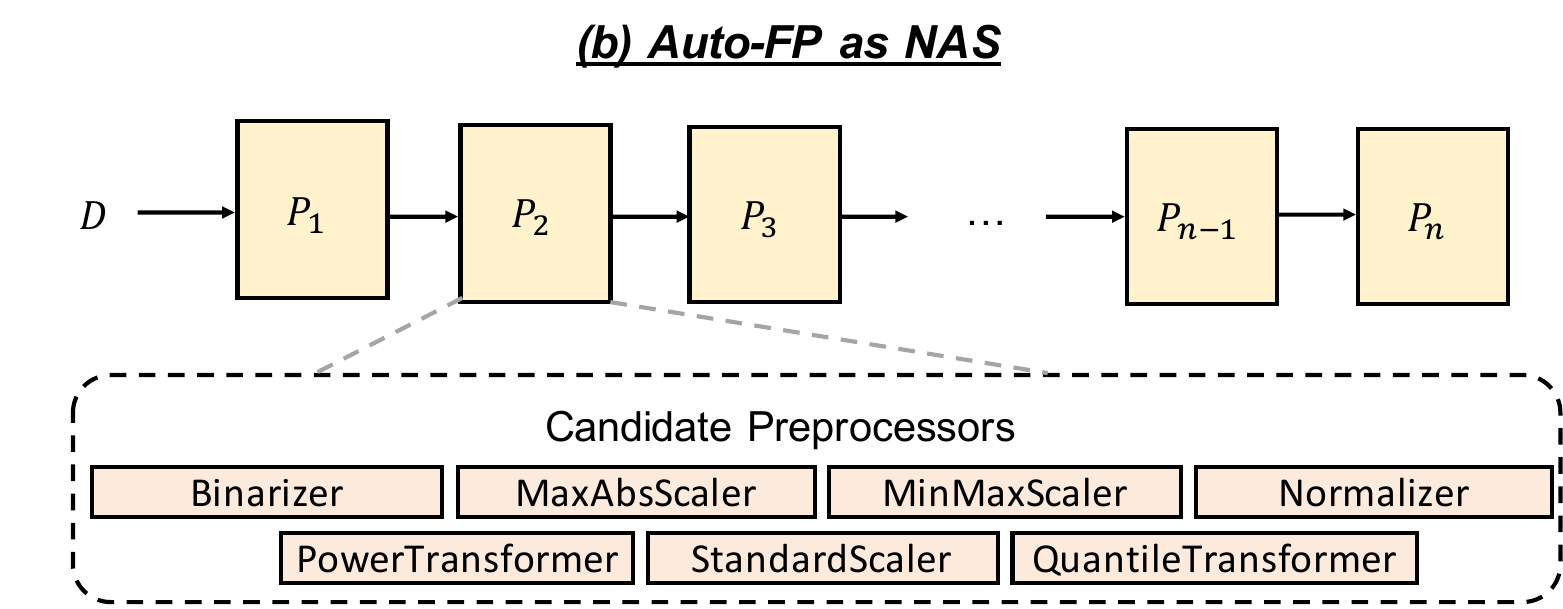}}

\caption{The anology between NAS and Auto-FP.} 
\label{fig:autofp_as_nas}
\end{figure}

\stitle{Auto-FP as NAS.} NAS aims to find the best neural architecture for a given dataset. Auto-FP can be modelled as the chain-structure NAS problem, whose neural architecture has no skip connection among different layers and no multi-branch in each layer. That is, a chain-structure neural architecture is a sequence of operators (i.e., a pipeline) as shown in Figure~\ref{fig:autofp_as_nas}(a).  Its search process optimizes two factors holistically: (1) the max depth of a chain structure; (2) the operator put in each position. Similarly, for Auto-FP, the goal is also to find the best chain-structure neural architecture. The difference is that it puts a feature preprocessor (e.g., Normalizer) rather than a neural-architecture operator at each position.

\noindent\stitle{Remark.} HPO, NAS and AutoFP tasks all aim to find the best combination in a large search space. HPO aims to find the best combination of the classifier and its related hyperparameters for a given dataset. NAS aims to find the best combination of operators from the large neural architecture space. AutoFP aims to search the best combination of feature preprocessors from the large FP pipeline space. Thus, HPO and NAS can cover AutoFP, which is the reason that we derive search algorithms for AutoFP from HPO and NAS area in Section~4.

\section{Auto-FP Search Algoithms}

\begin{table*}[t]
\footnotesize
 \caption{Categories of Automated Feature Preprocessing Search Algorithms.}
  \centering
  \begin{tabular}{|p{0.14\linewidth}<{\centering}
                  |p{0.03\linewidth}<{\centering}
                  |p{0.2\linewidth}<{\centering}
                  |p{0.12\linewidth}<{\centering}
                  |p{0.12\linewidth}<{\centering}
                  |p{0.12\linewidth}<{\centering}
                  |p{0.12\linewidth}<{\centering}|}
    \hline
     \textbf{Category} & \textbf{Area} & \textbf{Search Alg} & \textbf{Surrogate Model} & \textbf{Initialization} & \textbf{\# of samples / iter} & \textbf{\# of evaluations / iter}\\
    \hline
    Traditional & HPO & Random Search (RS) \cite{Random-Search} & None & None & =1 & =1\\
    \hline
    Traditional & HPO & Anneal \cite{Anneal}& None & None & =1 & =1\\
    \hline
    Surrogate-Model-based & HPO & SMAC \cite{SMAC} & Random Forest & Random Search & >1 & =1\\
    \hline
    Surrogate-Model-based & HPO & TPE \cite{TPE}  & KDE & Random Search & >1 & =1\\
    \hline
    Surrogate-Model-based & NAS & \makecell{Progressive NAS + \\MLP no ensemble (PMNE) \cite{PNAS}} & MLP no ensemble & Single Preprocessors & >1 & >1\\
    \hline
    Surrogate-Model-based & NAS & \makecell{Progressive NAS + \\MLP ensemble (PME)  \cite{PNAS}} & MLP ensemble & Single Preprocessors & >1 & >1\\
    \hline
    Surrogate-Model-based & NAS & \makecell{Progressive NAS + \\LSTM no ensemble (PLNE) \cite{PNAS}} & LSTM no ensemble & Single Preprocessors & >1 & >1\\
    \hline
    Surrogate-Model-based & NAS & \makecell{Progressive NAS + \\LSTM ensemble (PLE) \cite{PNAS}} & LSTM ensemble & Single Preprocessors & >1 & >1\\
    \hline
    Evolution-based & HPO & PBT \cite{PBT} & None & Random Search & >1 & >1\\
    \hline
    Evolution-based & NAS & \makecell{Tournament Evolution-\\Higher (TEVO\_H) \cite{TEVO}}& None & Random Search & =1 & =1\\
    \hline
    Evolution-based & NAS & \makecell{Tournament Evolution-\\Younger (TEVO\_Y) \cite{TEVO}}& None & Random Search & =1 & =1\\
    \hline
    RL-based & HPO & REINFORCE \cite{REINFORCE}& Parameter Matrix & None & =1 & =1\\ 
    \hline
    RL-based & NAS & ENAS \cite{ENAS} & LSTM & None & =1 & =1\\
    \hline
    Bandit-based & HPO & Hyperband \cite{Hyperband} & None & None & >1 & >1\\
    \hline
    Bandit-based & HPO & BOHB \cite{BOHB} & KDE & Random Search & >1 & >1\\
    \hline
  \end{tabular}
  \label{tab:summary}
\end{table*}

We identify 15 representative search algorithms from HPO and NAS for Auto-FP. To select the algorithms in our study, we referred to the popular Microsoft NNI tool\cite{Microsoft_Neural_Network_Intelligence_2021}, which has over 12.6k stars on Github, and two widely cited survey papers on AutoML\cite{AutoML-survey} and Neural Architecture Search\cite{NAS-survey}. Our paper covers a significant proportion of HPO and NAS algorithms in NNI, i.e. 8 out of 12 HPO algorithms and 5 out of 10 NAS algorithms. Also, our search algorithms cover all categories of search algorithms in NNI. We initially divide them into 5 categories according to their optimizing strategies. Then we conclude them into a unified framework for analyzing their performance bottleneck in Section~\ref{sec:rank_exp}.  Table~\ref{tab:summary} shows a summary of these algorithms.

\subsection{Categories of Search Algorithms}
Roughly, Auto-FP search algorithms can be divided into 5 categories including traditional, surrogate-model-based, evolution-based, RL-based and band-based algorithms. We introduce these algorithms in detail in the following content.

\subsubsection{\underline{Traditional Algorithms}}
Traditional algorithms samples and evaluates one pipeline for each iteration without any initialization. 

\stitle{Random Search~\cite{Random-Search}} randomly picks one FP pipeline from the search space and evaluates the pipeline with downstream ML model in each iteration.

\stitle{Anneal~\cite{Anneal}} progressively approaches the best FP solution by comparing the current best pipeline to its neighbourhoods. In each iteration, it accepts the better neighbourhoods as the new best state and rejects the worse neighbourhoods. 

\subsubsection{\underline{Surrogate-model-based Algorithms}}
The existing surrogate-model-based algorithms utilize one surrogate model to model the relationship $p(e_x|x)$ between FP pipelines and the downstream model accuracy. The $x$ represents FP pipelines and $e_x$ represents the downstream model accuracy. In each iteration, they follow two steps: 1) fit the surrogate model with evaluated trials 2) generate the next promising pipeline to evaluate according to the new surrogate model.

\stitle{SMAC~\cite{SMAC}} uses the random forest as the surrogate model. With random forest, \textit{SMAC} can handle high-dimensional and categorical input, which fits better for our Auto-FP scenario than Gaussian Process. At each iteration, \textit{SMAC} fits $p(e_x|x)$ with random forest and historical FP pipeline evaluations. Then \textit{SMAC} generates the next promising FP pipeline for evaluation with the newly fitted random forest.

\stitle{TPE~\cite{TPE}} takes \textit{Kernel Density Estimation} (KDE) as the surrogate model which gives the linear time complexity. It performs similar actions like \textit{SMAC} at each iteration, i.e. refitting KDE and generating the next promising FP pipeline.

\stitle{Progressive NAS~\cite{PNAS}} initially starts by considering single preprocessors as pipelines, evaluating them and training the surrogate model (\textit{MLP} or \textit{LSTM}). Then it expands the simple pipelines by adding more possible preprocessors. The surrogate model is used to select the next top-k pipelines for evaluation instead of one single pipeline generated by \textit{SMAC} and \textit{TPE} based on their score prediction. There are four variants of \textit{Progressive NAS} according to the variants of the surrogate model, which are all listed in Table~\ref{tab:summary}. 

\subsubsection{\underline{Evolution-based Algorithms}}
Evolution-based algorithms consider each individual FP pipeline as single DNA in a population. In each evolution step, some outstanding pipelines are selected, mutated and evaluated to update the population.

\stitle{Tournament Evolution~\cite{TEVO}} randomly chooses $S$ FP pipelines from the population in each evolution step and the pipelines with the highest downstream model accuracy are used for mutation. There are two variants of \textit{Tournament Evolution} because of two killing strategies, i.e. kill the oldest pipeline in the population (\textit{TEVO\_Y}) or kill pipeline with worst downstream model accuracy (\textit{TEVO\_H}).

\stitle{PBT~\cite{PBT}} updates the population gradually by replacing bad FP pipelines with the mutation of good pipelines. The ``good pipelines'' means the downstream model accuracies of these pipelines exceed the lowest bar. In each evolution step, \textit{PBT} also injects more exploration by randomly generating FP pipelines with a fixed probability instead of just mutating.

\subsubsection{\underline{RL-based Algorithms}} The existing RL-based algorithms aims find an optimal policy $\pi_{\theta}$ which can maximize the expected cumulative reward. With the trial-and-error strategy, FP pipelines are sampled and evaluated to produce rewards.

\stitle{REINFORCE~\cite{REINFORCE}} is a very traditional policy-gradient algorithm based on the Monte Carlo strategy, which tries to directly update the parameter $\theta$ of a policy. In each iteration, it samples one FP pipeline and updates $\theta$ according to the downstream model accuracy. The larger the downstream model accuracy of the sampled pipeline, the higher probability to choose feature preprocessors in this pipeline.


\stitle{ENAS~\cite{ENAS}} considers all the FP pipeline architectures as DAGs and considers the whole search space as a large super-graph. An FP pipeline architecture here is the chain structure connecting feature preprocessors mentioned in Definition 2 and Example 3.2, which is naturally a DAG structure. The nodes in this super-graph represent the preprocessors, and the edges represent the edge flow. Each iteration utilizes a LSTM to predict the next edges that should be activated and the next preprocessors that should be used. If the next edge is activated, it means that the length of FP pipeline is increased by 1 and the next preprocessor needed to be indicated. 

\subsubsection{\underline{Bandit-based Algorithms}} Bandit-based algorithms aim to create a trade-off between the number of evaluated FP pipelines and the evaluation time for each pipeline. It allocates more time for promising pipelines and successively discards others before the evaluation process is finished. Note that there are many types of bandit-base algorithms such as Thompson Sampling\cite{thompsonSampling} and Upper Confidence Bound\cite{UCB}. However, Thompson Sampling and UCB are both used to solve multi-armed bandit problems, while Hyperband\cite{Hyperband}(1952 citations) and BOHB\cite{BOHB}(839 citations) are two very popular HPO algorithms used specifically for HPO. Due to we take the analogy between Auto-FP and HPO, we select Hyperband and BOHB as the Bandit-based search algorithms in this work.

\stitle{Hyperband~\cite{Hyperband}} considers \textit{Succesive Halving}~\cite{successive-halving} as a subroutine and each running of SH is called a bracket. In each bracket, the outer loop controls the number of randomly sampled FP pipelines and the initial resource allocation, while the inner loop runs SH.

\stitle{BOHB~\cite{BOHB}} indicates the shortcoming of pure \textit{Hyperband}, that is, randomly generating sampled FP pipelines in each bracket wastes the limited budgets. Thus, it gives a mixture of randomly selected pipelines and pipelines generated by \textit{TPE}, which helps to direct pipeline search without losing exploration. 

\subsection{Search Algorithm Framework}

In fact, we notice that all these algorithms roughly follow the same search framework, as shown in Algorithm~\ref{alg:framework}. It is an iterative framework mainly consisting of four steps: \emph{Step 1}. Generate initial pipelines; \emph{Step 2}. Update a surrogate model (optional); \emph{Step 3}. Sample new pipelines; \emph{Step 4}. Evaluate sampled pipelines and go back to Step 2 until the budget is exhausted. Finally, the pipeline with the lowest error is returned. In the following, we describe how each algorithm works at each step in detail. 

\begin{algorithm}[t]
	\renewcommand{\algorithmicrequire}{\textbf{Input: minimal budget $b_min$, maximum budget $b_max$, discarding proportion $\eta$ (default $\eta = 3$)}}
	\caption{\textbf{: A Unified Auto-FP Search Framework}}
	\label{alg:framework}
	\begin{algorithmic}[1]
	    \STATE \textbf{Input:} dataset $D = (D_{train}, D_{valid})$, time budget $T$, surrogate model $\mathcal{M}$ (optional), downstream ML model $\mathcal{C}$
		\STATE \textbf{Initialization}: Randomly sample and evaluate $n_{init}$ pipelines $\mathcal{P}_{init} = \{(\mathcal{L}_{1}, eval_{1}), (\mathcal{L}_{2}, eval_{2}), \cdots, (\mathcal{L}_{n_{init}}, eval_{n_{init}})\}$ with random search. \ \ \ \ \ \ \ \ \ \ \ \ \ \ \ \ \ \ \ \ \ \ \ \ \ \ \ \ \ \ \ \ \ \ \ // \texttt{Step 1}
		
		\STATE elapsedTime = 0, $\mathcal{P}_{new} = \emptyset$, $\mathcal{P}_{eval} = \mathcal{P}_{init}$
		\WHILE{elapsedTime < $T$}
		\STATE $\mathcal{M} = \textsf{Update}(\mathcal{M}, \mathcal{P}_{eval})$ \ \ \ \ \  \  \ \ \ \ \ \ \ \ \ \ // \texttt{Step 2}
		\STATE $\mathcal{S}_{new}$ =  \textsf{get\_sampled\_pipelines}() \ \ \ // \texttt{Step 3}
		\STATE $\mathcal{P}_{new} = \textsf{Eval}(\mathcal{S}_{new})$ \ \ \ \ \ \ \ \ \ \ \ \ \ \ \ \ \ \ \ \ \ \ \ // \texttt{Step 4}
		\STATE $\mathcal{P}_{eval} = \mathcal{P}_{eval} \cup \mathcal{P}_{new}$
		\ENDWHILE
		\RETURN \textit{Pipeline with the lowest error from $\mathcal{P}_{eval}$}
	\end{algorithmic}  
\end{algorithm}

\stitle{Step 1: Generate initial pipelines} (Line 2): As shown in the ``Initialization'' column of Table~\ref{tab:summary}, most algorithms need an initialization step, i.e., generating initial pipelines. The evolution-based algorithms randomly generate initial pipelines to form an initial population. \textit{RS}, \textit{Anneal}, and \textit{Hyperband} are the ones that do not need any initial pipelines. However, they still follow the framework in  Algorithm \ref{alg:framework} by setting $\mathcal{P}_{init} = \emptyset$. RL-based algorithms do not need any initial pipelines because \textit{REINFORCE} uses a randomly generated parameter matrix as the initial policy and \textit{ENAS} utilizes a randomly parameterized LSTM as the initial controller. Surrogate-model-based algorithms all need the initial pipelines and their evaluation results to build an initial surrogate model. For example, \textit{TPE} leverages initial pipelines to generate an initial KDE, while \textit{SMAC} uses them to generate an initial random forest model. \textit{Progressive NAS} build initial \textit{MLP} or \textit{LSTM} with all single-preprocessor pipelines. Note that the initialization of all these algorithms except for \textit{Progressive NAS} uses random search.

\stitle{Step 2: Update a surrogate model (optional)} (Line 5):
Surrogate-model-based algorithms leverage surrogate models to generate pipelines. As mentioned in Step 1, all of them need initial pipelines to initialize a surrogate model. After each iteration, there are one or more newly sampled pipelines evaluated. The surrogate model should be updated with the historical and newly generated pipelines. For instance, \textit{SMAC} retrains its random forest, \textit{TPE} refits its KDEs, and \textit{Progressive NAS} refreshes its MLP or LSTM surrogate model. Other algorithms which include surrogate models also need to update their surrogate model: \textit{BOHB} refits its KDE with pipelines trained with the highest iterations or estimators,
\textit{REINFORCE} updates its policy, and \textit{ENAS} updates its LSTM model. Note that the algorithms without any surrogate model skip this step and directly go to Step 3. 

\begin{figure*}[t]
\centering
\subfigure{
\label{Fig.sub.4}
\includegraphics[width=0.22\linewidth]{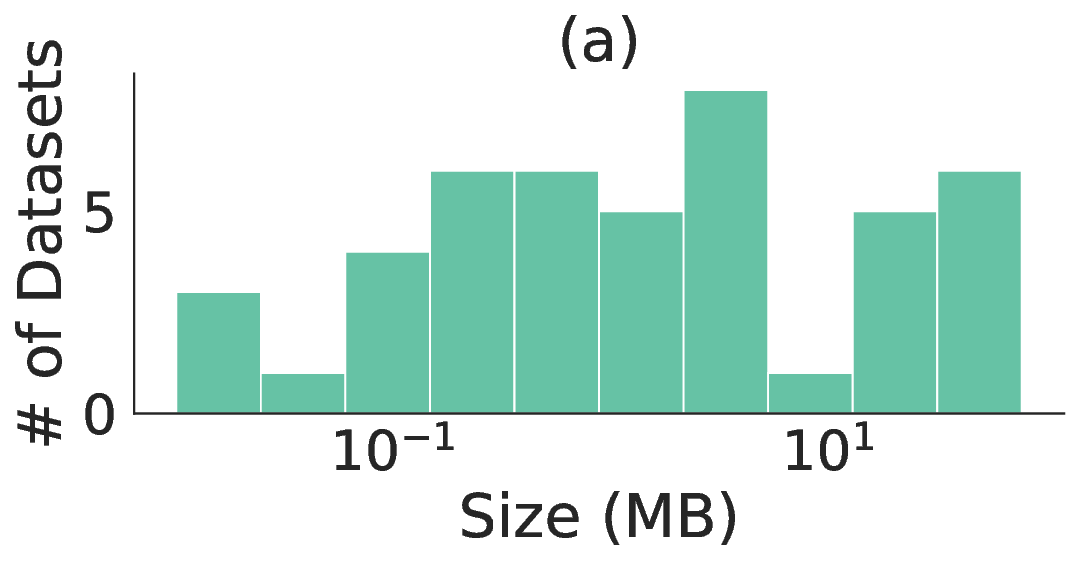}}
\subfigure{
\label{Fig.sub.5}
\includegraphics[width=0.22\linewidth]{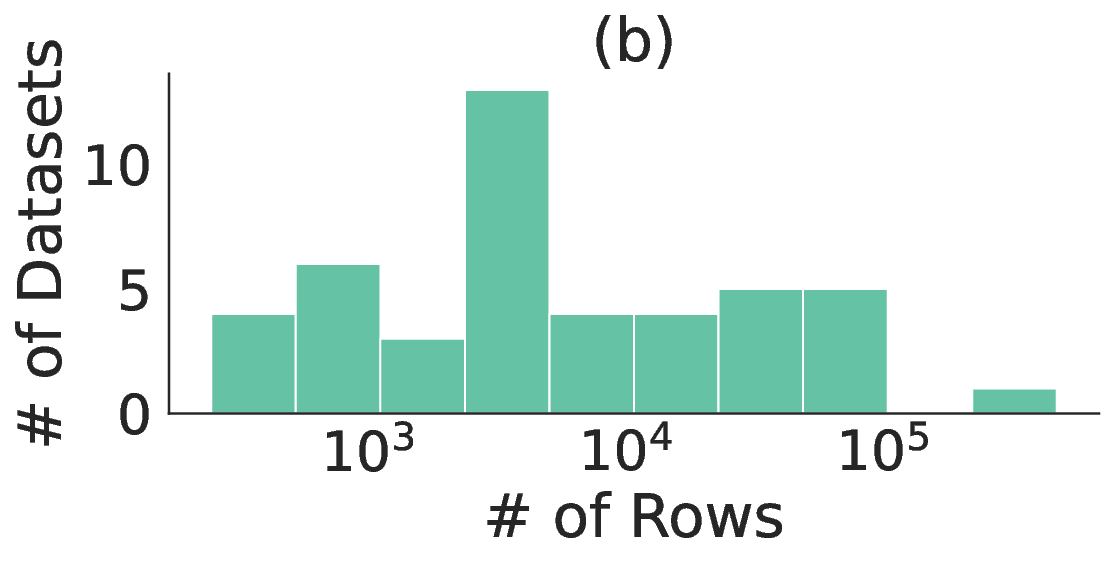}}
\subfigure{
\label{Fig.sub.8}
\includegraphics[width=0.22\linewidth]{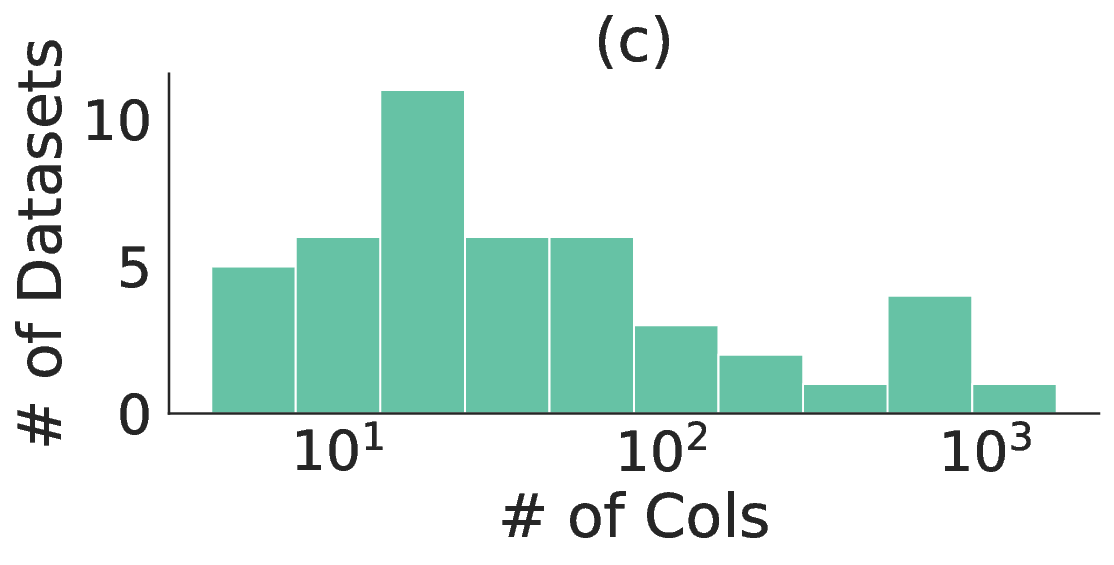}}
\subfigure{
\label{Fig.sub.8}
\includegraphics[width=0.22\linewidth]{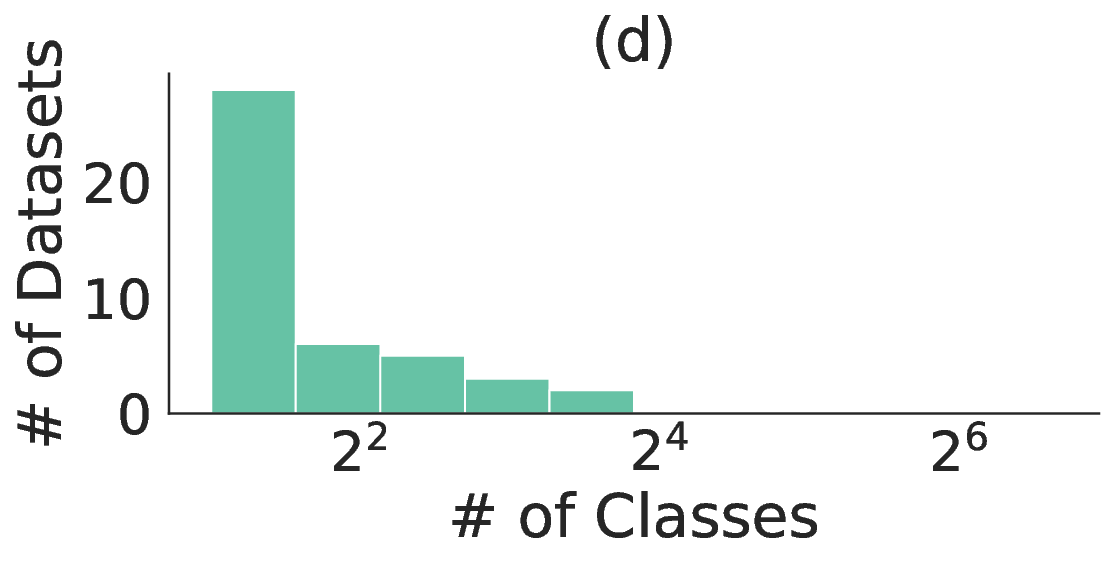}}
\caption{Statistics of 45 real-world ML datasets.}
\label{fig:dataset_stat}
\end{figure*}

\stitle{Step 3: Sample new pipelines} (Line 6):
Different algorithms sample new pipelines with different strategies. In addition, they could sample one or multiple pipelines.  \textit{RS} and \textit{Anneal} randomly sample a single pipeline at each iteration. \textit{Hyperband} randomly generates multiple pipelines in order to apply \textit{Successive Halving} to early terminate poor-performing pipelines. \textit{Tournament Evolution} tries to mutate the best one into another new child, while \textit{PBT} generates multiple pipelines with its exploitation and exploration process. 

The algorithms with surrogate models generate new sampled pipeline(s) with updated surrogate models. For example, \textit{TPE}, \textit{SMAC} sample a single pipeline with the best acquisition score based on updated KDE or random forest. \textit{BOHB} follows the framework of \textit{Hyperband}, thus at each iteration, it needs to generate multiple pipelines for Successive Halving. \textit{REINFORCE} produces one sampled pipeline with the updated policy. \textit{ENAS}  generates one sampled pipeline for evaluation using its LSTM controller, and \textit{Progressive NAS}  generates its candidate pipelines with the prediction of LSTM or MLP.

Note that there is also one important difference in Step 3 among the surrogate-model-based algorithms, i.e., the number of sampled pipelines for evaluation. 
\textit{Progressive NAS} generates top-k (k > 1) pipelines for evaluation. In the running process of \textit{TPE} and \textit{SMAC}, actually they produce several candidate pipelines when sampling. However, they pick up only one pipeline with the best acquisition score for the next-step evaluation.

\stitle{Step 4: Evaluate sampled pipelines} (Line 7):
The last step is to evaluate newly sampled pipelines. The goal is to update the population 
(like \textit{TEVO\_H}, \textit{TEVO\_Y} and \textit{PBT}) or collect new data to fit a better surrogate model. In this step, the preprocessed training data is used to train downstream ML models. Then, the preprocessed validation data is used to test the accuracy of trained ML models. The higher the accuracy, the better the preprocessing pipeline.

\section{How do different search algorithms perform?}\label{sec:rank_exp}


In practice, when users are looking for a feature preprocessing pipeline with good quality, they often constrain search time. In this section, we investigate how these search methods perform within specified time limits. At first, we compare these algorithms and rank them. Then, we analyze the performance bottleneck of these algorithms and identify the opportunities to further improve them. 


\subsection{Experimental Setup} \label{exp-setup}

\stitle{Datasets.} 
We search for a large collection of real-world datasets (in total 45 datasets) from the widely-known AutoML challenge website~\cite{AutoML-Challenge-Website}, an AutoML Benchmark~\cite{AutoML-Benchmark}, and Kaggle datasets~\cite{Kaggle-Datasets}. Without losing generalization, for categorical and textual features, we need to first transform them into numerical features and then search FP pipelines for numerical features. That is why we focus on the 45 numerical datasets because the conclusion drawn from numerical datasets can be widely used in all scenarios.

As shown in Figure~\ref{fig:dataset_stat}, these selected datasets have diverse characteristics in terms of file size, the number of rows/columns and binary/multi-classification. The file size of these datasets is from 0.01 MB to 75.2 MB. The number of rows of all datasets is from 242 to 464,809. The number of columns of all datasets is from 4 to 1,636. There are 28 binary classification datasets and 17 multi-classification datasets with up to 100 classes. These broad considerations promise the generalization of our experiments. The detailed information of datasets is shown in our technical report \cite{Auto-FP}.

\stitle{Search Algorithms.} As shown in Table \ref{tab:summary}, we consider 15 search algorithms in total. Some of them have been included in famous \texttt{Python} libraries. E.g., \texttt{HyperOpt} \cite{Hyperopt} 
includes \textit{Anneal} \cite{Anneal} and \textit{TPE}, \texttt{SMAC3} \cite{SMAC3} 
includes \textit{SMAC} algorithm, and \texttt{HyperbandSter} \cite{HpBandSter}
includes \textit{Hyperband} and \textit{BOHB}. We slightly change these libraries to make their algorithms support our scenario. For NAS algorithms like \textit{Progressive NAS} and \textit{ENAS}, they are originally implemented in \texttt{PyTorch}. However, they do not fit into our scenario, and thus we re-implement them based on their papers. We also implement \textit{REINFORCE}, \textit{Tournament Evolution}, and \textit{PBT}. Note that all algorithms are implemented in \texttt{Python}, which is fair for comparison.


\stitle{Downstream Classifiers.} We evaluate search algorithms using three downstream classifiers: Logistic Regression (LR), XGBoost (XGB) and Multi-layer Perceptron (MLP). We choose the three ML models based on the recent survey~\cite{Kaggle-2021-survey}. \textit{LR} is a linear model which takes the top popularity of all ML models, \textit{XGB} is a tree-based model which takes the first popularity of complex ML models, and \textit{MLP} is a neural network which gains more popularity recently. Choosing them means that our experimental results can provide insights on a wide range of scenarios. In terms of implementation, we use \textit{LR} (set n\_jobs = 1) and \textit{MLP} in the Scikit-learn library with default parameters and use the \textit{XGB} model (set n\_jobs = 1) in the XGBoost library. 

\stitle{Training and Evaluation.} 
For each dataset, we split it into training and validation with the proportion 80:20. The training set is used to generate trained downstream models after being preprocessed. And the preprocessed validation set is used to evaluate the trained models. Based on evaluations, search algorithms can derive information and choose the proper search direction for the next steps. After an iterative search of preset time limits, search algorithms stop their work and output the feature preprocessing pipeline with the highest validation accuracy.

\stitle{Experimental Environment.} We conduct experiments on a guest virtual machine with 110 vCPUs and 970GB main memory.
The guest virtual machine runs on a Linux Kernel-based Virtual Machine (KVM) enabled server equipped with four Intel Xeon E7-4830 v4 CPUs clocked at 2.0GHz and 1TB main memory.
Each CPU has 14 cores (28 hyperthreads) and 35MB Cache.
All of the experiments are repeated five times and we report the average to avoid the influence of hardware and network.

\subsection{Which Search Algorithm Performs Better?}

\begin{table*}[t]
\caption{Overall Average Performance Ranking of All Search Algorithms.}
\label{tab:overall_performance}
\centering
\footnotesize
\begin{tabular}{c|cc|ccc|cccccc|cc|cc}
\hline
Category            & \multicolumn{2}{c|}{Traditional} & \multicolumn{3}{c|}{Evolution-based} & \multicolumn{6}{c|}{Surrogate-model-based} & \multicolumn{2}{c|}{RL-based} & \multicolumn{2}{c}{Bandit-based} \\ \hline 
Search Alg          & RS            & Anneal           & PBT      & TEVO\_Y     & TEVO\_H     & PMNE  & PME  & PLNE  & PLE  & SMAC  & TPE  & REINFORCE        & ENAS       & HYPERBAND         & BOHB         \\ \hline 
LR Avg Ranking      & 6             & 12               & \textbf{1}        & 2           & 3           & 4     & 5    & 11    & 14   & 7     & 8    & 10               & 15         & 9                 & 13           \\
XGB Avg Ranking     & 6             & 13               & \textbf{1}        & 2           & 3           & 4     & 5    & 11    & 14   & 8     & 7    & 9                & 15         & 10                & 12           \\
MLP Avg Ranking     & 7             & 12               & 3        & 4           & 5           & \textbf{1}     & 2    & 6     & 10   & 8     & 9    & 14               & 15         & 11                & 13           \\ \hline
Overall Avg Ranking & 6             & 12               & \textbf{1}        & 2           & 3           & 4     & 5    & 9     & 13   & 7     & 8    & 11               & 15         & 10                & 14           \\ \hline
\end{tabular}
\end{table*}

We run the 15 algorithms on 45 datasets under 6 different time constraints: 60, 300, 600, 1200, 1800 and 3600 seconds. We choose these settings because it is more practical to mainly concern about the performance of FP with lower resources and leave some resources for tasks like feature generation and selection.
Due to the space constraint, we report the general findings derived from the comprehensive experimental results on all datasets with all time constraints. Detailed experimental results are shown in the technical report~\cite{Auto-FP}.

\stitle{Which search algorithm ranks better?}
To give a recommendation, we compute the average rankings of all 15 algorithms under all scenarios with at least 1.5\% improvement of validation accuracy compared to no-FP (215 scenarios on \textit{LR} + 90 scenarios on \textit{XGB} + 196 scenarios on \textit{MLP} = 501 scenarios). The reason we choose no-FP as a baseline instead of data processed by single preprocessors is that we want to compare the validation accuracy with/without FP instead of comparing the effectiveness between multiple and single preprocessors. Naturally, the scope of FP includes single preprocessors.
The ranking value in each scenario is also according to validation accuracy. If there is a tie, we give the same ranking value. Table~\ref{tab:overall_performance} shows the ranking results. We can see that \textit{PBT} is ranked at the top, followed by the other two evolution-based algorithms. More specifically, when the downstream model is \textit{LR} or \textit{XGB}, evolution-based algorithms are highly recommended; when the downstream model is \textit{MLP}, \textit{PMNE} and \textit{PME} are highly recommended. The overall average ranking of \textit{RS} is 6, which is still a strong baseline. Our following analysis also takes \textit{RS} as a baseline. Note that though the rankings of SMAC and TPE are medians in Table 3, we still take RS as the baseline because it is widely used and the most simple algorithm of the 15 search algorithms, which can facilitate our analysis of the advantages, bottlenecks and drawbacks of the other 14 algorithms.

\stitle{Why evolution-based algorithms outperform \textit{RS}?}
One possible reason that evolution-based algorithms exceed \textit{RS} is that they have more exploitation than \textit{RS}, which can produce promising search directions for the next steps. For example, \textit{TEVO\_H} mutates the best pipeline of sampled pipelines from the population. Moreover, evolution-based algorithms require a small overhead to select the next pipeline since they just sample and then mutate. This allows them to evaluate many more pipelines under the same budget.

\stitle{Why most surrogate-model-based algorithms do not outperform \textit{RS}?}
We observe that most surrogate-model-based algorithms do not outperform \textit{RS}, except \textit{PMNE} and \textit{PME}. 
The goal of utilizing surrogate models is to direct the search direction for the next steps through a model, thus precise directing is important for the performance of these surrogate-model-based algorithms. However, these surrogate-model-based algorithms that need initialization do not have enough data to promise the initial directions or only start with randomness. For example, the starting points of \textit{PLNE} are just the 7 pipelines with only one preprocessor. Furthermore, the fitting process of these surrogate models is time-consuming, which causes fewer pipelines evaluated. For example, \textit{SMAC} needs to train a random forest, \textit{TPE} needs to fit many KDEs, \textit{PLNE} and \textit{PNE} need to fit a LSTM and multi number of LSTMs. 
This insight also inspires us that the search space of feature preprocessing pipelines is hard to learn by general surrogate models like the random forest, KDE and LSTM, and there is still space to improve these surrogate models for Auto-FP scenario especially.
However, the special cases here are \textit{PMNE} and \textit{PME}. Even though they have non-precise starting points like \textit{PLNE} and \textit{PLE}, the overhead of the fitting process of MLP is very small (approximate to \textit{RS} as shown in Figure \ref{fig:overhead_aus_made_lr_mlp}), which leaves more time to train more precise MLP(s) with enough number of pipelines evaluated. 
\stitle{Why RL-based algorithms show poor performance?}
It is obvious that \textit{REINFORCE} and \textit{ENAS} do not perform well. There are two reasons. Firstly, the initialization of \textit{REINFORCE} and \textit{ENAS} is random, which is not so effective for finding a promising direction at the starting stage. Secondly, \textit{REINFORCE} and \textit{ENAS} employ the idea of stochastic gradient descent, which updates the policy after only one evaluation, which means the process of finding a good policy is slow and with lots of iterations. 



\begin{figure*}[t]
\centering
\includegraphics[width=\linewidth]{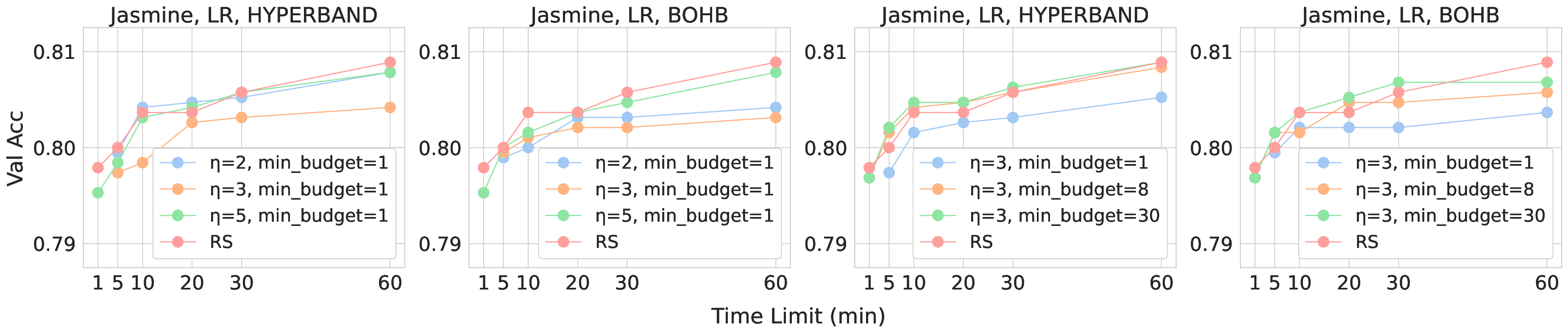}
\caption{Parameter adjustment for \textit{Hyperband} and \textit{BOHB}. The upper two vary $\eta$. The lower two vary \texttt{min\_budget}. Even with several parameter adjustments, \textit{Hyperband} and \textit{BOHB} still cannot outperform \textit{RS}.
}
\label{fig:hyperband_vary_eta_and_budget}
\end{figure*}

\stitle{Why bandit-based algorithms show low average ranking?}
 The average rankings of \textit{Hyperband} and \textit{BOHB} are also behind \textit{RS}. Their main early-stopping idea cannot grasp the correct pipeline ranking at the early stage for downstream ML models under our setting. Even though we try many possibilities of the two important parameters $\eta$ and \texttt{min\_budget} (See Figure~\ref{fig:hyperband_vary_eta_and_budget}. Due to the space limit, we only exhibit the result of \textit{Jasmine} with the \textit{LR} model), it is still hard to make the two algorithms exceed \textit{RS}. Also, it is hard to determine which parameter is better with different downstream ML models and time limits. Clearly, how to improve \textit{Hyperband} and \textit{BOHB}, especially for the Auto-FP scenario, still needs further exploration.
 
\stitle{Are there any frequent excellent feature preprocessor patterns?}
We tried to dig out if there are frequent patterns in the best pipelines of all 45 datasets searched by \textit{PBT} (top 1 ranking search algorithm) with FP-growth~\cite{FPTree} (a famous frequent pattern mining algorithm). However, the support of discovered patterns is very low, i.e. there are no obvious frequent patterns. This result further motivates our search idea and indicates that the Auto-FP problem is hard because of the large search space.

\subsection{Performance Bottleneck Analysis}
\begin{figure*}[t]
\centering
\includegraphics[width=\linewidth]{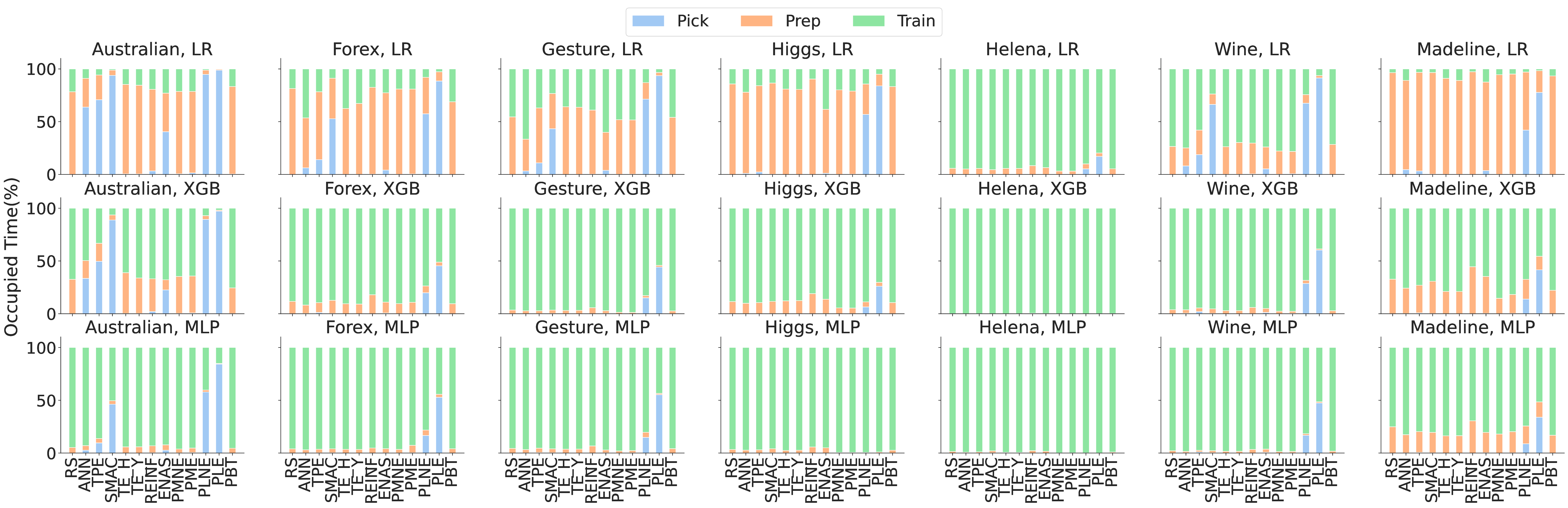}
\caption{Overhead Percentage on 7 datasets with different downstream ML models. ``Pick'' means the overhead of picking up next pipelines. ``Prep'' means the overhead of preprocessing dataset with feature preprocessors. ``Train'' means the overhead of evaluating FP pipelines.}
\label{fig:overhead_aus_made_lr_mlp}
\end{figure*}

We investigate the performance bottleneck of different algorithms to identify opportunities to enhance their performance. We break down the performance into three parts. (1) ``Pick'': picking up FP pipelines to be evaluated, i.e. picking-up time, which includes \texttt{Step 2} and \texttt{Step 3} in Algorithm~\ref{alg:framework}. (2) ``Prep'': preprocessing training and validation datasets with picked pipelines, i.e. preprocessing time, which is included in \texttt{Step 4} of Algorithm~\ref{alg:framework}. (3) ``Train'': training ML models with preprocessed training dataset, i.e., training time, which is also included in \texttt{Step 4} of Algorithm~\ref{alg:framework}. We conduct experiments on all datasets under different time limits. Due to the space constraint, we only show the results of 10 mins on 7 datasets in Figure~\ref{fig:overhead_aus_made_lr_mlp}. Note that \textit{Hyperband} and \textit{BOHB} are not shown because the two methods mix the picking-up and evaluation time and adopt partial training,  making it impossible to record each part separately as other algorithms.

\stitle{What is the most common bottleneck?}
Different search algorithms have different time distributions among the three parts, as shown in Figure~\ref{fig:overhead_aus_made_lr_mlp}. Obviously, ``Train'' is the bottleneck in most cases, followed by ``Prep'', then ``Pick''. ``Train'' and ``Prep'' are highly related to data size, i.e. the smaller data size, the shorter ``Train'' and ``Prep'' time. Therefore, reducing data size (e.g. by sampling) is meaningful for improving the performance. 

\begin{table}[t]
\caption{Performance Bottleneck of Different Scenarios. }
\label{tab:bottleneck_all}
\centering
\footnotesize

\begin{tabular}{|p{0.17\linewidth}<{\centering}
                |p{0.12\linewidth}<{\centering}
                |p{0.1\linewidth}<{\centering}
                |p{0.05\linewidth}<{\centering}
                 p{0.05\linewidth}<{\centering}
                 p{0.1\linewidth}<{\centering}
                 p{0.1\linewidth}<{\centering}|}
\hline
\textbf{Dataset Dimensions} & \textbf{Dataset Size}   & \textbf{ML Model} & \multicolumn{1}{c|}{\textit{\textbf{RS}}} & \multicolumn{1}{c|}{\textit{\textbf{PBT}}} & \multicolumn{1}{c|}{\textit{\textbf{TEVO\_H}}} & \textit{\textbf{TEVO\_Y}} \\ \hline
\multirow{3}{*}{High}       & \multirow{3}{*}{All}    & LR                & \multicolumn{4}{c|}{Prep}                                                                                                                                           \\ \cline{3-7} 
                            &                         & XGB               & \multicolumn{4}{c|}{\multirow{2}{*}{Train}}                                                                                                                         \\ \cline{3-3}
                            &                         & MLP               & \multicolumn{4}{c|}{}                                                                                                                                               \\ \hline
\multirow{9}{*}{Low}        & \multirow{3}{*}{Small}  & LR                & \multicolumn{4}{c|}{Prep/Train}                                                                                                                                     \\ \cline{3-7} 
                            &                         & XGB               & \multicolumn{4}{c|}{\multirow{2}{*}{Train}}                                                                                                                         \\ \cline{3-3}
                            &                         & MLP               & \multicolumn{4}{c|}{}                                                                                                                                               \\ \cline{2-7} 
                            & \multirow{3}{*}{Medium} & LR                & \multicolumn{4}{c|}{Prep}                                                                                                                                           \\ \cline{3-7} 
                            &                         & XGB               & \multicolumn{4}{c|}{\multirow{2}{*}{Train}}                                                                                                                         \\ \cline{3-3}
                            &                         & MLP               & \multicolumn{4}{c|}{}                                                                                                                                               \\ \cline{2-7} 
                            & \multirow{3}{*}{Large}  & LR                & \multicolumn{4}{c|}{Prep/Train}                                                                                                                                     \\ \cline{3-7} 
                            &                         & XGB               & \multicolumn{4}{c|}{\multirow{2}{*}{Train}}                                                                                                                         \\ \cline{3-3}
                            &                         & MLP               & \multicolumn{4}{c|}{}                                                                                                                                               \\ \hline
\end{tabular}
\end{table}

\stitle{Are there explicit data characteristic rules that can be used to figure out bottlenecks?}
To further indicate bottlenecks under different scenarios, we try to conclude the relationship between data characteristics and the type of bottleneck. According to the number of dataset dimensions, we split the 45 datasets into \textit{high-dimensional datasets} (\# of dimensions > 100) and \textit{low-dimensional datasets} (\# of dimensions <= 100). The \textit{low-dimensional datasets} can also be split into three groups according to their size: small (size <= 1.6MB), medium (1.6MB < size <= 4MB) and large (size > 4MB). Combined with the complexity of the downstream ML model, we draw Table~\ref{tab:bottleneck_all} to help researchers develop optimized searched algorithms tailored for different scenarios. For example, if some researchers tend to utilize \textit{PBT} to search FP pipeline for high-dimensional dataset with \textit{XGB} model, it is better for her to enhance the performance by solving the ``Train'' bottleneck.

\subsection{Main Findings}
Our main findings of this section are summarized as follows:
\begin{itemize}[leftmargin=*]
    \item Evolution-based algorithms, especially \textit{PBT}, give the highest overall average ranking under all scenarios. They have more exploitation with a similar small overhead for picking up the next pipeline compared to \textit{RS}.
    \item \textit{RS} is still a strong baseline.
    \item Most surrogate-model-based algorithms except \textit{PMNE} and \textit{PME} do not outperform \textit{RS} because of the imprecise initialization and time-consuming surrogate model fitting process. 
    \item RL-based algorithms do not exceed \textit{RS} because of the imprecise initialization and time-consuming policy learning process.
    \item Bandit-based algorithms do not exceed \textit{RS} because the early-stopping idea cuts-off good pipelines in the very early stage of training when the downstream ML models are \textit{LR, XGB} and \textit{MLP}.
    \item There is no obvious frequent feature preprocessor pattern always performing well.
    \item Different scenarios have different bottlenecks and ``Train'' is the bottleneck in most cases. Users can check Table~\ref{tab:bottleneck_all} for indicating the most promising direction to enhance performance. 
\end{itemize}
\section{Extending Auto-FP to Support Parameter Search}
So far we have assumed that each preprocessor adopts default parameter values. It is not uncommon that users may want to explore other parameter values. 
In this section, we explore two extended search spaces and evaluate two approaches to extend Auto-FP to support parameter search. Furthermore, we also discuss in which situation one is superior to the other and explain the reasons.

\subsection{Two Extended Search Spaces}
We extend our search space by allowing preprocessors' parameters to have multiple possible values. 
For example, \textit{Binarizer} has a parameter called ``threshold''. As mentioned in Section~\ref{operator}, its default value is 0. The extended search space extends the parameter space to a set of values such as \{0, 0.2, 0.4, 0.6, 0.8, 1.0\}. Based on the number of available values, i.e. the cardinality for each parameter, we present two types of extended search spaces: low-cardinality and high-cardinality search space.

\stitle{Low-Cardinality Search Space.}
We first construct an extended search space as shown in Table~\ref{tab:extended_space}, which is the low-cardinality search space. In low-cardinality search space, the value of max cardinality is small. For example, the max cardinality in Table~\ref{tab:extended_space} is the cardinality of \texttt{n\_quantiles}, which is 8.

\stitle{High-Cardinality Search Space.}
We construct another extended search space by significantly increasing the number of possible values for some parameters. We changed the \texttt{threshold} space of \textit{Binarizer} into a list from 0 to 1 with a gap of 0.05, and the \texttt{n\_quantiles} space of \textit{QuantileTransformer} into a list from 10 to 2000 with a gap of 1. As shown in Table~\ref{tab:extended_space_imbalanced}, in the high-cardinality search space, there are parameters with very high cardinality, i.e. the value of max cardinality is large. For example, the max cardinality in Table~\ref{tab:extended_space_imbalanced} is the cardinality of \texttt{n\_quantiles}, which is 1990.


\subsection{Two Approaches to Support Parameter Search}
We adopt two approaches to extend Auto-FP to support parameter search. The first one, called \textit{One-step}, combines parameter search and pipeline search together. The second one, called \textit{Two-step}, treats the parameter search and the pipeline search separately. 

\stitle{One-step.} This approach considers each preprocessor with different selected parameters as different preprocessors. For example, \textit{Binarizer(\texttt{threshold=0})} and \textit{Binarizer(\texttt{threshold=1})} are considered two different preprocessors. In this way, for the low-cardinality search space, the number of preprocessors is even and will be increased from 7 to 6 + 1 + 1 + 3 + 2 + 2 + 16 = 31. After that, any search algorithm presented before can be directly applied to search for the best sequence of preprocessors along with associated parameter values.

\stitle{Two-step.} This approach consists of two steps. In the first step, it randomly selects the parameter values for each preprocessor. For example, \texttt{threshold=1} is selected for \textit{Binarizer} and \texttt{ with\_mean=False} is selected for \textit{StandardScaler}. In the second step, it runs a search algorithm with a short time limit (like 60s) to search for the best pipeline w.r.t. the selected parameter values. It repeats the two steps until the time budget is exhausted and finally returns the best overall pipeline. Note that other sample techniques in the first step besides random can be further explored, but it is outside the scope of this work.

\begin{table}[t]
  \caption{Extended Low-Cardinality Search Space. The max cardinality is the cardinality of 
  \texttt{n\_quantiles}, which is 8.}
  \label{tab:extended_space}
  \centering
  \footnotesize
  \begin{tabular}{|c|c|}
    \hline
    {\bf Preprocessor} & {\bf Space Configures} \\
    \hline
    Binarizer & threshold = [0,0.2,0.4,0.6,0.8,1.0] \\
    \hline
    MinMaxScaler & \makecell{range\_min=0 \\ range\_max = 1} \\
    \hline
    MaxAbsScaler & No parameter \\
    \hline
    Normalizer & norm = [`l1', `l2', `max'] \\
    \hline
    StandardScaler & with\_mean = [True, False] \\
    \hline
    PowerTransformer & standardize = [True, False] \\
    \hline
    QuantileTransformer & \makecell{n\_quantiles = [10, 100, 200, 500, 1000, 1200, 1500, 2000] \\ output\_distribution = [`uniform', `normal']} \\
    \hline
  \end{tabular}
\end{table}




\begin{figure*}[t]
\centering
\includegraphics[width=\linewidth]{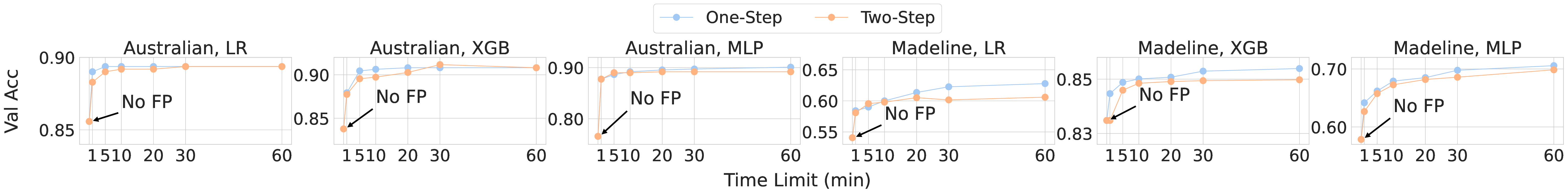}
\caption{Comparison of One-step and Two-step in the extended low-cardinality search space in Table~\ref{tab:extended_space}. One-step is preferred.}
\label{fig:max_scores_by_time_extended_space}
\end{figure*}

\subsection{One-step vs. Two-step.} 
We compare the two approaches on all datasets. We choose the \textit{PBT} as the search algorithm because it shows the best overall average ranking.  

\stitle{For low-cardinality search space, One-step or Two-step?}
We first varied the time limit and used each approach to search for the best pipeline for the extended low-cardinality search space in Table~\ref{tab:extended_space}. Due to the space constraint, Figure~\ref{fig:max_scores_by_time_extended_space} shows the results on \textit{Australian} and \textit{Madeline}. The complete results are shown in our technical report \cite{Auto-FP}. The ``Val Acc'' refers to the average accuracy over five runs. We can see that for most cases, \textit{One-step} outperforms \textit{Two-step}. This is because \textit{Two-step} only exploits at most one group of parameter values every minute. Even if we increase the time limit to 1 hour, it only goes through at most 60 groups of parameter values, which does not do enough exploration. In comparison,  \textit{One-step} can pick up more reasonable pipelines with more exploration. 

\begin{figure*}[t]
\centering
\includegraphics[width=\linewidth]{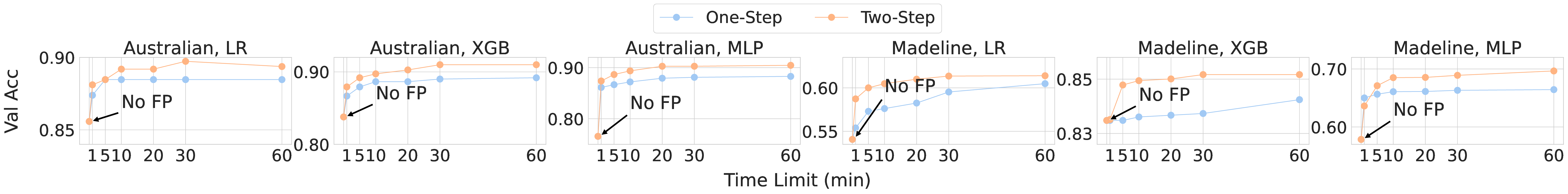}
\caption{Comparison of One-step and Two-step in the extended high-cardinality search space in Table~\ref{tab:extended_space_imbalanced}. Two-step is preferred.}
\label{fig:max_scores_by_time_extended_space_imbalanced}
\end{figure*}

\stitle{For high-cardinality search space, One-step or Two-step?}
However, \textit{One-step} has its own limitation. 
Compared to the low-cardinality search space in Table~\ref{tab:extended_space}, the \texttt{n\_quantiles} parameter of \textit{QuantileTransformer} in the high-cardinality search space has about 2k possible values, which leads \textit{One-step} to choose \textit{QuantileTransformer} with a much higher opportunity. So does the \texttt{threshold} parameter of \textit{Binarizer}. We run the same experiments as in Figure~\ref{fig:max_scores_by_time_extended_space} w.r.t. the high-cardinality search space and get the results in Figure~\ref{fig:max_scores_by_time_extended_space_imbalanced}.
We can see that in most cases, \textit{One-step} performs worse than \textit{Two-step} because \textit{One-step} selects pipelines with many duplicate preprocessors. For example, it may return a pipeline like ``QuanitleTransformer(n\_quantiles=10) $\rightarrow$ QuantileTransformer(n\_quantiles=50) $\rightarrow$ QuantileTransformer(n\_quantiles=200)''. It is natural because \textit{QuantileTransformer} takes a large proportion of preprocessors in this space, which is $4000/4027 \approx 99.3\%$, making the search algorithm less likely to select other preprocessors. \textit{Two-step} can avoid this issue and thus perform better. 

\fussy

In summary, our experimental study shows that different types of extended search space fit different approaches. In low-cardinality search space, \textit{One-step} is a preferred approach. However, in high-cardinality settings (e.g., one preprocessor dominates the search space), it may perform worse than \textit{Two-step}. It is an open research problem to combine pipeline search and parameter search in a systematic way.

\begin{table}[t]
  \caption{Extended High-Cardinality Search Space. The max cardinality is the cardinality of
  \texttt{n\_quantiles}, which is 1990.}
  \label{tab:extended_space_imbalanced}
  \centering
  \footnotesize

  \begin{tabular}{|c|c|}
    \hline
    {\bf Preprocessor} & {\bf Space configures}\\
    \hline
    Binarizer & 'threshold' = from 0 to 1 with 0.05 step \\
    \hline
    MinMaxScaler & \makecell{'range\_min'=0 \\ 'range\_max' = 1}\\
    \hline
    MaxAbsScaler & No parameter \\
    \hline
    Normalizer & 'norm' = ['l1', 'l2', 'max'] \\
    \hline
    StandardScaler & 'with\_mean' = [True, False] \\
    \hline
    PowerTransformer & 'standardize' = [True, False] \\
    \hline
    QuantileTransformer & \makecell{'n\_quantiles' = from 10 to 2000 with 1 step \\ 'output\_distribution' = ['uniform', 'normal']} \\
    \hline
  \end{tabular}
\end{table}

\subsection{Main Findings}
Our main findings of this section are summarized as follows:
\begin{itemize}[leftmargin=*]
    \item Different types of extended search space fit different approaches. 
    \item For extended low-cardinality search space, \textit{One-step} is a preferred approach because it can do more exploration than \textit{Two-step}.
    \item For extended high-cardinality search space, \textit{Two-step} is preferred because it can avoid selecting many duplicated preprocessors in pipelines. 
    \item Better combining pipeline search and parameter search is still an important open problem which deserves further exploration.
\end{itemize}

\section{Putting Auto-FP in an AutoML Context}\label{sec:automl}
The mission of AutoML is to democratize machine learning. AutoML automatically searches for the best ML pipeline for a given training set. Due to the large search space, recently, the decomposed Auto-ML works~\cite{ADMM-Decompose, VolcanoML, Oracle-AutoML, FLAML} introduces the space decomposition idea to speed up the search process of Auto-ML by searching good pipelines in each subspace and combining the pioneers. However, the search strategy for all subspaces in current works is still one-size-fits-all. A more reasonable choice is to design specific solutions for each subspace. By considering FP space as one individual search space in the AutoML context, we first figure out the answers to the two essential questions: 
\textit{1) Does Auto-FP outperform the feature preprocessing module in AutoML? 2) Is Auto-FP Important in AutoML Context?} To answer these questions, we investigate three popular open-source AutoML systems and evaluate the effectiveness of Auto-FP in an AutoML context using our benchmark datasets. After getting the answers to the two essential questions, we provide the discussion on \textit{What AutoML can learn from Auto-FP} to indicate the opportunities the Auto-FP experience can provide for building more powerful AutoML.

Note that the goal of our paper is not to propose an end-to-end solution and compare it with existing AutoML tools. Instead, we want to show the insight that it is valuable to separate individual steps in the AutoML flow and Auto-FP is an important part in the AutoML context. Thus, it is valuable to find a good solution for FP separately instead of a one-size-fits-all solution. How to combine Auto-FP with AutoML solutions like TPOT is out of our scope.


\subsection{Does Auto-FP Outperform FP in AutoML?} 
A typical ML pipeline consists of different components, such as  feature preprocessing, feature selection, model selection, and hyperparameter tuning. Since FP is part of an ML pipeline, a general-purpose AutoML tool can be configured to solve an Auto-FP problem, by only enabling the FP component and disabling all other components.

After investigating several popular AutoML systems, however, we find that their FP modules are quite limited. Table~\ref{tab:automl_systems} shows the capability of the FP module of Auto-WEKA~\cite{Auto-WEKA}, Auto-Sklearn~\cite{Auto-Sklearn}, and TPOT~\cite{TPOT}, respectively.
We can see that Auto-WEKA does not provide any preprocessor, thus it cannot be applied to find an FP pipeline. In comparison, Auto-Sklearn provides five preprocessors, but each FP pipeline only contains a single preprocessor and one search algorithm. TPOT allows a pipeline to contain an arbitrary number of preprocessors, but compared to Auto-FP, it has fewer preprocessors and only considers one search algorithm.

\begin{figure}[t]
\centering
\includegraphics[width=\linewidth]{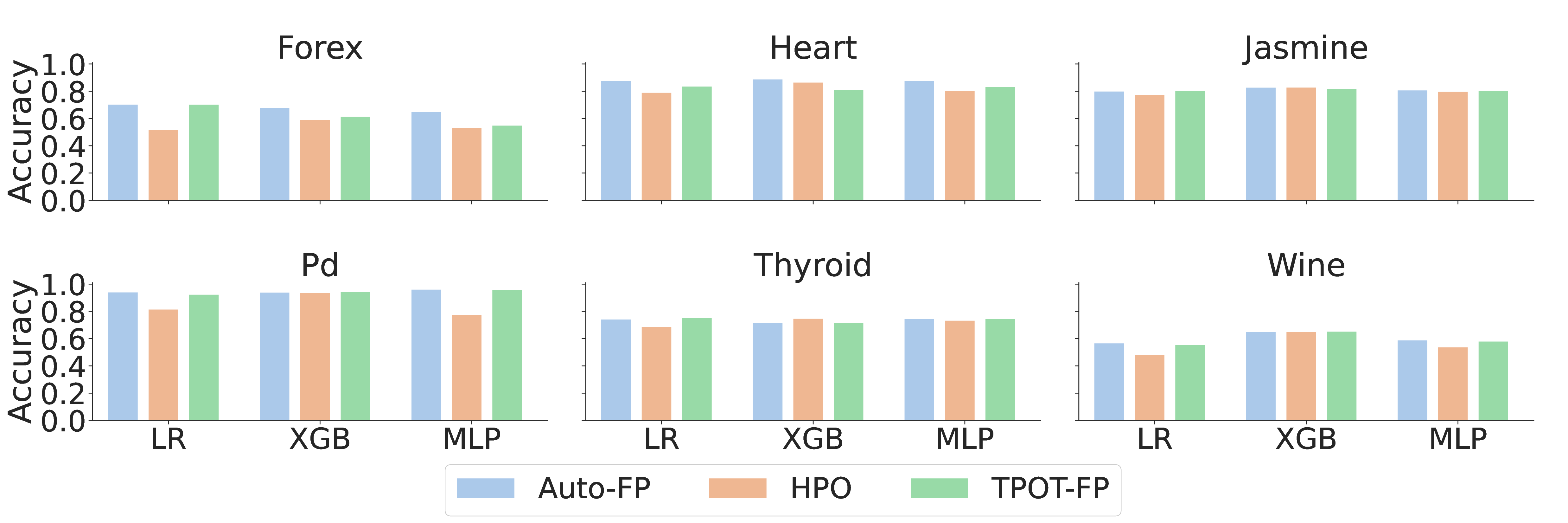}
\caption{Evaluate Auto-FP in an AutoML context (default search space). In most cases, Auto-FP outperforms TPOT-FP and comparable to HPO module in default search space. }
\label{fig:end-to-end}
\end{figure}

We compare FP in AutoML with Auto-FP. We use TPOT which has five preprocessors and applies the genetic programming~\cite{banzhaf1998genetic} algorithm for this comparison because it has a more sophisticated FP module compared to Auto-WEKA and Auto-Sklearn. We adopt the same experimental setting as Section~\ref{exp-setup} and set the time limit to 600 seconds. For TPOT, we disable all components except the FP module. For Auto-FP, the leading search algorithm \textit{PBT} is employed. Figure~\ref{fig:end-to-end} shows the results of six datasets and the complete results on all 45 datasets are shown in \cite{Auto-FP}.  We can see that Auto-FP outperforms TPOT-FP in four, three and five datasets out of six datasets (24, 25 and 25 datasets out of 45 datasets) when downstream models are LR, MLP and XGB, respectively. Besides default search space, in Figure~\ref{fig:end-to-end_balanced_space}, Auto-FP outperforms TPOT-FP in four, three and five datasets out of six datasets (24, 29, 24 out of 45 datasets) when downstream models are LR, MLP and XGB. Obviously, the conclusion that Auto-FP outperforms FP in AutoML can be generalized into a wider search space. There are two reasons that Auto-FP outperforms TPOT-FP. Firstly, Auto-FP considers more feature preprocessors. Secondly, Auto-FP adopts a better search algorithm. This result validates the necessity of designing specific solutions for each subspace.

\subsection{Is Auto-FP Important in AutoML Context?} 
\begin{figure}[t]
\centering
\includegraphics[width=\linewidth]{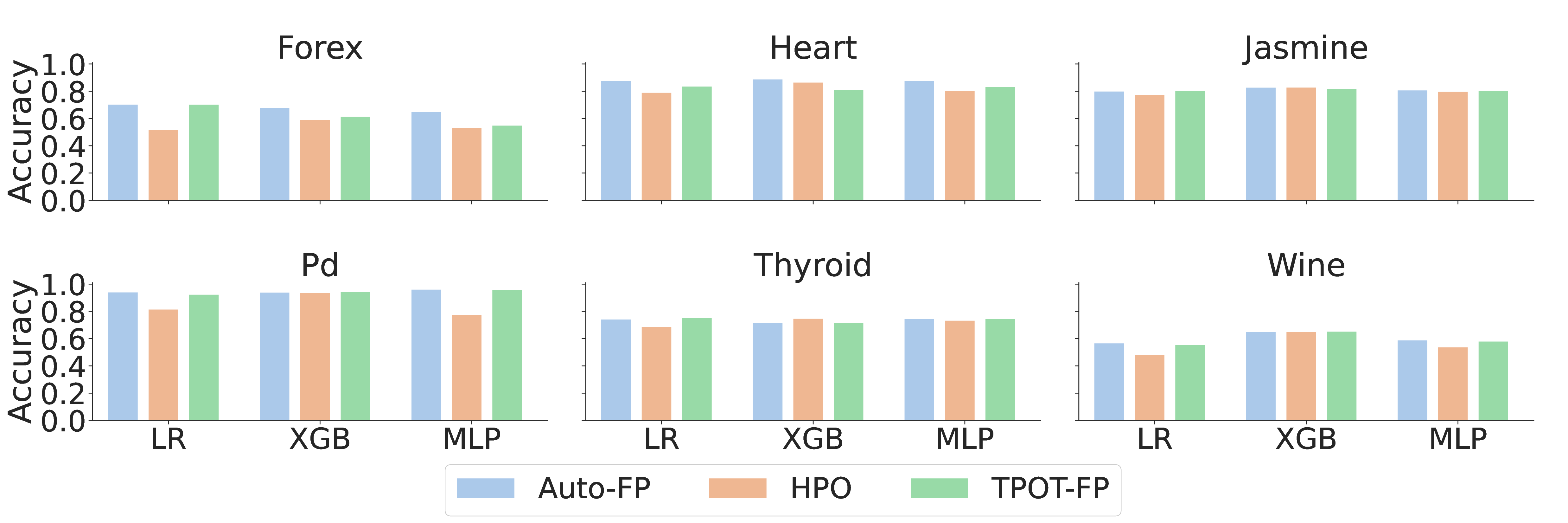}
\caption{Evaluate Auto-FP in an AutoML context (extended search space in Table~\ref{tab:extended_space}). In most cases, Auto-FP outperforms TPOT-FP and comparable to HPO module also in extended search space.}
\label{fig:end-to-end_balanced_space}
\end{figure}

To illustrate whether Auto-FP is important in the AutoML context, we explore whether Auto-FP is as important as other well-known important modules in AutoML. To answer the question, we compare Auto-FP against HPO, a well-known and highly effective module in AutoML. We also adopt the same experimental setting as Section~\ref{exp-setup} and set the time limit to 600 seconds. For TPOT, we disable all other parts except HPO. For Auto-FP, we still use \textit{PBT} as the search algorithm. Figure~\ref{fig:end-to-end_balanced_space} shows the results. We can see that Auto-FP outperforms HPO in all the datasets (40 and 37 outperform datasets out of 45 datasets) when the downstream models are LR and MLP. When the downstream model is XGB, out of six datasets, Auto-FP achieves better accuracy in three datasets (22 outperform and 2 competitive datasets out of 45 datasets). 
This result indicates that FP is as important as HPO and contributes greatly to downstream model performance improvement. Still in extended search space (Table~\ref{tab:extended_space}), Auto-FP outperforms HPO in 
all datasets (40 and 36 datasets out of 45 datasets) when downstream models are LR and MLP. For XGB, Auto-FP is better in four datasets (29 outperform and 2 competitive datasets out of 45 datasets). Obviously, Auto-FP is still as important as HPO even in other search space.

\begin{table}[t]
  \caption{Feature preprocessing module in popular open-source AutoML systems.}
  \label{tab:automl_systems}
  \centering
  \small
  \begin{tabular}{|c|c|c|c|}
    \hline
    {\bf AutoML System} & {\bf Preprocessors\#} & {\bf Pipeline Len.} & {\bf Search Algo.}  \\
    \hline
    Auto-WEKA \cite{Auto-WEKA} & 0 & 0 & SMAC \cite{SMAC} \\
    \hline
    Auto-Sklearn \cite{Auto-Sklearn} & 5 & 1 & SMAC \cite{SMAC} \\
    \hline
    TPOT \cite{TPOT} & 5 & arbitrary & GP~\cite{banzhaf1998genetic} \\
    \hline
  \end{tabular}
\end{table}

\subsection{What can AutoML learn from Auto-FP?} 
Mainstream AutoML systems are monolithic and they aim to use a one-size-fits-all search algorithm to automate every step (feature preprocessing, feature selection, hyperparameter tuning, etc.) in an ML pipeline. 
Due to the use of a monolithic architecture,
mainstream AutoML systems suffer from two limitations. Firstly, to control the overall search space, they tend to use a smaller search space for each step. For example, the search space of the FP module of Auto-Sklearn contains only five pipelines while that of Auto-FP contains about 1 million pipelines. Secondly, the one-size-fits-all search algorithm may not be suitable for every step of the pipeline. For example, the GP algorithm adopted by TPOT may not be the best search algorithm for hyperparameter tuning.

To overcome the two limitations, combining the previously mentioned space decomposition idea and designing specific solutions for each subspace is a promising direction. By employing space decomposition, AutoML can use a larger search space for each component without extending the whole search space exponentially. By designing specific solutions for each subspace, there is an opportunity for AutoML to improve its performance. To achieve this goal, the research community should conduct more benchmarks and develop the best solution tailored for each task, such as Automated Feature Generation and Automated Feature Selection.

\subsection{Main Findings}
Our main findings of this section are summarized as follows:
\begin{itemize}[leftmargin=*]
    \item Auto-FP outperforms FP in AutoML in most cases by considering larger search space and adopting better search algirithms. 
    \item Auto-FP is important in the AutoML context. Specifically, it is as important as the well-known HPO module.
    \item Limited search space and a one-size-fits-all search algorithm are two limitations mainstream AutoML systems are suffering. Combining the popular space decomposition idea and deploying specific solutions such as Auto-FP for each subspace is a promising direction to refurbish current systems.
\end{itemize}

\section{Research Opportunities}

\stitle{Warm-start search algorithms.} 
Evolution-based algorithms show a leading position in the Auto-FP scenario. Intuitively, a better initial population instead of random initialization is important for searching good pipelines faster. How to better warm start evolution-based algorithms for Auto-FP scenario is worthy to investigate. Meta-learning is an alternative way which is leveraged for warm-starting HPO~\cite{Meta-HPO1, Meta-HPO2, Meta-HPO3, Meta-HPO4, Meta-HPO5, Meta-HPO6, Meta-HPO7, Meta-HPO8} and database tuning~\cite{ResTune}. For Auto-FP, the initial population of newly-coming tasks can also be warm-started by historical tasks encoded by meta-features.

\stitle{Reduce data size intelligently to mitigate performance bottleneck.}
We have indicated in Section 6 that the performance bottleneck of most cases is ``Train'' and ``Prep''. To reduce the evaluation cost and explore more pipelines, simple approximation with a random sample~\cite{AutoML-downsampling} instead of full data can be leveraged. However, the influence function~\cite{InfluenceFunction} and SHAP~\cite{SHAP} reveal that each training example and feature have its own impact on downstream model accuracy. Thus, how to reduce data size intelligently for the Auto-FP scenario is an interesting research direction.

\stitle{Allocate pipeline and parameter search time budget reasonably.}
For better supporting parameter search in the Auto-FP scenario, how to allocate the limited search time to different stages, i.e. pipeline search and parameter search is worthy of exploration~\cite{Two-stage1, Two-stage2}. Allocating too much time for searching pipelines may reduce the opportunity to fine-tune good pipelines and get better performance. While allocating too much time for searching parameter may miss promising pipelines. There is still a trade-off between the time budget for searching pipelines and parameters.

\stitle{Benchmark other tasks in the AutoML context.}
In Section 8, we mentioned that decomposing the current AutoML search space and designing specific solutions for each subspace is a promising direction. To figure out the best algorithms for other tasks besides Auto-FP, it is valuable to do more benchmarking and combine optimal solutions of individual components, e.g.data cleaning, feature generation and feature selection, to construct a more efficient AutoML strategy.

\stitle{Benchmark Auto-FP on Other Types of Data.} While our paper focuses on tabular data, we recognize that evaluating the performance of Auto-FP on other types of data, such as text and image data, would provide a more comprehensive understanding of its capabilities. Both text and image data need specific feature preprocessors. For text data, there are various feature preprocessing methods such as TF-IDF and word embeddings, etc. Similarly, for image data, feature preprocessors like random cropping and normalization are widely used. Benchmarking Auto-FP by combining these specific feature preprocessors with current Auto-FP feature preprocessors could also be a valuable direction to explore if there is a universe-recommended search strategy.

\stitle{Benchmark Auto-FP for Deep Models for Specific Tasks.} Our paper considers the widely-used ML models including LR, XGB and MLP. In some specific domains, the deep model is the mainstream model. For example, DeepFM~\cite{Deepfm} and DCN~\cite{DCN} are two mainstream deep models dealing with recommendation tasks.  Actually, Auto-FP is applicable for deep models. Consider two recommendation datasets \textit{Tmall}\footnote{https://tianchi.aliyun.com/competition/entrance/231576/introduction} and \textit{Instacart}\footnote{https://www.kaggle.com/c/instacart-market-basket-analysis} with DeepFM as the downstreaming model. With/without 200 random FP pipelines, the valid AUC of \textit{Tmall} is 0.5875/0.5 and the valid AUC of \textit{Instacart} is 0.4756/0.7085. However, deep models for specific tasks may require specific search algorithms. Thus, benchmarking Auto-FP on deep models for specific tasks would provide a more practical direction for applying Auto-FP.

\section{Related Work}

\stitle{HPO and NAS.} Auto-FP is similar in spirit to HPO and NAS. The goal of HPO is to find the best combination of a classifier and its related hyperparameters for a given dataset. There are many search algorithms proposed for HPO~\cite{Random-Search, SMBO, SMAC, TPE, Hyperband, BOHB}. NAS aims to automatically find the best DNN architecture. Zoph and Le~\cite{NAS-RL} did the pioneering work in NAS. Later, many improved NAS algorithms are proposed~\cite{NAS-Transfer, zhong2018practical, PNAS, ENAS, TEVO}. In this paper, we model Auto-FP as HPO and NAS respectively in order to utilize these search algorithms. However, Auto-FP is fundamentally different because it has a totally difference search space.  The search space of HPO includes classifiers and hyperparameters and the search space of NAS includes DNN operators, while the search space of Auto-FP includes feature preprocessors.

\stitle{AutoML/Data Preparation Pipeline Search.} Recently, AutoML attracts great attention from the database community~\cite{AutoML-Zeyuan_Shang,Oracle-AutoML,VolcanoML,lee2019human}.  However, existing studies are mainly focused on optimizing the entire AutoML pipeline rather than diving into the feature preprocessing stage. We have discussed how AutoML systems can benefit from our study in Section~\ref{sec:automl}. Our work is also related to \textit{AutoPipeline}~\cite{AutoPipeline}, which aims to generate multiple data preparation steps automatically. It would be interesting to explore whether our Auto-FP search algorithms can be applied to solve their problem. 




\stitle{AutoML Benchmark.} OpenML~\cite{AutoML-Benchmark} is a popular AutoML benchmark which utilizes 39 public datasets to evaluate classification performance with different time slots and different metrics. Zogaj et al.~\cite{AutoML-downsampling} conduct an extensive empirical study to investigate the impact of downsampling on AutoML results.  Several benchmarks are published in the NAS area called NAS-Bench 101~\cite{NAS-Bench-101}, 201~\cite{NAS-Bench-201} and 301~\cite{NAS-Bench-301}. There are also benchmarks for some stages in the data science life cycle such as data cleaning~\cite{CleanML} and feature type recognition~\cite{FeatureTypeBench}. Different from existing work, we are the first to benchmark automated feature processing. 

\section{Conclusion}

In this paper, we studied an essential task in classical ML--feature preprocessing. We justified the importance of FP and pointed out that FP should be automated due to its large search space. We benchmarked Auto-FP with 15 search algorithms from HPO and NAS, 3 popular downstream ML models, and 7 widely-used preprocessors on 45 public datasets. We found that evolution-based algorithms take the top position. To further enhance Auto-FP, We also explored two kinds of extended parameter search space and compared two methods to support parameter search. We concluded that different search spaces fit different methods. In the end, we evaluated Auto-FP in an AutoML context and figured out that Auto-FP is an important part in the AutoML context which deserves a specific solution. How to decompose the AutoML search space reasonably and conduct effective solutions for other tasks in the AutoML context is a promising direction for the community to further explore.

\bibliographystyle{ACM-Reference-Format}
\bibliography{ref}

\begin{appendix}

\section{Detailed Information of All Datasets}

\begin{table*}[t]
  \vspace{-1em}
  \caption{Basic information of all 45 datasets.}
  \vspace{-1em}
  \centering
  \footnotesize
\begin{tabular}{|p{0.2\linewidth}<{\centering}
                  |p{0.15\linewidth}<{\centering}
                  |p{0.15\linewidth}<{\centering}
                  |p{0.15\linewidth}<{\centering}
                  |p{0.15\linewidth}<{\centering}|}
    \hline
    \textbf{Dataset} & \textbf{Size (MB)} & \textbf{\# of rows} & \textbf{\# of cols} & \textbf{\# of classes} \\
    \hline
    ada & 0.34 & 3317 & 48 & 2 \\ \hline 
    austrilian & 0.02 & 552 & 14 & 2 \\ \hline 
    blood & 0.01 & 598 & 4 & 2 \\ \hline 
    christine & 32.5 & 4334 & 1636 & 2 \\ \hline 
    Click\_prediction\_small & 2.4 & 31958 & 11 & 2 \\ \hline 
    covtype & 75.2 & 464809 & 54 & 7 \\ \hline 
    credit & 2.7 & 24000 & 23 & 2 \\
\hline 

  \end{tabular}
  \label{tab:datasets_45}\vspace{-1em}
\end{table*}


\begin{table*}[t]
  \vspace{-1em}
  \caption{Information of 40 meta-features (detailed meta-features of all datasets can be seen in {\color{blue}\url{https://github.com/AutoFP/Auto-FP/blob/main/metafeatures.csv}}).}
  \vspace{-1em}
  \centering
  \footnotesize
  \begin{tabular}{|p{0.1\linewidth}<{\centering}
                  |p{0.3\linewidth}<{\centering}
                  |p{0.45\linewidth}<{\centering}|}
    \hline
    \textbf{Group} & \textbf{Name} & \textbf{Description}\\ \hline
    \multirow{18}{*}{Simple} & NumberOfMissingValues & Number of missing cells in dataset\\ \cline{2-3}
    & PercentageOfMissingValues & Number of missing cells dividing number of cells\\ \cline{2-3} 
    & NumberOfFeaturesWithMissingValues & Number of features including missing cells\\ \cline{2-3}
    & PercentageOfFeaturesWithMissingValues & Number of features including missing cells dividing number of features\\ \cline{2-3}
    & NumberOfInstancesWithMissingValues & Number of rows including missing cells\\ \cline{2-3}
    & PercentageOfInstancesWithMissingValues & Number of rows including missing cells dividing number of rows\\ \cline{2-3}
    & NumberOfFeatures & Number of features in dataset\\
\hline
    
  \end{tabular}
  \label{tab:meta-features}\vspace{-1em}
\end{table*}

\begin{table*}[t]
  \vspace{-1em}
  \caption{Performance of 200-iterations \textit{RS} for all collected datasets.}
  \vspace{-1em}
  \centering
  \footnotesize
  \begin{tabular}{|p{0.2\linewidth}<{\centering}
                  |p{0.1\linewidth}<{\centering}
                  |p{0.1\linewidth}<{\centering}
                  |p{0.1\linewidth}<{\centering}
                  |p{0.1\linewidth}<{\centering}
                  |p{0.1\linewidth}<{\centering}
                  |p{0.1\linewidth}<{\centering}|}
    \hline
    \textbf{Dataset} & 
    \textbf{\makecell{LR Acc \\ (no prep)}} & \textbf{\makecell{LR Acc \\ (200 iter RS)}} & \textbf{\makecell{XGB Acc \\ (no prep)}} & \textbf{\makecell{XGB Acc \\ (200 iter RS)}} & \textbf{\makecell{MLP Acc \\ (no prep)}} & \textbf{\makecell{MLP Acc \\ (200 iter RS)}}\\
    \hline
    ada & 0.8464 & 0.8608 & 0.8554 & 0.8654 & 0.7455 & 0.872\\ \hline 
    austrilian & 0.8559 & 0.8738 & 0.8378 & 0.8873 & 0.7658 & 0.8738\\ \hline 
    blood & 0.75 & 0.7833 & 0.725 & 0.7625 & 0.75 & 0.775\\ \hline 
    christine & 0.6713 & 0.7303 & 0.7474 & 0.7494 & 0.6171 & 0.7144\\ \hline 
    Click\_prediction\_small & 0.8312 & 0.8333 & 0.8343 & 0.835 & 0.8312 & 0.8351\\ \hline 
    covtype & 0.6196 & 0.811 & 0.8113 & 0.8137 & 0.7771 & 0.8189\\ \hline 
    credit & 0.776 & 0.6501 & 0.9332 & 0.9339 & 0.5244 & 0.9248\\
\hline 

  \end{tabular}
  \label{tab:datasets_45}\vspace{-1em}
\end{table*}

\section{Complete Experiment Results}

\begin{table*}[htbp] 
\vspace{-1em} 
  \caption{Validation accuracy improvement using different search algorithms on 45 datasets under 60 seconds time constrain.}
  \vspace{-.5em}
  \label{tab:delta_acc_60}
  \footnotesize
  \begin{tabular}{|p{0.06\textwidth}<{\centering}
                  |p{0.034\textwidth}<{\centering}
                  |p{0.03\textwidth}<{\centering}
                  p{0.034\textwidth}<{\centering}
                  p{0.03\textwidth}<{\centering}
                  p{0.034\textwidth}<{\centering}
                  p{0.045\textwidth}<{\centering}
                  p{0.04\textwidth}<{\centering}
                  p{0.032\textwidth}<{\centering}
                  p{0.07\textwidth}<{\centering}
                  p{0.034\textwidth}<{\centering}
                  p{0.08\textwidth}<{\centering}
                  p{0.035\textwidth}<{\centering}
                  p{0.03\textwidth}<{\centering}
                  p{0.03\textwidth}<{\centering}
                  p{0.03\textwidth}<{\centering}
                  p{0.035\textwidth}<{\centering}|}
    \hline
    \textbf{Dataset} & \textbf{ML Model} & \textbf{RS} & \textbf{Anneal} & \textbf{TPE} & \textbf{SMAC} & \textbf{TEVO\_H} & \textbf{TEVO\_Y} & \textbf{PBT} & \textbf{REINFORCE} & \textbf{ENAS} & \textbf{HYPERBAND} & \textbf{BOHB} & \textbf{PMNE} & \textbf{PME} & \textbf{PLNE} & \textbf{PLE} \\
  \hline
	& LR & 1.33 & 1.39 & 1.2 & 0.75 & 1.6 & \textbf{1.72} & \textbf{1.72} & 0.69 & 0.0 & 0.0 & 0.0 & 1.66 & 1.66 & 0.45 & 0.45\\
	Ada 	& XGB & 0.72 & 0.36 & 0.36 & \textbf{0.9} & 0.54 & 0.54 & 0.42 & 0.0 & 0.0 & 0.0 & 0.0 & 0.6 & 0.6 & 0.3 & 0.3\\
	& MLP & 11.02 & 10.93 & 11.3 & 10.84 & 10.78 & 10.78 & 11.2 & 9.22 & 8.83 & 0.0 & 0.0 & \textbf{11.57} & 11.3 & 11.36 & 11.36\\
\hline
	& LR & \textbf{3.06} & 1.8 & 1.8 & 1.8 & 2.34 & 2.7 & 2.52 & 2.88 & 2.34 & 1.98 & 1.44 & 1.98 & 1.98 & 1.26 & 0.9\\
	Austrilian 	& XGB & 3.96 & 3.24 & 3.6 & 3.6 & 4.5 & \textbf{4.68} & 4.5 & 3.6 & 2.34 & 3.42 & 1.26 & 3.6 & 3.6 & 1.8 & 0.0\\
	& MLP & 11.17 & 10.99 & 10.45 & 9.91 & \textbf{11.71} & \textbf{11.71} & 11.17 & 10.09 & 9.55 & 9.91 & 9.91 & 10.09 & 10.09 & 10.45 & 9.91\\
\hline
\end{tabular}
\end{table*}

\begin{table*}[htbp] 
\vspace{-1em} 
  \caption{Validation accuracy improvement using different search algorithms on 45 datasets under 60 seconds time constrain (cont.).}
  \vspace{-.5em}
  \label{tab:delta_acc_60}
  \footnotesize
  \begin{tabular}{|p{0.06\textwidth}<{\centering}
                  |p{0.034\textwidth}<{\centering}
                  |p{0.03\textwidth}<{\centering}
                  p{0.034\textwidth}<{\centering}
                  p{0.03\textwidth}<{\centering}
                  p{0.034\textwidth}<{\centering}
                  p{0.045\textwidth}<{\centering}
                  p{0.04\textwidth}<{\centering}
                  p{0.032\textwidth}<{\centering}
                  p{0.07\textwidth}<{\centering}
                  p{0.034\textwidth}<{\centering}
                  p{0.08\textwidth}<{\centering}
                  p{0.035\textwidth}<{\centering}
                  p{0.03\textwidth}<{\centering}
                  p{0.03\textwidth}<{\centering}
                  p{0.03\textwidth}<{\centering}
                  p{0.035\textwidth}<{\centering}|}
    \hline
    \textbf{Dataset} & \textbf{ML Model} & \textbf{RS} & \textbf{Anneal} & \textbf{TPE} & \textbf{SMAC} & \textbf{TEVO\_H} & \textbf{TEVO\_Y} & \textbf{PBT} & \textbf{REINFORCE} & \textbf{ENAS} & \textbf{HYPERBAND} & \textbf{BOHB} & \textbf{PMNE} & \textbf{PME} & \textbf{PLNE} & \textbf{PLE} \\
  \hline
	& LR & 6.32 & 6.18 & 6.18 & 6.62 & 6.62 & \textbf{7.5} & 6.32 & 6.62 & 5.44 & 6.03 & 5.88 & 6.62 & 6.62 & 5.88 & 5.88\\
	Vehicle 	& XGB & 3.53 & 2.79 & 3.38 & \textbf{4.41} & 3.68 & 3.68 & 3.97 & 3.38 & 0.0 & 3.38 & 3.09 & \textbf{4.41} & \textbf{4.41} & 2.21 & 2.21\\
	& MLP & \textbf{19.85} & \textbf{19.85} & \textbf{19.85} & \textbf{19.85} & \textbf{19.85} & \textbf{19.85} & 19.56 & 17.5 & 18.82 & 18.68 & 18.82 & \textbf{19.85} & \textbf{19.85} & \textbf{19.85} & \textbf{19.85}\\
\hline
	& LR & 3.74 & 0.0 & 0.0 & 0.0 & 0.0 & 0.0 & 2.06 & 0.0 & 0.0 & 0.0 & 0.0 & \textbf{4.51} & \textbf{4.51} & \textbf{4.51} & \textbf{4.51}\\
\hline
\end{tabular}
\end{table*}

\begin{table*}[htbp] 
\vspace{-1em} 
  \caption{Validation accuracy improvement using different search algorithms on 45 datasets under 3600 seconds time constrain.}
  \vspace{-.5em}
  \label{tab:delta_acc_60}
  \footnotesize
  \begin{tabular}{|p{0.06\textwidth}<{\centering}
                  |p{0.034\textwidth}<{\centering}
                  |p{0.03\textwidth}<{\centering}
                  p{0.034\textwidth}<{\centering}
                  p{0.03\textwidth}<{\centering}
                  p{0.034\textwidth}<{\centering}
                  p{0.045\textwidth}<{\centering}
                  p{0.04\textwidth}<{\centering}
                  p{0.032\textwidth}<{\centering}
                  p{0.07\textwidth}<{\centering}
                  p{0.034\textwidth}<{\centering}
                  p{0.08\textwidth}<{\centering}
                  p{0.035\textwidth}<{\centering}
                  p{0.03\textwidth}<{\centering}
                  p{0.03\textwidth}<{\centering}
                  p{0.03\textwidth}<{\centering}
                  p{0.035\textwidth}<{\centering}|}
    \hline
    \textbf{Dataset} & \textbf{ML Model} & \textbf{RS} & \textbf{Anneal} & \textbf{TPE} & \textbf{SMAC} & \textbf{TEVO\_H} & \textbf{TEVO\_Y} & \textbf{PBT} & \textbf{REINFORCE} & \textbf{ENAS} & \textbf{HYPERBAND} & \textbf{BOHB} & \textbf{PMNE} & \textbf{PME} & \textbf{PLNE} & \textbf{PLE} \\
  \hline
	& LR & 1.81 & 1.51 & 1.72 & 1.84 & 1.87 & 1.78 & \textbf{1.96} & 1.17 & 1.23 & 1.72 & 1.78 & \textbf{1.96} & \textbf{1.96} & 1.66 & 0.93\\
	Ada 	& XGB & 0.96 & 0.6 & 1.2 & 1.3 & 1.75 & 1.63 & \textbf{2.05} & 1.08 & 0.0 & 0.9 & 0.9 & 1.6 & 1.6 & 1.11 & 0.69\\
	& MLP & 12.68 & 12.41 & 12.62 & 12.68 & \textbf{12.8} & 12.77 & 12.71 & 11.54 & 10.78 & 12.02 & 12.08 & 12.65 & 12.65 & 12.65 & 12.17\\
\hline
	& LR & 3.96 & 1.98 & 2.88 & 2.34 & 3.6 & 3.6 & 3.96 & 2.88 & 2.52 & 3.6 & 2.88 & \textbf{4.5} & \textbf{4.5} & 2.34 & 1.98\\
	Austrilian 	& XGB & \textbf{5.59} & 3.24 & 4.68 & 3.6 & 4.5 & 4.68 & 5.41 & 3.6 & 3.42 & 5.05 & 3.78 & 5.41 & 5.41 & 4.68 & 3.6\\
	& MLP & 12.43 & 11.71 & 11.71 & 12.25 & \textbf{12.61} & 12.25 & 12.25 & 11.89 & 9.73 & 12.07 & 11.71 & 12.43 & 12.43 & 12.07 & 11.53\\
\hline
\end{tabular}
\end{table*}

\begin{table*}[htbp] 
\vspace{-1em} 
  \caption{Validation accuracy improvement using different search algorithms on 45 datasets under 3600 seconds time constrain (cont.).}
  \vspace{-.5em}
  \label{tab:delta_acc_60}
  \footnotesize
  \begin{tabular}{|p{0.06\textwidth}<{\centering}
                  |p{0.034\textwidth}<{\centering}
                  |p{0.03\textwidth}<{\centering}
                  p{0.034\textwidth}<{\centering}
                  p{0.03\textwidth}<{\centering}
                  p{0.034\textwidth}<{\centering}
                  p{0.045\textwidth}<{\centering}
                  p{0.04\textwidth}<{\centering}
                  p{0.032\textwidth}<{\centering}
                  p{0.07\textwidth}<{\centering}
                  p{0.034\textwidth}<{\centering}
                  p{0.08\textwidth}<{\centering}
                  p{0.035\textwidth}<{\centering}
                  p{0.03\textwidth}<{\centering}
                  p{0.03\textwidth}<{\centering}
                  p{0.03\textwidth}<{\centering}
                  p{0.035\textwidth}<{\centering}|}
    \hline
    \textbf{Dataset} & \textbf{ML Model} & \textbf{RS} & \textbf{Anneal} & \textbf{TPE} & \textbf{SMAC} & \textbf{TEVO\_H} & \textbf{TEVO\_Y} & \textbf{PBT} & \textbf{REINFORCE} & \textbf{ENAS} & \textbf{HYPERBAND} & \textbf{BOHB} & \textbf{PMNE} & \textbf{PME} & \textbf{PLNE} & \textbf{PLE} \\
  \hline
		& LR & 7.5 & 6.18 & 6.76 & 7.06 & \textbf{8.24} & \textbf{8.24} & \textbf{8.24} & 7.65 & 6.32 & 7.5 & 6.91 & \textbf{8.24} & \textbf{8.24} & 7.06 & 6.18\\
	Vehicle 	& XGB & 5.44 & 3.53 & 4.71 & 5.0 & \textbf{7.35} & 5.88 & 6.62 & 5.88 & 5.15 & 4.85 & 4.56 & 5.44 & 6.32 & 5.74 & 4.71\\
	& MLP & 21.32 & 19.85 & 20.0 & 21.32 & 20.0 & 19.85 & 20.44 & \textbf{22.21} & 19.85 & 21.32 & 20.88 & 21.32 & 21.32 & 20.74 & 20.44\\
\hline
	& LR & 5.6 & 5.76 & 5.71 & 5.53 & \textbf{5.86} & 5.8 & 5.83 & 5.13 & 1.08 & 5.59 & 5.44 & 5.8 & 5.79 & 5.42 & 5.42\\
	Volkert 	& XGB & 0.3 & 0.0 & 0.0 & 0.3 & 0.0 & 0.0 & 0.49 & 0.0 & 0.0 & 0.0 & 0.0 & \textbf{0.6} & \textbf{0.6} & \textbf{0.6} & \textbf{0.6}\\
	& MLP & 0.28 & 0.17 & 0.0 & 0.71 & 0.67 & 0.67 & \textbf{1.05} & 0.0 & 0.0 & 0.35 & 0.24 & 0.93 & 0.71 & 0.79 & 0.79\\
\hline
\end{tabular}
\end{table*}

\begin{figure*}[h]
\vspace{-.5em}
\centering
\includegraphics[width=\textwidth]{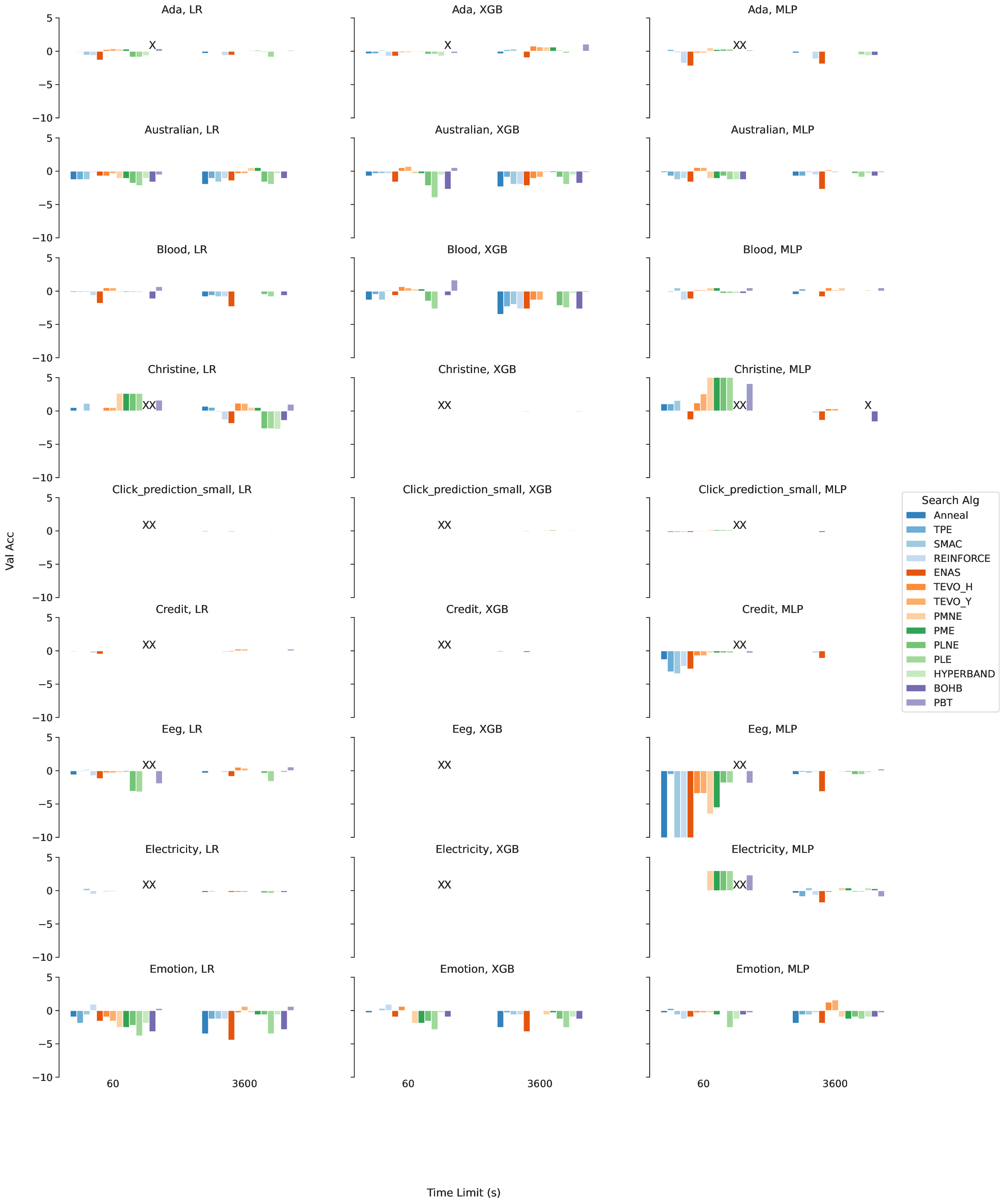}
\vspace{-1.5em}
\caption{The gap of the validation accuracy between random search and other search algorithms on all datasets with 60s and 3600s time limits - part 1.}
\label{fig:bar_vary_time_all_datasets0}
\end{figure*}

\begin{figure*}[h]
\vspace{-.5em}
\centering
\includegraphics[width=\textwidth]{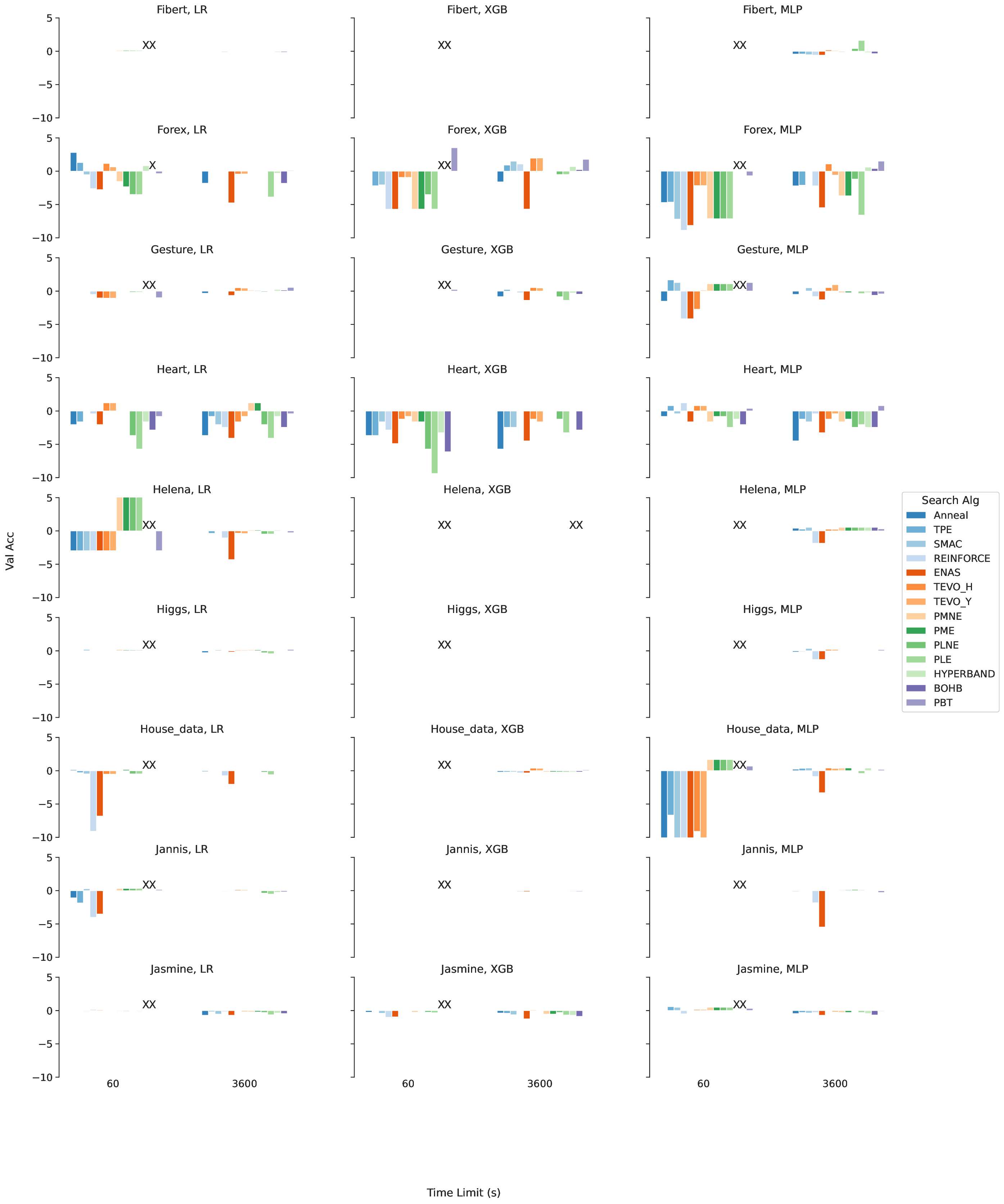}
\vspace{-1.5em}
\caption{The gap of the validation accuracy between random search and other search algorithms on all datasets with 60s and 3600s time limits - part 2.}
\label{fig:bar_vary_time_all_datasets1}
\end{figure*}

\begin{figure*}[h]
\vspace{-.5em}
\centering
\includegraphics[width=\textwidth]{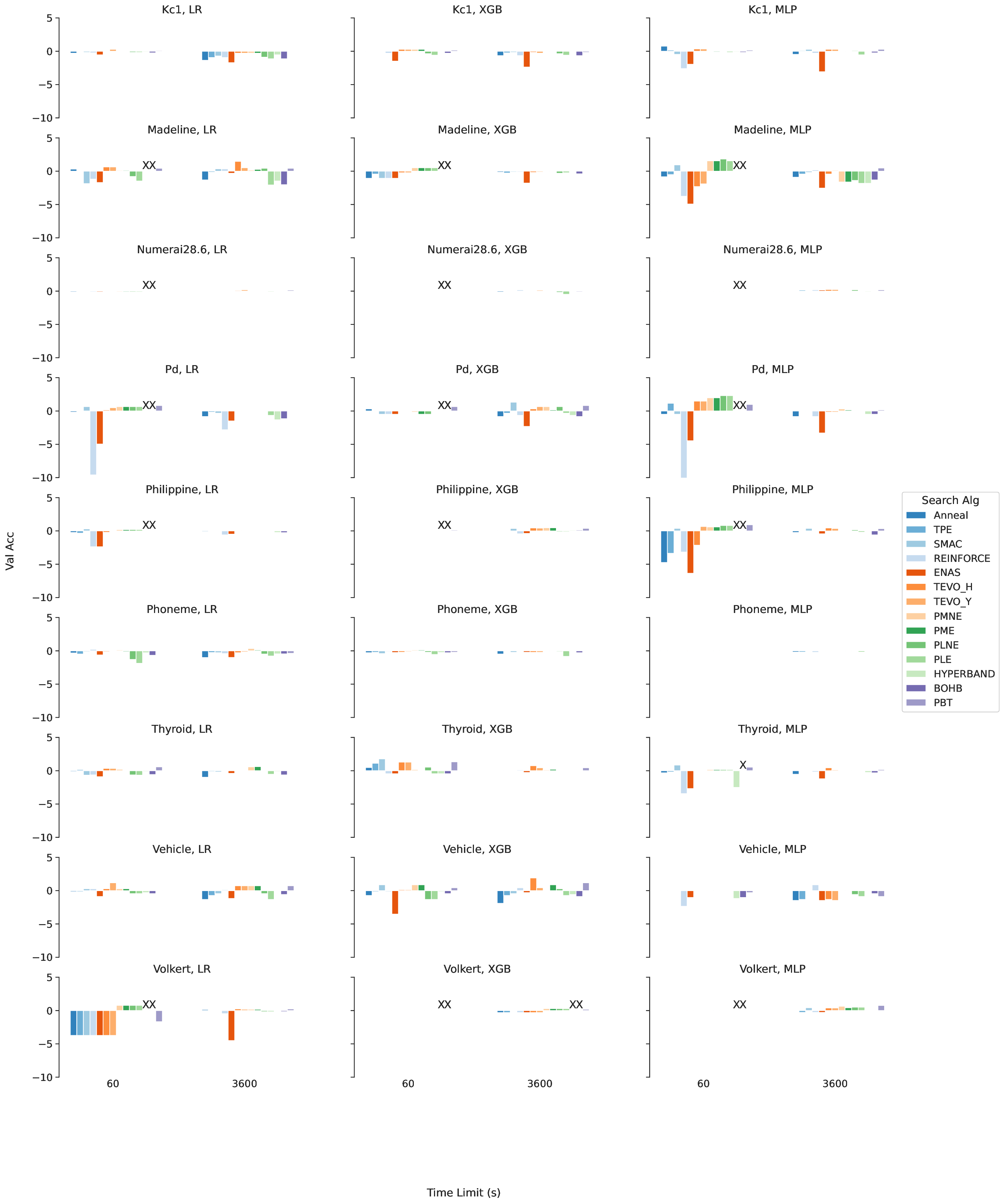}
\vspace{-1.5em}
\caption{The gap of the validation accuracy between random search and other search algorithms on all datasets with 60s and 3600s time limit - part 3.}
\label{fig:bar_vary_time_all_datasets2}
\end{figure*}

\begin{figure*}[h]
\vspace{-.5em}
\centering
\includegraphics[width=\textwidth]{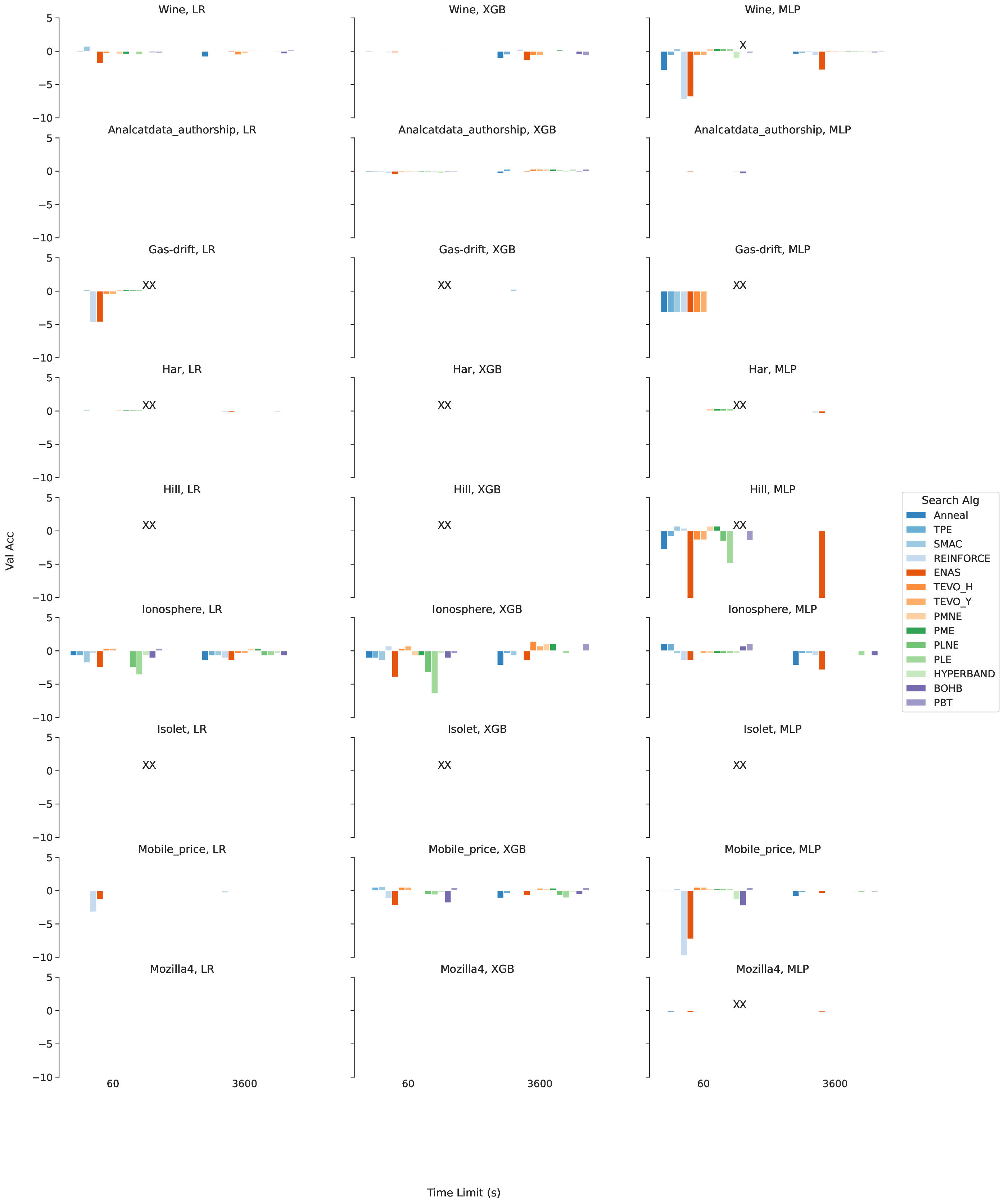}
\vspace{-1.5em}
\caption{The gap of the validation accuracy between random search and other search algorithms on all datasets with 60s and 3600s time limits - part 4.}
\label{fig:bar_vary_time_all_datasets3}
\end{figure*}

\begin{figure*}[h]
\vspace{-.5em}
\centering
\includegraphics[width=\textwidth]{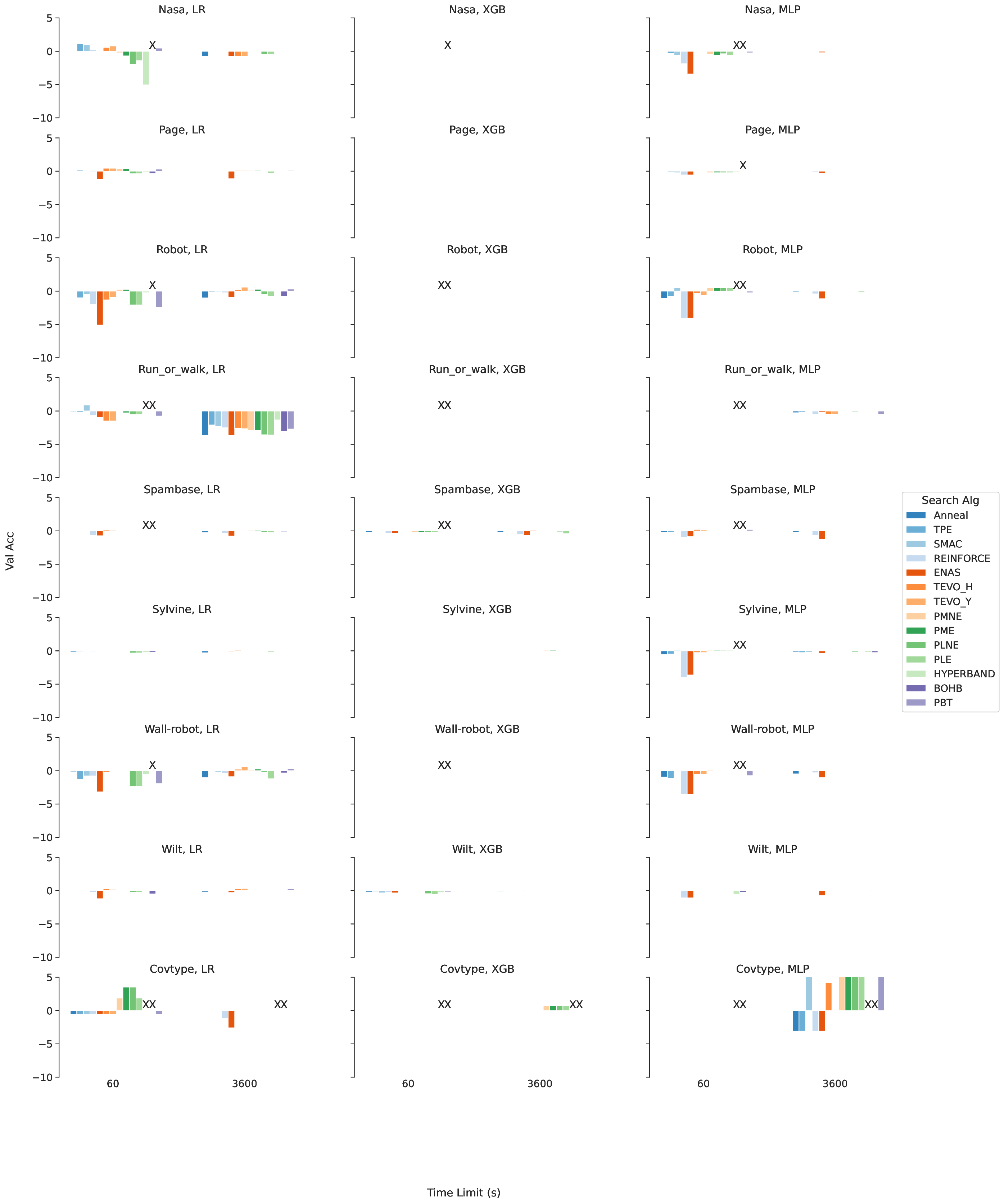}
\vspace{-1.5em}
\caption{The gap of the validation accuracy between random search and other search algorithms on all datasets with 60s and 3600s time limits - part 5.}
\label{fig:bar_vary_time_all_datasets3}
\end{figure*}


\begin{figure*}[h]
\vspace{-.5em}
\centering
\includegraphics[width=\textwidth]{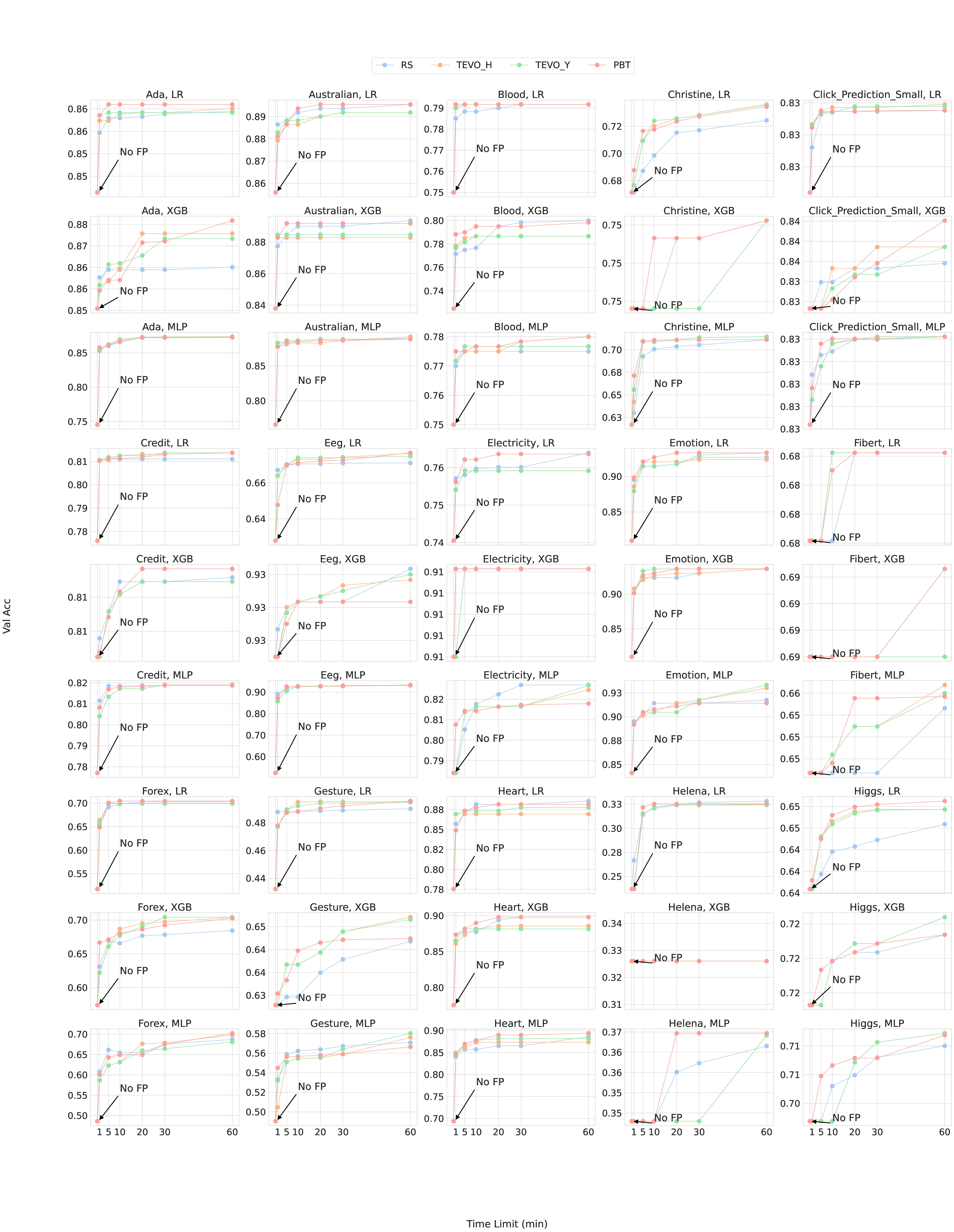}
\vspace{-1.5em}
\caption{The trend of validation accuracy with the increase
of time limit on all datasets - part 1.}
\label{fig:max_score_trend_all_datasets0}
\end{figure*}

\begin{figure*}[h]
\vspace{-.5em}
\centering
\includegraphics[width=\textwidth]{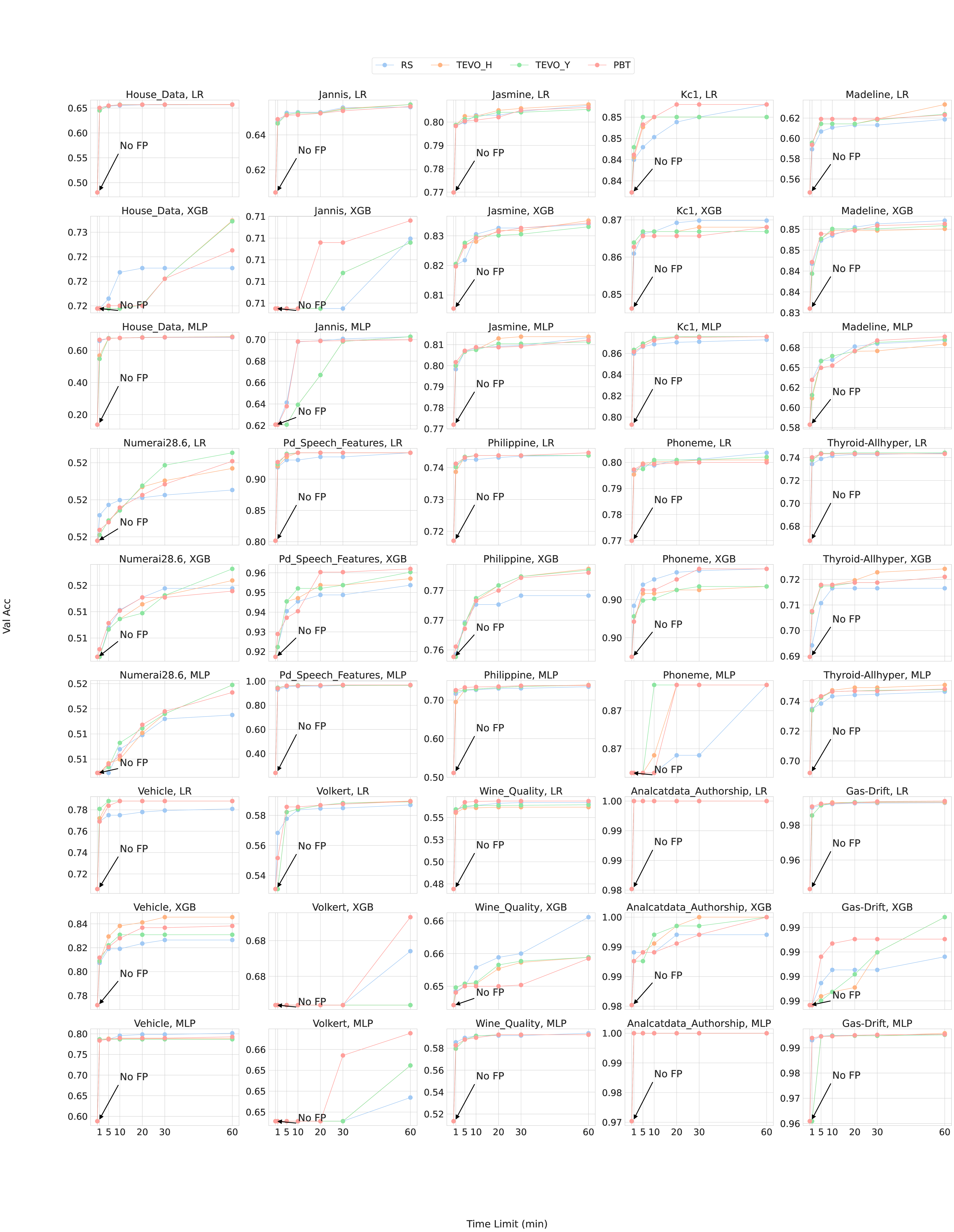}
\vspace{-1.5em}
\caption{The trend of validation accuracy with the increase
of time limit on all datasets - part 2.}
\label{fig:max_score_trend_all_datasets1}
\end{figure*}

\begin{figure*}[h]
\vspace{-.5em}
\centering
\includegraphics[width=\textwidth]{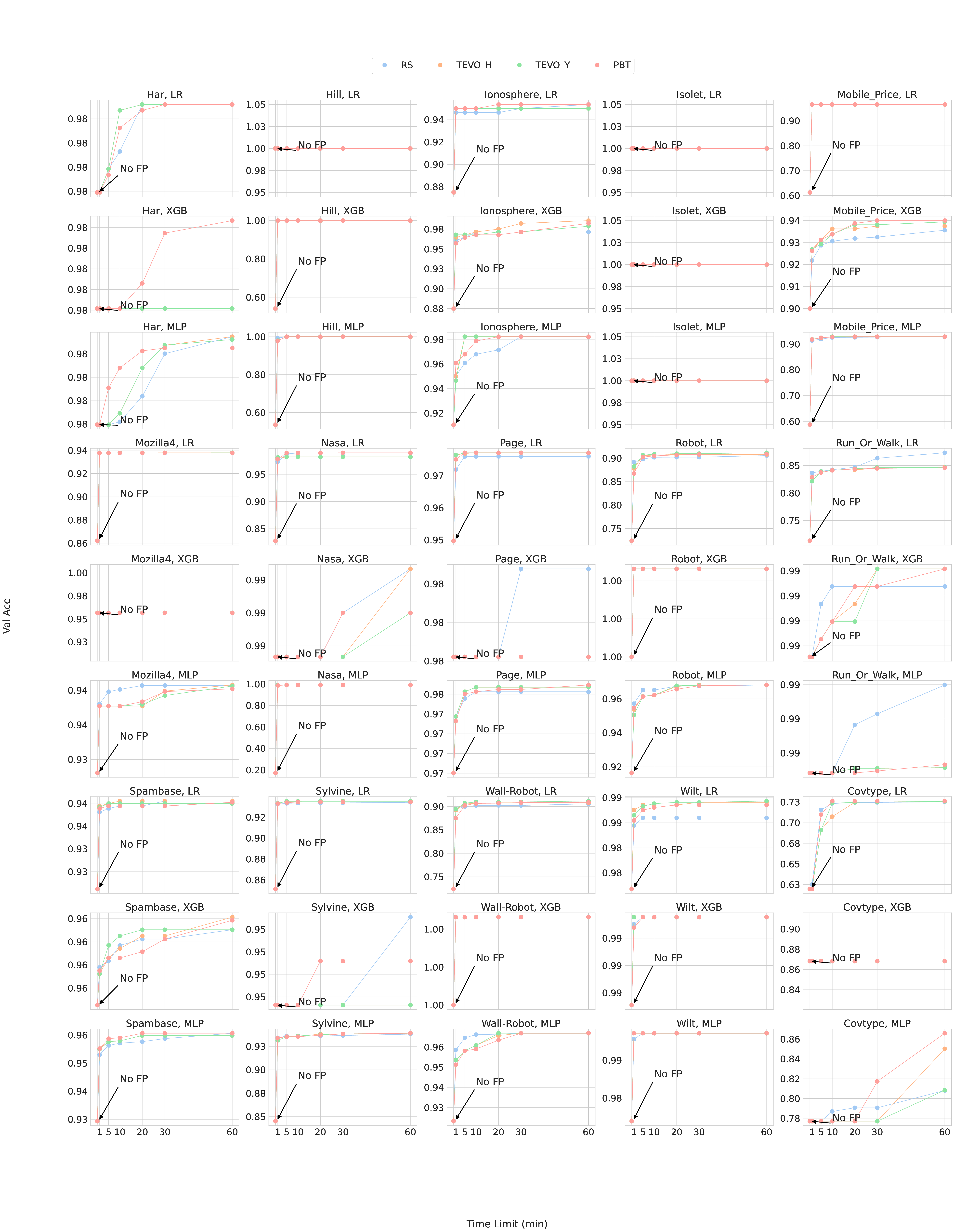}
\vspace{-1.5em}
\caption{The trend of validation accuracy with the increase
of time limit on all datasets - part 3.}
\label{fig:max_score_trend_all_datasets2}
\end{figure*}

\begin{figure*}[h]
\vspace{-.5em}
\centering
\includegraphics[width=\textwidth]{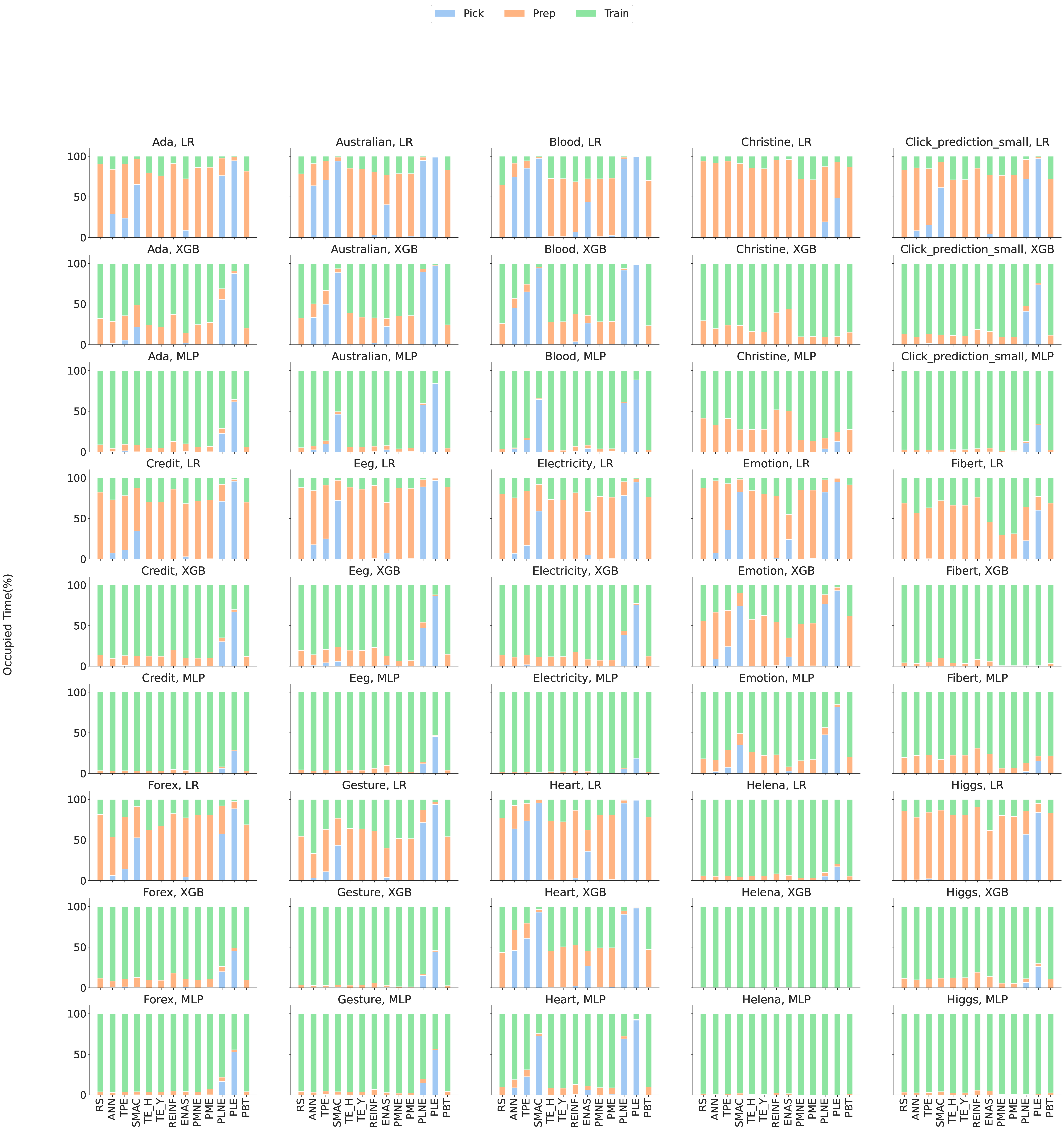}
\vspace{-1.5em}
\caption{Overhead percentage of all datasets with different downstream ML models. ``Pic'' means the overhead of picking up next pipelines. ``Prep'' means the overhead of preprocessing dataset with feature preprocessors. ``Train'' means the overhead of evaluating FP pipelines. (related to Figure 7) - part 1.}
\label{fig:overhead_all_datasets1}
\end{figure*}

\begin{figure*}[h]
\vspace{-.5em}
\centering
\includegraphics[width=\textwidth]{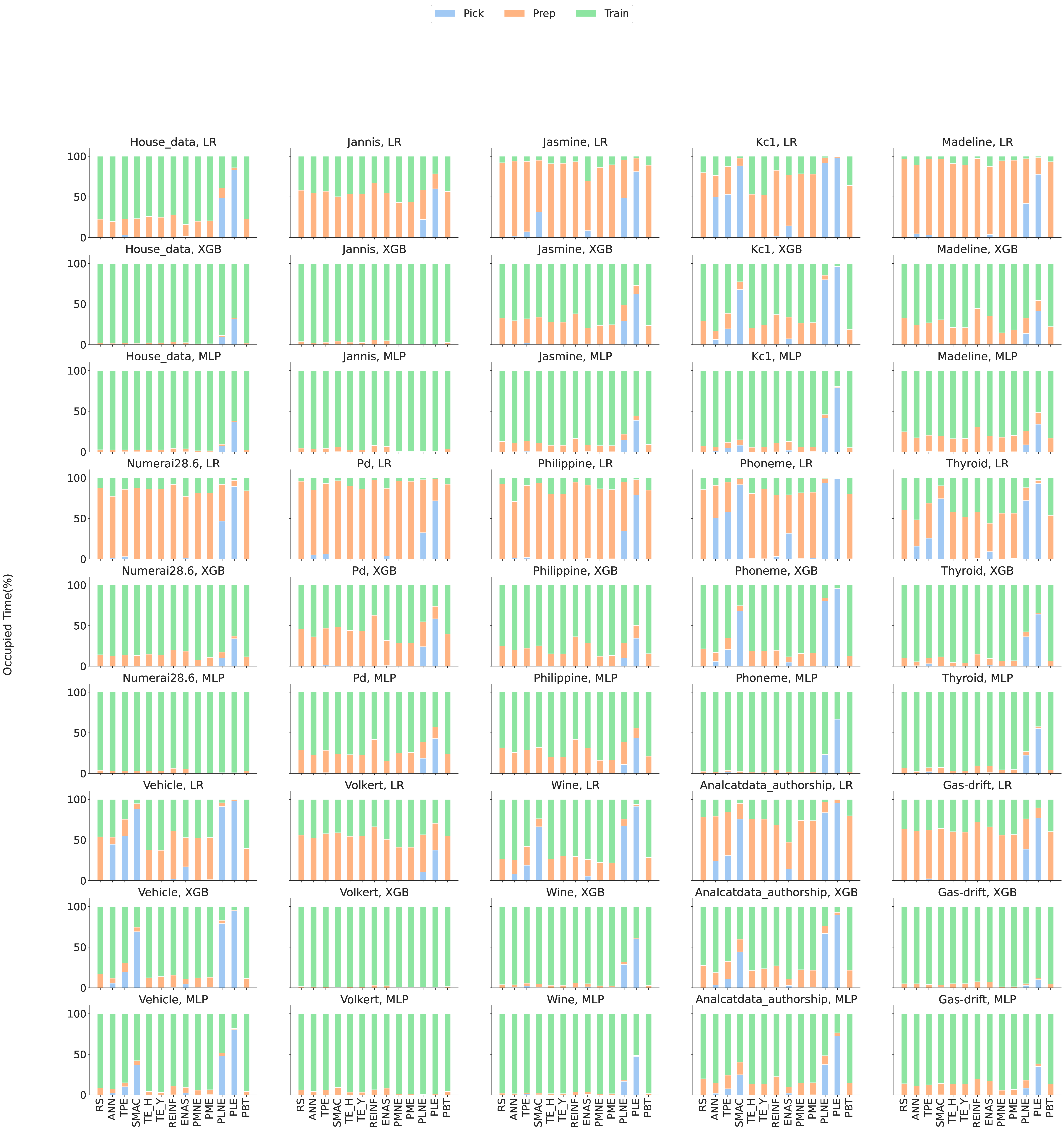}
\vspace{-1.5em}
\caption{Overhead percentage of all datasets with different downstream ML models. ``Pic'' means the overhead of picking up next pipelines. ``Prep'' means the overhead of preprocessing dataset with feature preprocessors. ``Train'' means the overhead of evaluating FP pipelines. (related to Figure 7) - part 2.}
\label{fig:overhead_all_datasets2}
\end{figure*}

\begin{figure*}[h]
\vspace{-.5em}
\centering
\includegraphics[width=\textwidth]{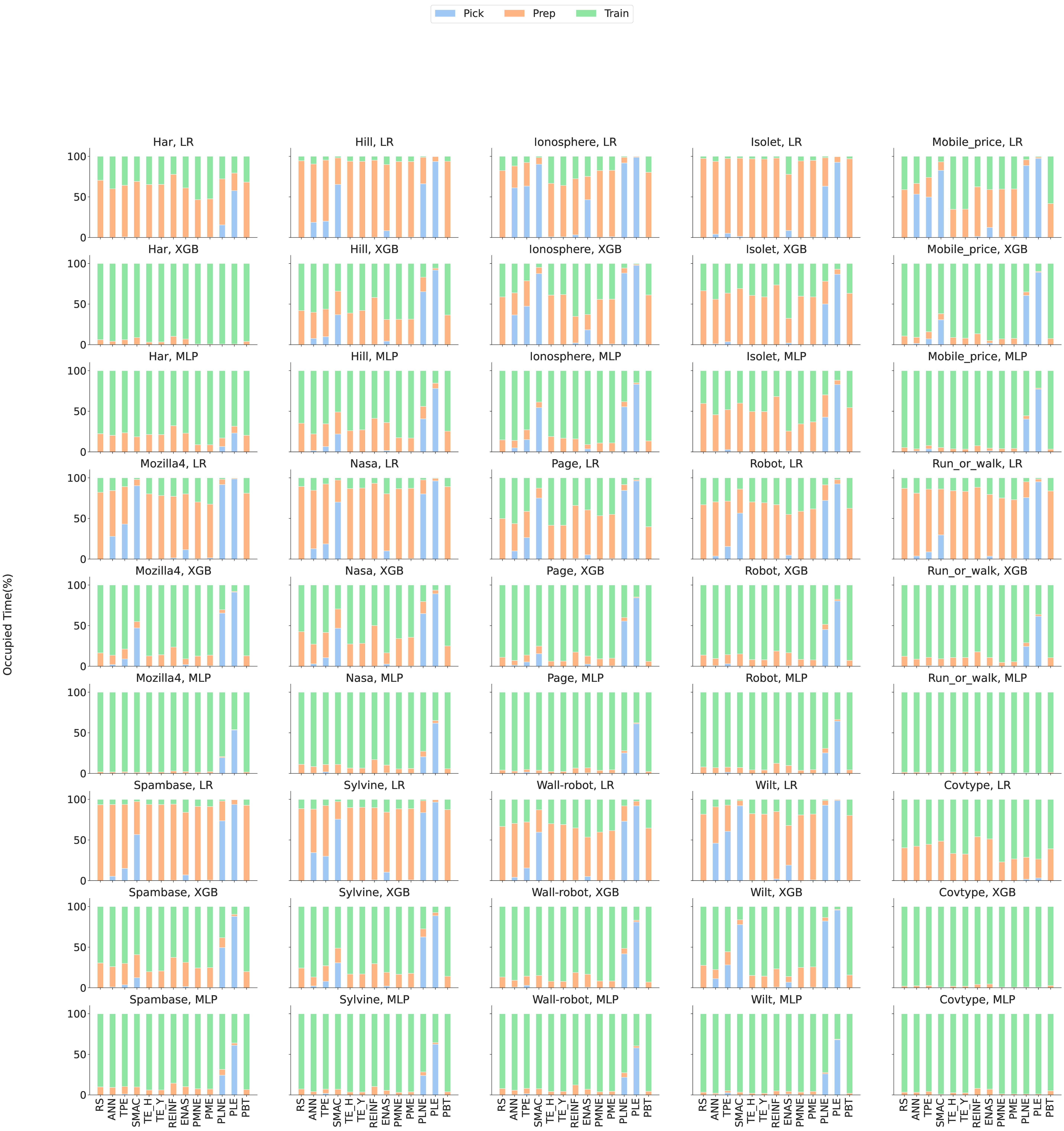}
\vspace{-1.5em}
\caption{Overhead percentage of all datasets with different downstream ML models. ``Pic'' means the overhead of picking up next pipelines. ``Prep'' means the overhead of preprocessing dataset with feature preprocessors. ``Train'' means the overhead of evaluating FP pipelines. (related to Figure 7) - part 3.}
\label{fig:overhead_all_datasets3}
\end{figure*}

\begin{figure*}[h]
\vspace{-.5em}
\centering
\includegraphics[width=\textwidth]{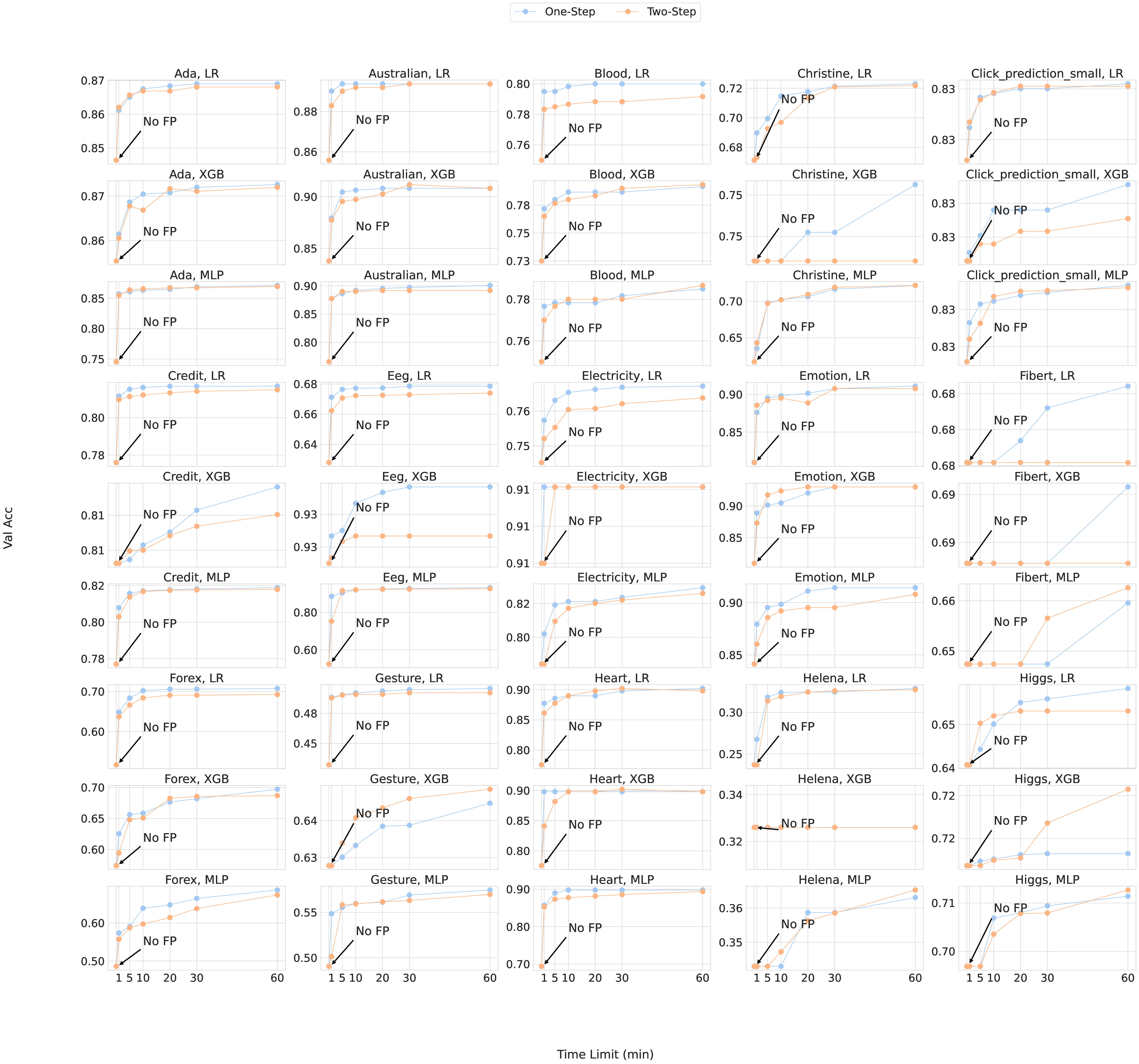}
\vspace{-1.5em}
\caption{Comparison of One-step and Two-step in the extended low-cardinality search space in Table 5 (related to Figure 8) - part 1.}
\label{fig:max_scores_by_time_extended_space_all_datasets0}
\end{figure*}

\begin{figure*}[h]
\vspace{-.5em}
\centering
\includegraphics[width=\textwidth]{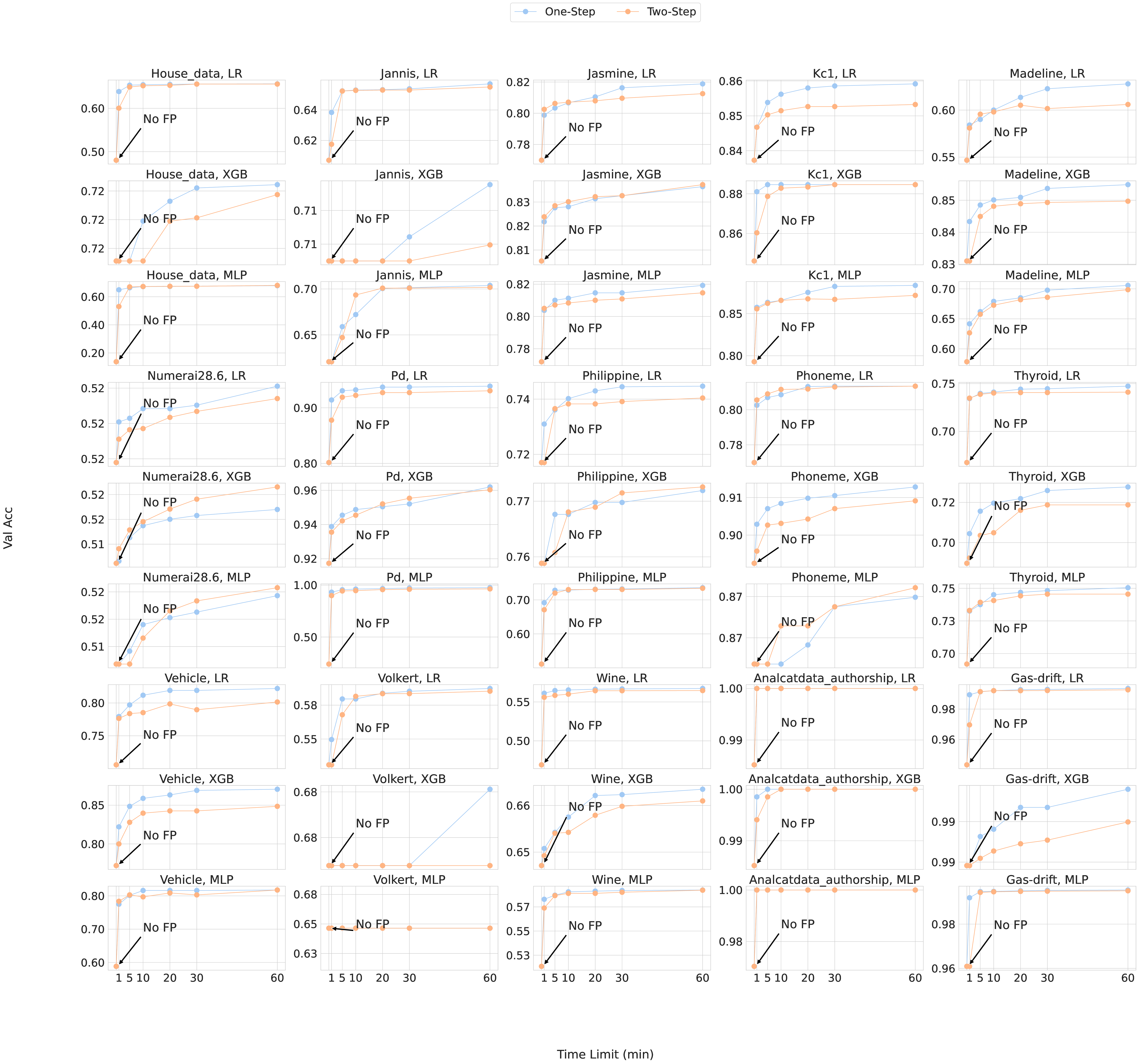}
\vspace{-1.5em}
\caption{Comparison of One-step and Two-step in the extended low-cardinality search space in Table 5 (related to Figure 8) - part 2.}
\label{fig:max_scores_by_time_extended_space_all_datasets1}
\end{figure*}

\begin{figure*}[h]
\vspace{-.5em}
\centering
\includegraphics[width=\textwidth]{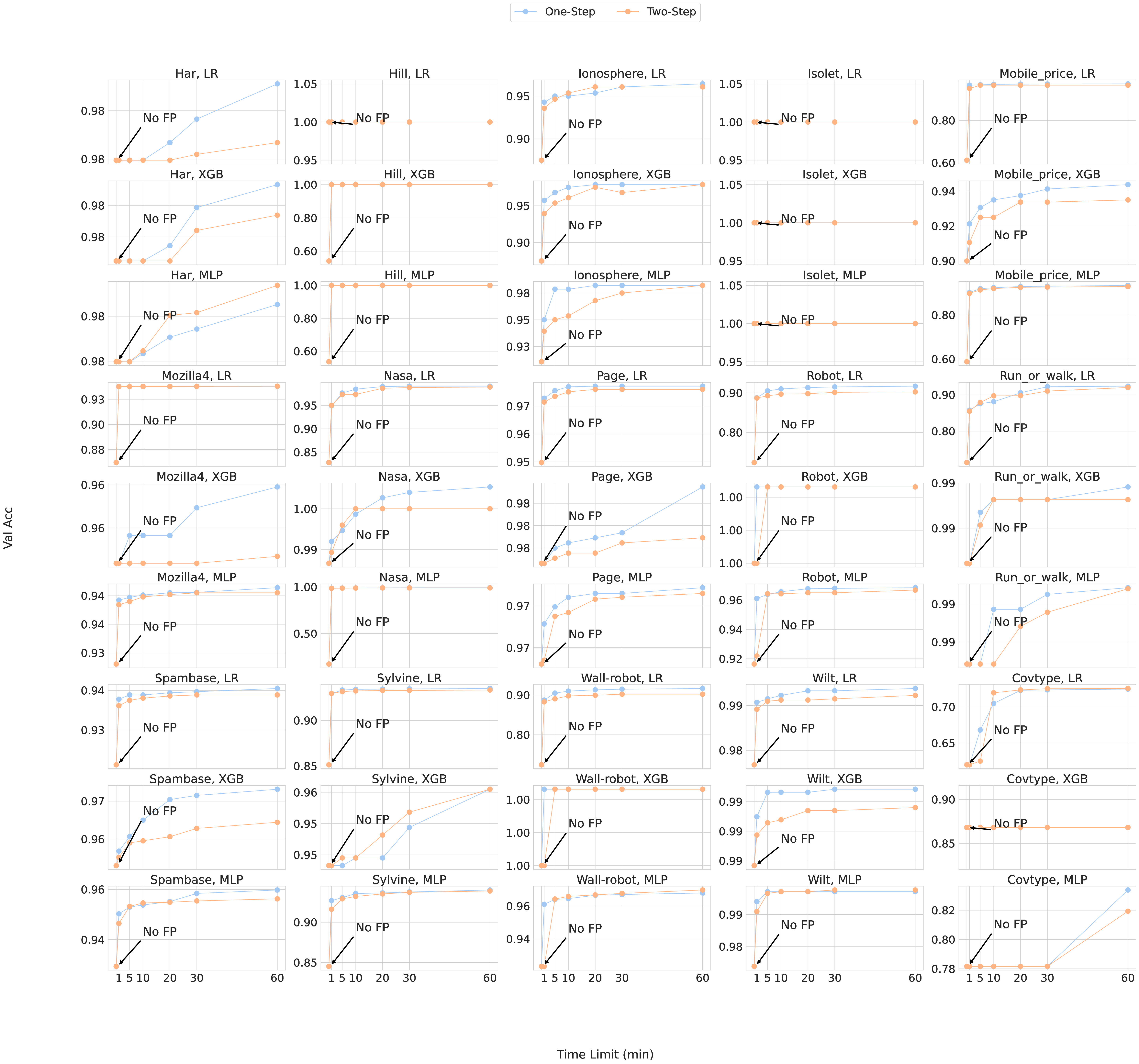}
\vspace{-1.5em}
\caption{Comparison of One-step and Two-step in the extended low-cardinality search space in Table 5 (related to Figure 8) - part 3.}
\label{fig:max_scores_by_time_extended_space_all_datasets2}
\end{figure*}


\begin{figure*}[h]
\vspace{-.5em}
\centering
\includegraphics[width=\textwidth]{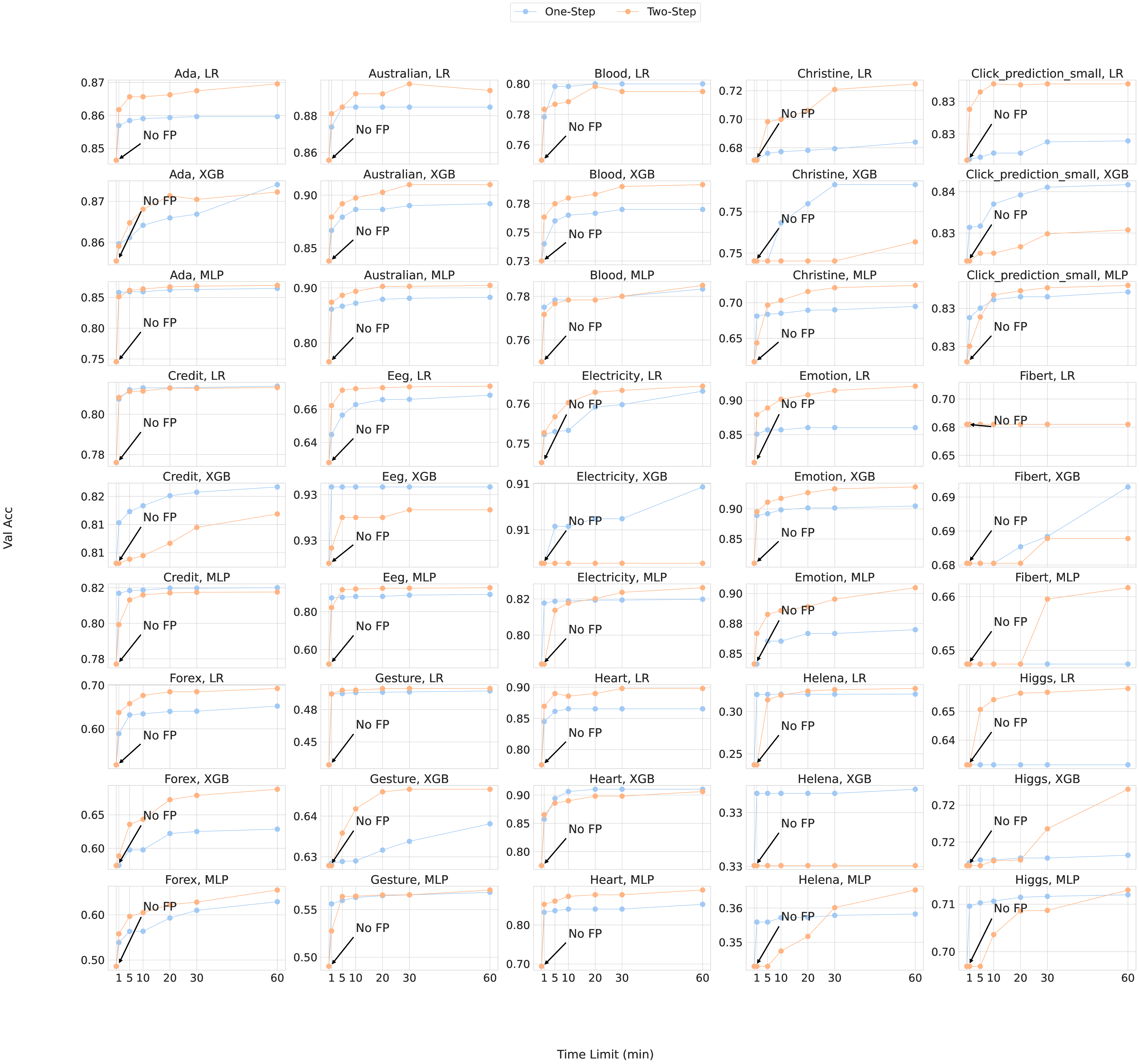}
\vspace{-1.5em}
\caption{Comparison of One-step and Two-step in the extended low-cardinality search space in Table 6 (related to Figure 9) - part 1.}
\label{fig:max_scores_by_time_extended_space_imbalanced_all_datasets0}
\end{figure*}

\begin{figure*}[h]
\vspace{-.5em}
\centering
\includegraphics[width=\textwidth]{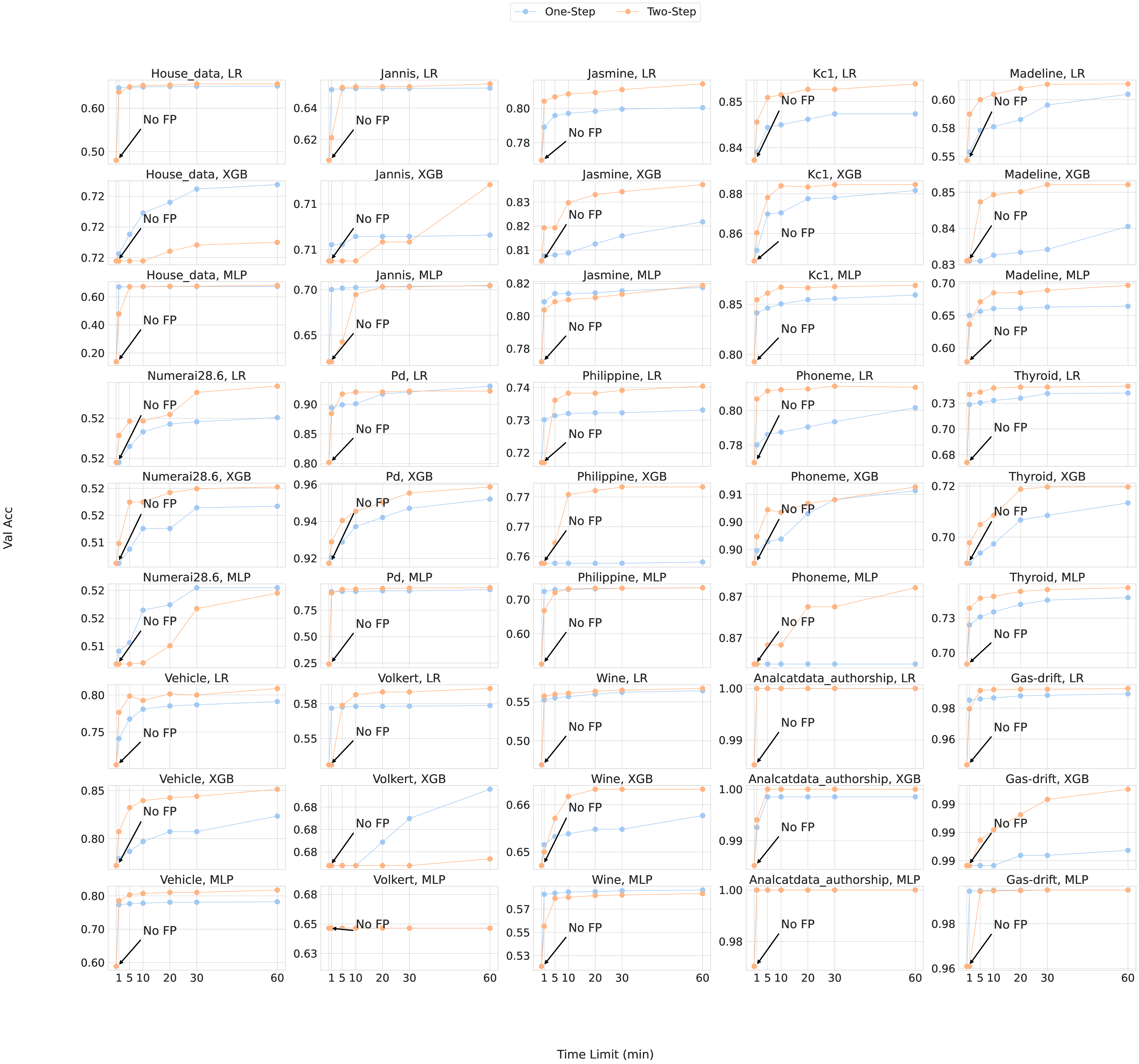}
\vspace{-1.5em}
\caption{Comparison of One-step and Two-step in the extended low-cardinality search space in Table 6 (related to Figure 9) - part 2.}
\label{fig:max_scores_by_time_extended_space_imbalanced_all_datasets1}
\end{figure*}

\begin{figure*}[h]
\vspace{-.5em}
\centering
\includegraphics[width=\textwidth]{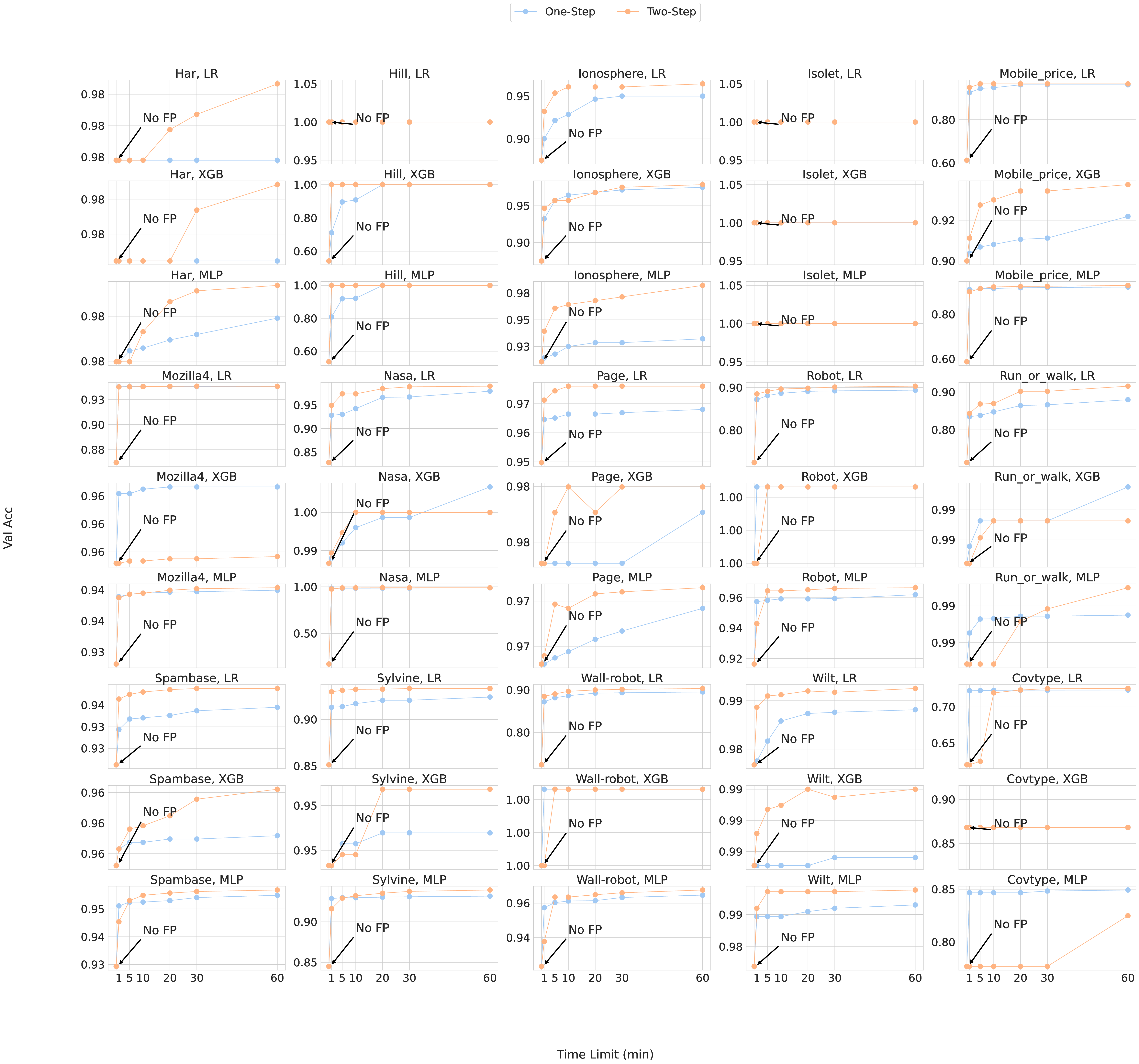}
\vspace{-1.5em}
\caption{Comparison of One-step and Two-step in the extended low-cardinality search space in Table 6 (related to Figure 9) - part 3.}
\label{fig:max_scores_by_time_extended_space_imbalanced_all_datasets2}
\end{figure*}

\begin{figure*}[h]
\vspace{-.5em}
\centering
\includegraphics[width=\textwidth]{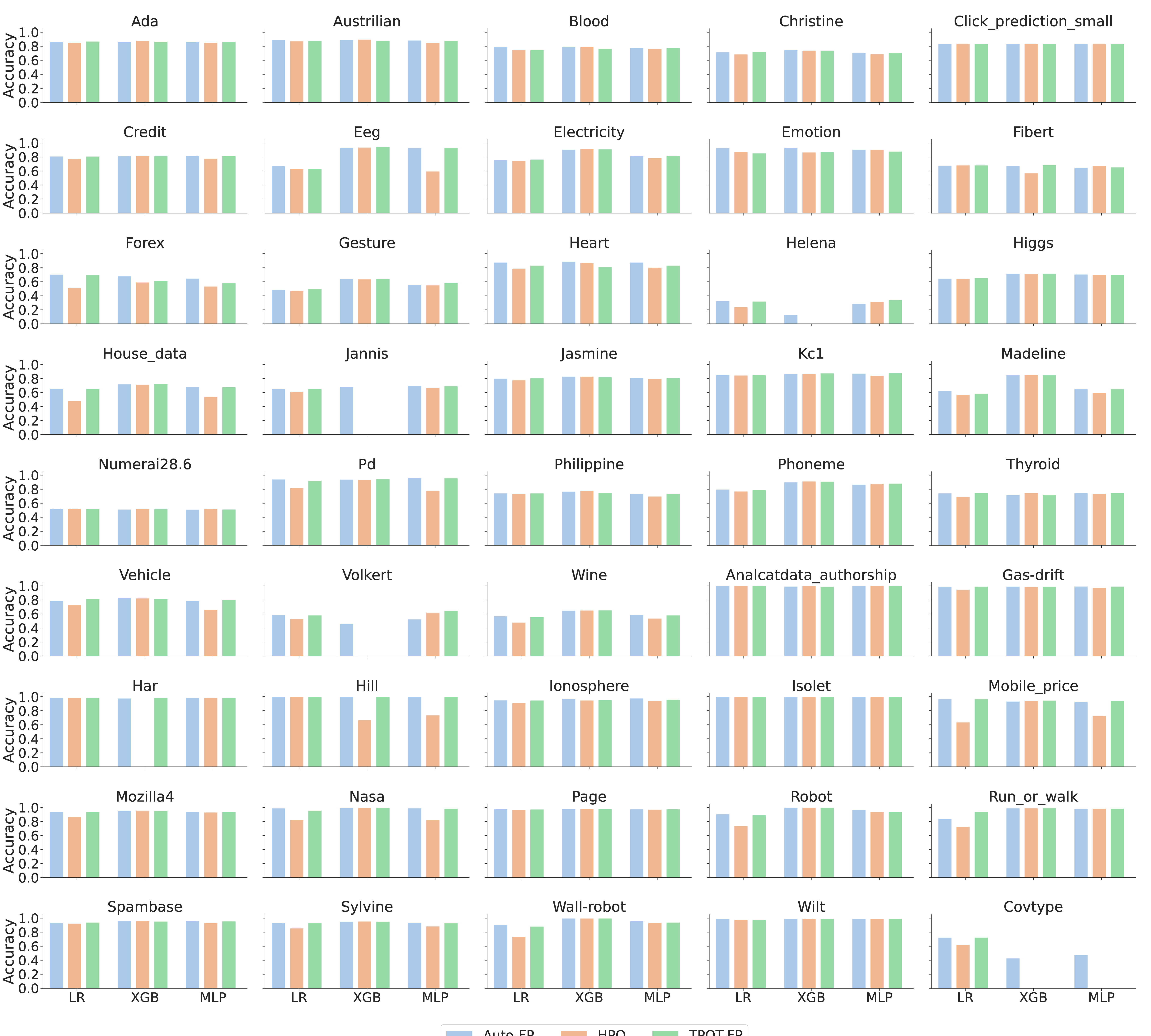}
\vspace{-1.5em}
\caption{Evaluate Auto-FP in an AutoML context (default search space, related to Figure 10).}
\label{fig:automl_default_space}
\end{figure*}

\begin{figure*}[h]
\vspace{-.5em}
\centering
\includegraphics[width=\textwidth]{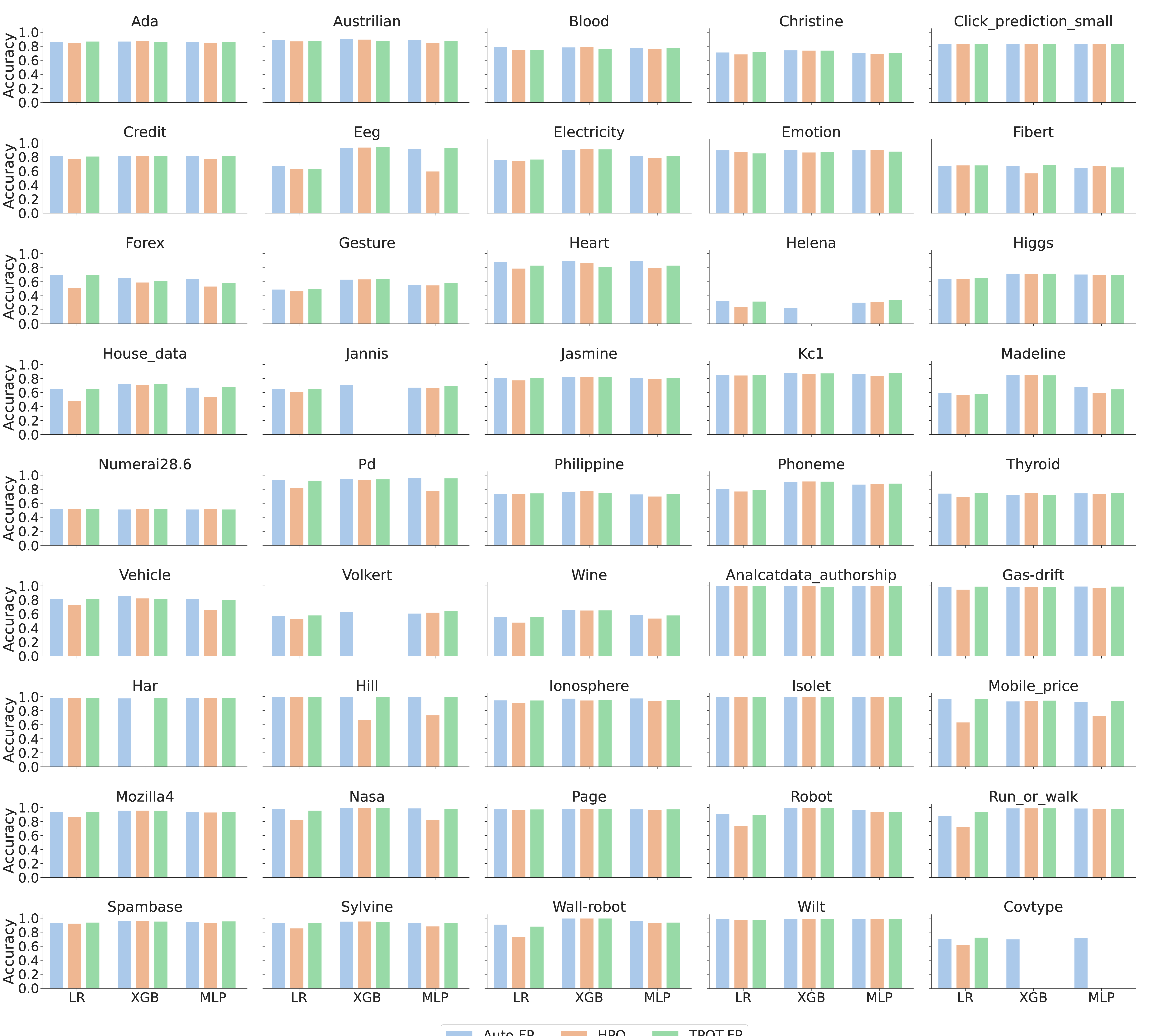}
\vspace{-1.5em}
\caption{Evaluate Auto-FP in an AutoML context (extended search space in Table 5, related to Figure 11).}
\label{fig:automl_balanced_space}
\end{figure*}

\end{appendix}



\end{document}